\documentclass[11pt,]{article}
\pdfoutput=1
\usepackage{todonotes}

\usepackage[linktocpage,colorlinks,linkcolor=blue,anchorcolor=blue,citecolor=blue,urlcolor=blue,pagebackref]{hyperref}
\usepackage{tikz}
\RequirePackage[authoryear]{natbib}
\setcitestyle{authoryear,open={(},close={)}} 

\usepackage{mathtools,amssymb}

\usepackage[margin=1in]{geometry}

\usepackage{booktabs}
\usepackage{mathtools,amsthm}

\newcommand{\mb}{\mathbb}
\newcommand{\mc}{\mathcal}

\newcommand{\dee}{\mathrm{d}}
\newcommand{\Knu}{K^{\nu}}

\usepackage{mathtools}

\allowdisplaybreaks

\renewcommand*{\backref}[1]{\ifx#1\relax \else Page #1 \fi}
\renewcommand*{\backrefalt}[4]{%
  \ifcase #1 \footnotesize{(Not cited.)}%
  \or        \footnotesize{(Cited on page~#2.)}%
  \else      \footnotesize{(Cited on pages~#2.)}%
  \fi
}

\renewcommand{\ker}{\textsf{k}}

\newcommand{\KL}{\mathsf{KL}}

\newcommand{\Tm}{\mathsf{T}}
\newcommand{\En}{\mathsf{E}}
\newcommand{\id}{\mathsf{id}}
\newcommand{\HS}{\mathsf{HS}}

\makeatletter
\newcommand*{\rom}[1]{\expandafter\@slowromancap\romannumeral #1@}
\makeatother

\newcommand{\namedproof}[1]{\renewcommand{\proofname}{#1}\begin{proof}}

\newtheorem{theorem}{Theorem}
\newtheorem{assump}{Assumption}
\newtheorem{assumption}{Assumption}
\newtheorem{lemma}{Lemma}
\newtheorem{corollary}{Corollary}

\newtheorem{proposition}{Proposition}

\newtheorem{remark}{Remark}

\theoremstyle{definition}

\title{Finite-Particle Rates for \\ Regularized Stein Variational Gradient Descent}
\author{Ye He$^1$, Krishnakumar Balasubramanian$^2$, Sayan Banerjee$^3$, Promit Ghosal$^4$\\ \\
$^1$Department of Mathematics, Georgia Institute of Technology. \\
$^2$Department of Statistics, University of California, Davis. \\
$^3$Department of Statistics and Operations Research, University of North Carolina, Chapel Hill. \\
$^4$Department of Statistics, University of Chicago. 
}

\begin{document}
\maketitle

\begin{abstract}%
We derive finite-particle rates for the regularized Stein variational gradient descent (R-SVGD) algorithm introduced by~\cite{he2024regularized} that corrects the constant-order bias of the SVGD by applying a resolvent-type preconditioner to the kernelized Wasserstein gradient. For the resulting interacting $N$-particle system, we establish explicit non-asymptotic bounds for time-averaged (annealed) empirical measures, illustrating convergence in the \emph{true} (non-kernelized) Fisher information and, under a $\mathrm{W}_1\mathrm{I}$ condition on the target, corresponding $\mathrm{W}_1$ convergence for a large class of smooth kernels. Our analysis covers both continuous- and discrete-time dynamics and yields principled tuning rules for the regularization parameter, step size, and averaging horizon that quantify the trade-off between approximating the Wasserstein gradient flow and controlling finite-particle estimation error.
\end{abstract}

\section{Introduction}

Given a potential $V:\mathbb{R}^d\to\mathbb{R}$, we consider the problem of sampling from the target density $\pi(x)=Z^{-1}e^{-V(x)}$ where $Z=\smallint_{\mathbb{R}^d} e^{-V(x)}\,dx,$
where the normalizing constant $Z$ is typically unknown or intractable. This task is ubiquitous in Bayesian statistics, machine learning, and numerical integration. A useful modern viewpoint interprets sampling as \emph{optimization over probability measures}~\citep{jordan1998variational,wibisono2018sampling} based on the Wasserstein Gradient Flow (WGF) of the relative entropy,
\begin{equation}\label{eq:intro_WGF}
\partial_t \rho_t
=\nabla\cdot\!\left(\rho_t \nabla_{W_2} \mathrm{KL}(\rho_t\mid \pi)\right)
=\nabla\cdot\!\left(\rho_t \nabla \log\frac{\rho_t}{\pi}\right).
\end{equation}
Exact space--time discretization of~\eqref{eq:intro_WGF} by empirical measures is impossible due to the diffusion term, motivating deterministic approximations via mollification and related constructions.

Stein Variational Gradient Descent (SVGD) is a prominent deterministic approximation scheme proposed by~\cite{liu2016stein,liu2017stein}, that has been widely adopted in machine learning and applied mathematics as a deterministic discretzation of~\eqref{eq:intro_WGF}. It corresponds to a kernelized, constant-order approximation of the WGF, often written as the Stein Variational Gradient Flow (SVGF)
\begin{equation}\label{eq:intro_SVGF}
\partial_t \rho_t
=\nabla\cdot\!\left(\rho_t\,\mathcal{T}_{k,\mu_t}\nabla \log\frac{\mu_t}{\pi}\right),
\end{equation}
where $\mathcal{T}_{k,\rho}$ is the integral operator induced by the reproducing kernel $k$; see Section~\ref{app:RKHS} for a short overview of the \emph{reproducing kernel Hilbert space} background required for this work. While this kernelization yields a practical deterministic particle dynamics, it also introduces a systematic bias: for common bounded translation-invariant characteristic kernels, $\|\mathcal{T}_{k,\mu_t}-I\|_{\mathrm{op}}$ need not be small, so the quality of the approximation depends critically on the kernel choice.

To address this,~\cite{he2024regularized} proposed the \emph{Regularized}-Stein Variational Gradient Flow (R-SVGF) that applies a resolvent-type correction,
\begin{equation}\label{eq:intro_RSVGF}
\partial_t \rho_t
=\nabla\cdot\!\left(\rho_t\,\big((1-\nu)\mathcal{T}_{k,\rho_t}+\nu I\big)^{-1}\mathcal{T}_{k,\rho_t}\nabla \log\frac{\rho_t}{\pi}\right),
\end{equation}
with $\nu\in(0,1]$. The factor $\big((1-\nu)\mathcal{T}_{k,\rho_t}+\nu I\big)^{-1}\mathcal{T}_{k,\rho_t}$ ``de-biases'' the kernelized gradient: whenever $\nabla\log(\rho_t/\pi)$ lies in (or is sufficiently regular relative to) the range of $\mathcal{T}_{k,\rho_t}$, this correction converges to the true Wasserstein gradient as $\nu\to 0$, with a rate that improves under additional smoothness.

Practical implementations of SVGF and R-SVGF approximate the evolving density with $N$ interacting particles and a time discretization, so a complete convergence theory ultimately requires \emph{finite-particle, discrete-time} rates—yet such results for SVGD have been notoriously difficult. As a first step toward this goal, several works studied \emph{mean-field} convergence (finite-particle dynamics to the limiting flow in continuous time) using PDE and propagation-of-chaos techniques~\citep{lu2019scaling,gorham2020stochastic,korba2020non,carrillo2025convergence}. Moving to discrete time, \cite{shi2024finite} obtained the first finite-particle rates in the KSD metric (cf.~\eqref{eq:Stein fisher}), but only at the sub-optimal order $1/\sqrt{\log\log N}$; \cite{liu2024towards} proved improved parametric rates in the special \emph{Gaussian SVGD} setting (Gaussian target and initialization, bilinear kernel); and most recently, \cite{balasubramanian2024improved} established state-of-the-art $1/\sqrt{N}$ finite-particle, continuous and discrete-time rates, in KSD (a double-exponential improvement over~\cite{shi2024finite}) and, leveraging~\cite{kanagawa2025controlling}, also derived $W_2$ guarantees with Mat\'ern kernels. See Table~\ref{tab:comparison} for a brief summary.

\begin{table}[t]
\centering
\caption{Convergence results for discrete-time SVGD-type algorithms. Here $I$ denotes the true (i.e., non-kernelized) Fisher information and KSD corresponds to Kernel Stein Discrepancy, also known as kernelized Fisher Information. For RKHS kernel bounded by $B$, $\mathrm{KSD}\le B I$.}
\label{tab:comparison}
\begin{tabular}{lcccc}
\toprule

Work & Method  & Particle Setting & Metric & Convergence Rate  \\

\midrule

\cite{shi2024finite} & SVGD & finite $N$ & KSD & $\sqrt{1/\log\log N}$ \\

\cite{liu2024towards} & Gaussian SVGD & finite $N$ & 
\begin{tabular}[c]{@{}c@{}}
convergence of \\
covariance parameter
\end{tabular} 
& $1/\sqrt{N}$  \\
 
\cite{balasubramanian2024improved} & SVGD & finite $N$ & KSD & $1/\sqrt{N}$ \\

\cite{he2024regularized} & R-SVGF & infinite $N$ & $I$ and KL-divergence & Not Proved  \\

\textbf{This work} & R-SVGD & finite $N$ & $I$ and $W_1^2$~\footnotemark[1] & $N^{-(1-\alpha-c(3-2\alpha))}$~\footnotemark[2] \\

\bottomrule

\end{tabular}
\end{table}

\footnotetext[1]{The convergence of $W_1^2$ requires extra functional inequality assumption on the target. See details in Section \ref{sec:rsvgdalgorithm}.}

\footnotetext[2]{The constant $\alpha\in [0,1/2]$ depends on the target potential's growth and the regularizer is of scale $N^{-c}$ for $c\in [0, \tfrac{1-\alpha}{3-2\alpha})$. }



Our main results provide \emph{finite-particle, non-asymptotic} guarantees for regularized SVGD in both continuous and discrete time, and they clarify how regularization enables control in \emph{true} non-kernalized Fisher information rather than only kernelized (Stein) quantities.

In continuous time, we show that an annealed (time-averaged) empirical law enjoys quantitative decay of the Fisher information along the interacting particle dynamics. Concretely, the average Fisher information over a time horizon $[0,T]$ can be bounded by a sum of (i) an initialization term that decreases as the run time increases, (ii) an interaction term of order $1/N$, and (iii) a regularization-induced term that grows with $T$ but is suppressed by larger $N$ and $\nu$. Choosing an appropriate averaging window (of order $N^{2/3}$) balances these effects and yields explicit rates for the time-averaged measure $\mu_{av}^N$, showing that the Fisher information (and, under a standard $\mathrm{W}_1\mathrm{I}$ inequality for the target, also the $\mathrm{W}_1$ distance) decreases with $N$ at rates that are \emph{kernel-agnostic in the exponent}. This stands in contrast with classical SVGD analyses, which typically control only the weaker Stein Fisher information and may require specialized structured kernels  to obtain Wasserstein guarantees.

In discrete time, we complement the continuous-time theory with step-size and regularization schedules that make the finite-$N$ bounds explicit for the practical R-SVGD algorithm. A resulting corollary identifies two regimes: when $\nu$ is kept very close to $1$, the method behaves like SVGD and achieves an $N^{-1}$-type accuracy in the regularized Stein metric (up to dimension- and kernel-dependent factors), matching the usual Monte Carlo-type rates for averaged quantities. Moreover, when $\nu$ is allowed to \emph{decrease} with $N$, the dynamics become progressively closer to the Wasserstein gradient flow, and we can guarantee convergence in the \emph{true} non-kernelized Fisher information and in $\mathrm{W}_1$ for the annealed iterate---at the price of a slower rate that explicitly captures the estimation cost of inverting the regularized operator. The theory pinpoints a sharp trade-off: $\nu$ should not decay too fast (roughly, slower than $N^{-1/3}$ in the stated regime) to ensure the finite-particle error still vanishes, thereby providing principled guidance for tuning $\nu$, $h$, and the averaging horizon in practice.

\subsection{R-SVGD Algorithm}
We end this introduction with a brief introduction the R-SVGD algorithm~\cite{he2024regularized}; more details and motivation are provided in Section~\ref{sec:rsvgdprelim}. Let $(X_n^i)_{i=1}^N\subset \mathbb{R}^d$ denote the particle locations at iteration $n$, and write
$\underline{X}_n \coloneqq [X_n^1,\ldots,X_n^N]^\intercal \in \mathbb{R}^{dN}$ for the stacked particle vector. For any vector field $f:\mathbb{R}^d\to\mathbb{R}^d$, define the evaluation operator $L_n f \;\coloneqq\; [f(X_n^1),\ldots,f(X_n^N)]^\intercal.$
The R-SVGD update is obtained by an explicit time discretization of the regularized particle dynamics (see Section~\ref{sec:particledynamic}), where the kernel interaction is encoded through the Gram matrix. Specifically, let $K_n\in\mathbb{R}^{N\times N}$ be the Gram matrix with entries $(K_n)_{ij}=k(X_n^i,X_n^j)$ for $i,j\in[N]$, and let $I_N$ denote the $N\times N$ identity. Given a step-size sequence $(h_n)_{n\ge 1}$ and a regularization schedule $(\nu_n)_{n\ge 1}\in (0,1]^{\mathbb{N}}$, the particles are updated according to
\begin{equation}\label{eq:algorithm}
\underline{X}_{n+1}
=
\underline{X}_n
-
h_{n+1}
\bigg(\frac{1-\nu_{n+1}}{N}K_n+\nu_{n+1}I_N\bigg)^{-1}
\bigg(
\frac{1}{N}K_n(L_n\nabla V)
-
\frac{1}{N}\sum_{j=1}^N L_n\nabla k(X_n^j,\cdot)
\bigg)
\end{equation}
where $\nabla k(X_n^j,\cdot)$ denotes the map $x\mapsto \nabla_2 k(X_n^j,x)$ evaluated at the particle locations via $L_n$. The term $\frac{1}{N}K_n(L_n\nabla V)$ corresponds to the attractive drift induced by the target potential $V$, while $\frac{1}{N}\sum_{j=1}^N L_n\nabla k(X_n^j,\cdot)$ is the repulsive interaction that promotes particle diversity. The matrix factor $\left(\tfrac{1-\nu_{n+1}}{N}K_n+\nu_{n+1}I_N\right)^{-1}$ plays the role of the regularized inverse (a resolvent-type preconditioner), correcting the kernelization bias: when $\nu_{n+1}\approx 1$ the update reduces to SVGD, whereas smaller $\nu_{n+1}$ increasingly de-kernelizes the direction and moves the method closer to the Wasserstein gradient flow regime analyzed later.

\subsection{Main challenges}

We emphasize some central challenges in the analysis of R-SVGD~\eqref{eq:algorithm}. At the finite-particle level, the velocity contains the empirical resolvent
$
\left(\frac{1-\nu}{N}K+\nu I_N\right)^{-1},
$
where \(K\) is the particle Gram matrix. Thus, differentiating the \(N\)-particle KL divergence produces not only the usual SVGD self-interaction terms, as in~\cite{balasubramanian2024improved}, but also resolvent-derivative terms involving higher-order particle interactions. These terms become increasingly singular as \(\nu\) decreases and cannot be controlled by a uniform \(O(d/N)\)-type estimate alone; instead, they must be bounded recursively through the regularized Fisher information along the particle trajectory; see Lemma~\ref{lem:bound C*}. This leads to a much more involved and nuanced technical analysis compared to ~\cite{balasubramanian2024improved}. Further, the analysis must also interpolate between two competing regimes: for \(\nu\approx1\), R-SVGD is close to classical SVGD and finite-particle errors are stable, but the geometry remains essentially kernelized; for small \(\nu\), the dynamics better approximate the true Wasserstein gradient flow, but the empirical inverse becomes ill-conditioned and amplifies finite-particle errors. Balancing this trade-off through appropriate choices of \(\nu\), the averaging horizon, and the step size in our finite-particle analysis, is another key distinction from~\cite{balasubramanian2024improved}.

\section{Convergence of Finite-Particle R-SVGF} \label{sec:rsvgddynamics}
We start with the following assumptions on the kernel $k$ and the potential function $V$. 

\begin{assumption}\label{assump:kernel} The kernel function $k:\mb{R}^d\times \mb{R}^d \to \mb{R}_+ $ satisfies the following conditions: [(1)] $k$ is symmetric, [(2)] $k$ is positive semi-definite, i.e., $\iint_{\mb{R}^d\times\mb{R}^d} f(x)k(x,y)f(y)\dee x\dee y \ge 0$ for all $f\in \mc{C}(\mb{R}^d)$, and [(3)] (Boundedness) $k$ and all its partial derivatives up to order $2$ are uniformly bounded by $B\in (0,\infty)$.
\end{assumption}
 
Assumption~\ref{assump:kernel} holds, for example, for the Gaussian/RBF kernel $k(x,y)=\exp\!\big(-\|x-y\|_2^2/(2\ell^2)\big)$, the rational quadratic kernel $k(x,y)=\big(1+\|x-y\|_2^2/(2\alpha\ell^2)\big)^{-\alpha}$, Mat\'ern kernels with smoothness parameter $\nu>1$ (ensuring bounded derivatives up to order two), the inverse multiquadric (IMQ) kernel $k(x,y)=(c^2+\|x-y\|_2^2)^{-\beta}$, and sufficiently smooth compactly supported Wendland kernels. In contrast, unbounded kernels such as polynomial kernels do not satisfy (3) on $\mathbb{R}^d$.
\begin{assumption}\label{assump:target} The target potential $V:\mb{R}^d\to \mb{R}$ is twice continuously differentiable and smooth, and there exists a constant $C_V\in (0,\infty)$ such that $\|\nabla^2 V(x)\|_{\mathrm{op}} \le \, C_V$ for all $x\in \mb{R}^d$, where $\|\cdot\|_{\mathrm{op}}$ denotes the operator norm.
\end{assumption}

This assumption enforces global $C^2$-smoothness of $V$ with a uniformly bounded Hessian, i.e., the gradient $\nabla V$ is Lipschitz continuous with Lipschitz constant at most $C_V$. We now introduce the following discrepancy measures which are used in the subsequent analysis:
\begin{align}
 I(\rho|\pi)&\coloneqq\left\langle \nabla \log \frac{\rho}{\pi}, \nabla \log \frac{\rho}{\pi} \right\rangle_{L_2^d(\rho)}, \label{eq:fisher}\\
 I_{\text{Stein}}(\rho|\pi) &\coloneqq \left\langle  \nabla \log \frac{\rho}{\pi},  \iota_{k,\rho}\iota_{k,\rho}^* \nabla \log \frac{\rho}{\pi}\right\rangle_{L_2^d(\rho)}, \label{eq:Stein fisher}\\
    I_{\nu,\text{Stein}}(\rho|\pi) &\coloneqq \left\langle  \nabla \log \frac{\rho}{\pi}, \left((1-\nu)\iota_{k,\rho}^*\iota_{k,\rho}+\nu I_d \right)^{-1} \iota_{k,\rho}\iota_{k,\rho}^* \nabla \log \frac{\rho}{\pi}\right\rangle_{L_2^d(\rho)}. \label{eq:regularized fisher}
\end{align}

These quantities measure the discrepancy between $\rho$ and $\pi$ through the score $\nabla\log(\rho/\pi)$ under different geometries. 
The standard Fisher information $I(\rho|\pi)$ is the $L_2(\rho)$-squared norm of this score and corresponds to the ``full'' Wasserstein gradient magnitude, while $I_{\mathrm{Stein}}(\rho|\pi)$ (also known as the $\text{KSD}^2$ distance) replaces it by its kernelized version via the integral operator $\iota_{k,\rho}\iota_{k,\rho}^*=\mathcal{T}_{k,\rho}$, matching the SVGF geometry. 
Finally, $I_{\nu,\mathrm{Stein}}(\rho\mid\pi)$, which was introduced by~\cite{he2024regularized} and referred to as the \emph{regularized Fisher information},  inserts the resolvent-type preconditioner $\big((1-\nu)\iota_{k,\rho}^*\iota_{k,\rho}+\nu I_d\big)^{-1}$, interpolating between the Stein and Wasserstein scales and recovering the latter as $\nu\downarrow 0$ under suitable regularity.

\subsection{Convergence of regularized Fisher information}

The convergence of the continuous-time finite particle system depends on the interaction between the kernel function $k$ and the potential function $V$. As in \cite[Theorem 1]{balasubramanian2024improved}, the (regularized) gradient flow structure of the dynamics leads to the Kullback-Leibler ($\KL$) distance between the \emph{joint density} $p^N(\cdot)$ of the particle locations and the $N$-fold target $\pi^{\otimes N}$ behaving as an \emph{approximate Lyapunov function} with a dominant decreasing component in time, characterized in terms of the regularized Fisher information $I_{\nu,\text{Stein}}$. However, the dependence of the driving vector field of any particle on its own location, which is negligible in the population limit, does exert a non-trivial effect on the finite particle dynamics. This is captured through the \emph{self-interaction coefficient} $C^*$ described as follows. For $\underline{x}=(x^1,\ldots,x^N)\in\mathbb{R}^{dN}$, let $K =\big(k(x^i,x^j)\big)_{i,j\in[N]}\in\mathbb{R}^{N\times N},$ and $K_\nu \coloneqq \Big(\tfrac{1-\nu}{N}K+\nu I_N\Big)^{-1},$ and define (with $\nabla_1,\nabla_2$ and $\Delta_2$ denoting derivatives with respect to the first/second argument) 
{\small
\begin{align}
    C^*(\underline{x})=&-\frac{1}{N}\sum_{i,j=1}^N \Knu_{ij}\big( \Delta_2 k(x^j,x^i)-\nabla_1 k(x^i,x^j)\cdot\nabla V(x^i) -k(x^j,x^i)\Delta V(x^i)  \big) \nonumber\\
    +&\frac{1-\nu}{\nu}\frac{1}{N^2}\sum_{i,j,l=1}^N k(x^i,x^l)\Knu_{li}\big(\nabla_1\cdot\nabla_2 k(x^i,x^j)-\nabla_1 k(x^i,x^j)\cdot\nabla V(x^j)\big) \nonumber\\
    +&\frac{1-\nu}{N^2}\sum_{i,j,k,l=1}^N \Knu_{ij}\Knu_{ik}\nabla_1 k(x^i,x^j)\cdot \big(\nabla_2 k({x^k},x^l)-k({x^k},x^l)\nabla V(x^l)\big)\nonumber\\
     -& \frac{(1-\nu)^2}{\nu}\frac{1}{N^3}\sum_{i,j,k,l,r=1}^N \Knu_{ri} \Knu_{jk} k(x^i,x_r)\nabla_1 k(x^i,x^j)\cdot\big(\nabla_2 k({x^k},x^l)-k({x^k},x^l)\nabla V(x^l)\big). \label{eq:C*} 
\end{align}}
Unlike in the usual SVGD framework~\cite{balasubramanian2024improved}, this functional takes a significantly more complex form in the regularized set-up, as described below. The first line approaches the ``SVGD-type'' self-interaction term when $\nu \rightarrow 1$, while the remaining lines are \emph{purely regularization-induced higher-order interactions} that originate from differentiating and coupling the resolvent preconditioner with the Stein drift. When $\nu=1$, the higher-order interactions disappear recovering the $C^*$ from~\cite{balasubramanian2024improved}. We also remark that the presence of these higher-order interaction terms make our proofs significantly more involved compared to the usual SVGD case in ~\cite{balasubramanian2024improved}, and we will highlight the differences in the rest of the draft.

As in \cite{balasubramanian2024improved}, the uniform boundedness of the SVGD type self-interaction term plays a crucial role which we capture in the following assumption.

\begin{assumption}\label{ass:cstarfin}
Assume there exists $c^* \in (0,\infty)$ such that
    \begin{align}\label{cstarfinite}
    \sup_{z \in \mathbb{R}^d}\bigg|\frac{1}{N}\sum_{i=1}^N \big( \Delta_2 k(z,z)-\nabla_1 k(z,z)\cdot\nabla V(z) -k(z,z)\Delta V(z)  \big)\bigg| =c^*d.
\end{align}
\end{assumption}


In the following theorem, we exhibit the relationship between the decay rate of $\KL, I_{\nu,\text{Stein}}$ and the interaction-coefficient $C^*$ which, in turn, furnishes rates of convergence in $I_{\nu,\text{Stein}}$.

\begin{theorem}\label{thm:convergence} Under Assumption \ref{assump:kernel} - (1) and (2), let $\{x^i(t)\}_{i=1}^N$ be the set of $N$ particles along the R-SVGD dynamics, $p^N(t)$ be the joint distribution of $\underline{x}(t)\coloneqq(x^1(t),\cdots,x^N(t))\in \mb{R}^{dN}$ and $\rho^N(t)=\frac{1}{N}\sum_{j=1}^N \delta
_{x^j(t)}$ be the empirical distribution at time $t$. Then for any $T>0$, we get
\begin{align}
    &\mb{E}[I_{\nu,\text{Stein}}(\rho_{av,T}^N|\pi)]\le \frac{1}{T}\int_0^T \mb{E} \big[ I_{\nu,\text{Stein}}(\rho^N(t)|\pi) \big] \dee t\le \frac{\KL(p_0^N|\pi^{\otimes N})}{NT}+\frac{\int_0^T \mb{E}[C^*(\underline{x}(t))]\dee t}{NT}, \label{eq:convergence of regularized Stein fisher}\\
    &\KL(p^N(T)|\pi^{\otimes N})\le \KL(p_0^N|\pi^{\otimes N})+\int_0^T \mb{E}[C^*(\underline{x}(t))]\dee t,\label{eq:KL decay}
\end{align}
where $C^*(\underline{x})$ is given in \eqref{eq:C*}, $p^N(0)=p_0^N$ and let $\rho^N_{av,T}\coloneqq \tfrac{1}{T}\int_0^T \rho^N(t) \dee t$, the time-averaged empirical measure of the particle system over the horizon. If we further assume Assumption \ref{assump:kernel}-(3) and Assumptions \ref{assump:target} and \ref{ass:cstarfin}, then $\mb{E}_{\underline{x}\sim p^N(t)}[C^*(\underline{x})]$ can be estimated appropriately (see Lemma \ref{lem:bound C*} and Remark \ref{rem:c^* bound}), and we get
\begin{align}
     &\mb{E}[I_{\nu,\text{Stein}}(\rho_{av,T}^N|\pi)] \le\frac{1}{T}\int_0^T \mb{E} \big[ I_{\nu,\text{Stein}}(\rho^N(t)|\pi) \big] \dee t \nonumber \\
    \lesssim &  \frac{\KL(p_0^N|\pi^{\otimes N})}{NT} + \frac{c^*d}{N} + (1-\nu)\bigg(\frac{\beta_1 d + \beta_2 \mb{E}[\mb{E}_{y\sim \rho^N(0)} [ \| \nabla V(y) \|^2 ]^{\frac{1}{2}}] d^{\frac{1}{2}}}{N} + \frac{\beta_2^2 C_V^2 d T^2} {\nu N^2 }\bigg) \label{eq:improved convergence regularized fisher}\\
    &\KL(p^N(T)|\pi^{\otimes N})\nonumber\\
    \lesssim & \KL(p^N_0|\pi^{\otimes N}) + c^*dT + (1-\nu)T\big(\beta_1  d + \beta_2  \mb{E}[\mb{E}_{y\sim \rho^N(0)} [ \| \nabla V(y) \|^2 ]^{\frac{1}{2}}] d^{\frac{1}{2}} + \frac{\beta_2^2 C_V^2 d T^2} {\nu N }\big) \label{eq:KL bound}
\end{align}
where $\beta_1,\beta_2$ are defined as
\begin{align}
\beta_1
&= B(1+C_V)
+ \frac{2B^2}{\nu^2}
+ \frac{B^3}{\nu^3}(1-\nu),\  \beta_2= B
+ \frac{2B^2}{\nu^2}
+ \frac{B^3}{\nu^3}(1-\nu). \label{eq:constants alpha beta}
\end{align}
\end{theorem}

\begin{remark}[Challenge in estimating $\mathbb{E}\lbrack C^*(\underline{x}(t))\rbrack$]\label{rem:c^* bound} Unlike in \cite[Theorem 1]{balasubramanian2024improved}, the quantity $\mb{E}[C^*(\underline{x}(t))]$ in \eqref{eq:convergence of regularized Stein fisher} and \eqref{eq:KL decay} cannot be simply bounded by a uniform constant that is linear in dimension $d$, when $\nu < 1$. Instead, we bound $\mb{E}[C^*(\underline{x}(t))]$ in terms of constants as well as the regularized Fisher information, in the form of $\int_0^t \mb{E}[I_{\nu,\text{Stein}}(\rho^N(s)|\pi)^{\frac{1}{2}}]\dee s$, along the trajectory of the process. Then the explicit bounds in \eqref{eq:improved convergence regularized fisher} and \eqref{eq:KL bound} follow by a recursive integral inequality. See Lemma \ref{lem:bound C*} and the proof of Theorem \ref{thm:convergence} in Appendix \ref{sec:proof thm 1} for details.  
\end{remark}

\begin{corollary}\label{cor:average regularize fisher} 
Define $\rho_{av}^{N}\coloneqq \tfrac{1}{(1-\nu)^{-\frac{1}{3}}N^{\frac{2}{3}}}\int_0^{T=(1-\nu)^{-\frac{1}{3}}N^{\frac{2}{3}}} \rho^N(t)\dee t $.  Then we have
\begin{align*}
    \mb{E}[I_{\nu,\text{Stein}}(\rho_{av}^N|\pi)]
    &\lesssim \frac{\KL(p_0^N|\pi^{\otimes N})}{(1-\nu)^{-\frac{1}{3}}N^{\frac{5}{3}}} + \frac{c^*d}{N} + (1-\nu)\frac{\beta_1 d + \beta_2 \mb{E}_{y\sim \rho^N(0)} [ \| \nabla V(y) \|^2 ]^{\frac{1}{2}}d^{\frac{1}{2}}}{N} +\frac{\nu^{-1}\beta_2^2 C_V^2 d}{(1-\nu)^{-\frac{1}{3}}N^{\frac{2}{3}}}.
\end{align*}
In particular, if $\nu = \nu_N = 1- \frac{1}{N}$, then we have
\begin{align*}
    \mb{E}[I_{\nu_N,\text{Stein}}(\rho_{av}^N|\pi)]
    &\lesssim \frac{\KL(p_0^N|\pi^{\otimes N})}{N^2} + \frac{c^*d}{N} + \frac{\beta_{1,N} d + \beta_{2,N} \mb{E}_{y\sim \rho^N(0)} [ \| \nabla V(y) \|^2 ]^{\frac{1}{2}}d^{\frac{1}{2}}}{N^2} +\frac{\nu_N^{-1}\beta_2^2 C_V^2 d}{N},
\end{align*}
where $\beta_{1,N}, \beta_{2,N}$ are defined in \eqref{eq:constants alpha beta} taking $\nu_N$ in place of $\nu$.
\end{corollary}

\begin{remark}\label{cor:optscale}
    Under the natural assumption that $\KL(p_0^N|\pi^{\otimes N})=\mc{O}(N)$, the choice of $T=(1-\nu)^{-\frac{1}{3}}N^{\frac{2}{3}}$ above is made in order to match the order of the first and fourth terms in \eqref{eq:improved convergence regularized fisher} (ones involving $T$). This `optimizes' the bound (in the sense of maximizing decay rate order) and gives a decay rate for $\mb{E}[I_{\nu,\text{Stein}}(\rho_{av}^N|\pi)]$ of order $(1-\nu)^{\frac{1}{3}}N^{-\frac{2}{3}}$.
\end{remark}
\begin{remark}[Matching SVGD bounds under regularization for $\nu$ close to $1$]
    From the above result, we note that the bounds match the performance of SVGD in \cite{balasubramanian2024improved} when $\nu=1$. However, to match the order of the bounds, it suffices to take $\nu = \nu_N = 1- \frac{1}{N}$. This choice will ensure that the regularization effect of R-SVGD still persists for every finite $N$, while its performance is at least as good as SVGD. However, as we will see in Section \ref{sec:FishWass} (see Remark \ref{fisherconrem}), the efficacy of R-SVGD as an approximation to the Wasserstein gradient flow worsens as $\nu \rightarrow 1$.
\end{remark}

\begin{remark}[Convergence rate for small $\nu$]
    Under the natural scaling $\KL(p_0^N | \pi^{\otimes N})=\mathcal{O}(N)$, the bound simplifies when $\nu\to 0$, as $(1-\nu)^{1/3}=1+o(1)$, to
\[
\mathbb{E}\!\left[I_{\nu,\mathrm{Stein}}(\rho_{av}^N\mid\pi)\right]
\;\lesssim\;
\mathcal{O}\!\left({N^{-2/3}}\right)
\;+\;
\mathcal{O}\!\left({N^{-1}}\right)
\;+\;
\mathcal{O}\!\left({\nu^{-1}\,N^{-2/3}}\right).
\]
Therefore, along any joint limit $N\to\infty$ and $\nu=\nu_N\to 0$, the right-hand side goes to $0$ \emph{if and only if} $\nu_N\,N^{2/3}\to\infty$
 or equivalently, $\nu_N=\omega(N^{-2/3})$. For example, choosing $\nu_N=N^{-\alpha}$ with any $\alpha\in(0,2/3)$ and the near-minimal choice $\nu_N=N^{-2/3}\log N$ respectively ensure convergence and yield the rates
\begin{align*}
\mathbb{E}\!\left[I_{\nu_N,\mathrm{Stein}}(\rho_{av}^N\mid\pi)\right]
=
\mathcal{O}\!\left(N^{-(2/3-\alpha)}\right),
\quad\text{and}\quad
\mathbb{E}\!\left[I_{\nu_N,\mathrm{Stein}}(\rho_{av}^N\mid\pi)\right]
=
\mathcal{O}\!\left({1}/{\log N}\right).
\end{align*}
This shows that as $\nu$ goes to zero, while the metric is strong (compared to $I_{\text{Stein}}$), the rates get worse.
\end{remark}

\noindent Under exchangeability of the law of initial particles, Theorem \ref{thm:convergence} implies weak convergence to the target distribution of the time-averaged single-particle marginals.
\begin{theorem}\label{thm:weak convergence under exchangeability} Under the conditions in Theorem \ref{thm:convergence}, suppose that the law $p_0^N$ of the initial particles $x^1(0),x^2(0), \cdots, x^N(0)$ is exchangeable for each $N\in \mb{N}$, and $\limsup_{N \rightarrow \infty} \frac{\KL(p_0^N|\pi^{\otimes N})}{N} < \infty$.   
Define $\bar{\rho}^N(\cdot)\coloneqq \tfrac{1}{(1-\nu)^{-1/3}N^{2/3}}\int_0^{(1-\nu)^{-1/3}N^{2/3}} \mb{P}(x^1(t)\in \cdot) \dee t$. Then $I_{\nu, \text{Stein}}(\bar{\rho}^N|\pi)\to 0$ and $\bar{\rho}^N\rightharpoonup \pi$ as $N\to\infty$, where `$\rightharpoonup$' denotes weak convergence. This result continues to hold if $\nu = \nu_N = 1 - \tfrac{1}{N}$.
\end{theorem}

\subsection{Convergence in Fisher information and Wasserstein distance}\label{sec:FishWass} In this section, we analyze convergence in Wasserstein-1 distance ($\mathrm{W}_1$) along the R-SVGD dynamics. Our analysis is performed in 2 parts: (1) deriving convergence in Fisher information based on the convergence in regularized Fisher information in Theorem \ref{thm:convergence} and the equivalence relation between Fisher information and the regularized Fisher information as studied in \cite{he2024regularized}; (2) deriving convergence of $\mathrm{W}_1$ under the transport-information inequality assumption on the target distribution, relating the Fisher information convergence in (1) to $\mathrm{W}_1$-convergence.

\begin{assumption}[Transport-Information $(\mathrm{W}_1\mathrm{I}$) inequality] A distribution $\mu\in \mb{R}^d$ satisfies the $W_1$-information inequality with parameter $C_{\mu}$ if $W_1(\rho,\mu)\le C_\mu \sqrt{I(\rho|\mu)},~~ \forall \rho\in \mc{P}(\mb{R}^d).$
\end{assumption}

\noindent We start from part (1) by stating the equivalence result in \cite{he2024regularized}. It is worth noting that the Fisher information $I(\cdot|\pi)$ is not well-defined for empirical distributions. Therefore, instead of looking at $I(\rho^N(t)|\pi)$, we look at $I(\mu^N(t)|\pi)$ where $\mu^N(t)$ is the population expectation of the empirical distribution $\rho^N(t)$ defined as 
\begin{align}\label{eq:average population density0}
    \mu^N(t,A)\coloneqq \mb{E}_{\underline{x}(0) \sim p_0^N} [ \rho^N(t,A) ] =\frac{1}{N}\sum_{i=1}^N\mb{P}_{\underline{x}(0) \sim p_0^N}[x^i(t) \in A], \ A \in \mathcal{B}(\mathbb{R}^d).
\end{align} 
When the initial condition $p_0^N$ admits a continuous differentiable density, $\mu^N(t)$ would admit a continuous differentiable density, hence making $I(\mu^N(t)|\pi)$ well-defined. Following \citep[Proposition 3]{he2024regularized}, to establish equivalence between the Fisher information and the regularized Fisher information requires a source-type regularity condition: for any $t\in [0,T]$, there exists $\gamma(t)\in (0,1/2]$ such that 
\begin{align*}
    \| \mc{J}(\mu^N(t),\pi)\|_{L_2^d(\mu^N(t))}<\infty ,\quad \text{for } \mc{J}(\mu^N(t),\pi) \coloneqq (\iota_{k,\mu^N(t)}^*\iota_{k,\mu^N(t)})^{-\gamma(t)}\nabla\log \frac{\mu^N(t)}{\pi}.
\end{align*}
Intuitively, this condition says that the $\nabla\log \frac{\mu^N(t)}{\pi}$ is sufficient regular relative to the kernel geometry: it should not concentrate too much in directions that are nearly invisible to the kernel operator. In addition, we assume such regularity along the trajectory i.e., $\inf_{t\in [0,T]}\Big(\tfrac{I(\mu^N(t)|\pi)}{2\| \mc{J}(\mu^N(t),\pi)\|_{L_2^d(\mu^N(t))}^2} \Big)^{\frac{1}{2\gamma(t)}}>0$. Equivalently, there exists a regularization parameter $\nu\in (0,1]$ such that
\begin{align}\label{eq:regularizer condition dynamics}
    \frac{\nu}{1-\nu}\le \inf_{t\in [0,T]}\left(\frac{I(\mu^N(t)|\pi)}{2\| \mc{J}(\mu^N(t),\pi)\|_{L_2^d(\mu^N(t))}^2} \right)^{\frac{1}{2\gamma(t)}}.
\end{align} 
Although \eqref{eq:regularizer condition dynamics} is stated abstractly, it can be checked on a case-by-case basis. See Appendix \ref{append:sufficient condition} for such an illustration.
When \eqref{eq:regularizer condition dynamics} holds, we have equivalence between $I(\mu^N(t)|\pi)$ and $I_{\nu,\text{Stein}}(\mu^N(t)|\pi)$  along the trajectory, i.e., for $t \in [0,T]$,
\begin{align}\label{eq:fisher equivalence}
    \frac{1}{2}(1-\nu)^{-1} I(\mu^N(t)|\pi)\le I_{\nu,\text{Stein}}(\mu^N(t)|\pi)\le (1-\nu)^{-1} I(\mu^N(t)|\pi).
\end{align}

Combining the equivalence relation in \eqref{eq:fisher equivalence} and the convergence results in Theorem \ref{thm:convergence}, we get convergence of the Fisher information along the R-SVGD dynamics.

\begin{theorem}\label{thm:fisher convergence dynamcis} Under the conditions in Theorem \ref{thm:convergence}, for any $T>0$, consider $(\mu^N(t))_{0\le t\le T}$ defined by \eqref{eq:average population density0}. Moreover, suppose \eqref{eq:regularizer condition dynamics} is satisfied. Then, we have
\begin{align*}
\frac{1}{T}\int_0^T I(\mu^N(t)|\pi) \dee t \lesssim \frac{\KL(p_0^N|\pi^{\otimes N})}{NT} + \frac{c^*d}{N} + \frac{\beta_1 d + \beta_2 \mb{E}[\mb{E}_{y\sim \rho^N(0)} [ \| \nabla V(y) \|^2 ]^{\frac{1}{2}}] d^{\frac{1}{2}}}{N} + \frac{\beta_2^2 C_V^2 d T^2} {\nu N^2 }.
\end{align*} 
Furthermore, defining $\mu_{av}^N\coloneqq \tfrac{1}{N^{2/3}}\int_0^{N^{2/3}} \mu^N(t)\dee t$ and assuming \eqref{eq:regularizer condition dynamics} with $T=N^{2/3}$, we have
\begin{align}\label{eq:average fisher convergence}
    I(\mu_{av}^N|\pi) \lesssim \frac{\KL(p_0^N|\pi^{\otimes N})}{N^{\frac{5}{3}}} + \frac{c^*d}{N} + \frac{\beta_1 d + \beta_2 \mb{E}[\mb{E}_{y\sim \rho^N(0)} [ \| \nabla V(y) \|^2 ]^{\frac{1}{2}}] d^{\frac{1}{2}}}{N} +\frac{\nu^{-1}\beta_2^2 C_V^2 d}{N^{\frac{2}{3}}}.
\end{align}
\end{theorem}

If $\pi$ satisfies the $\mathrm{W}_1\mathrm{I}$ with parameter $C_{\pi}$, we have $\mathrm{W}_1 (\mu_{av}^N,\pi)^2  \le C_\pi^2 I(\mu_{av}^N|\pi)$. Then immediately we obtain convergence of the time- and population- averaged empirical distribution $\mu_{av}^N$ in $\mathrm{W}_1$ along the R-SVGD dynamics from \eqref{eq:average fisher convergence} in Theorem \ref{thm:fisher convergence dynamcis}.

\begin{theorem}\label{thm:KL along RSVGF} Under conditions in Theorem \ref{thm:fisher convergence dynamcis}, if the target distribution $\pi$ satisfies $\mathrm{W}_1\mathrm{I}$ with parameter $C_{\pi}$, then for any $\nu$ satisfying \eqref{eq:regularizer condition dynamics} with $T=N^{2/3}$, we have with $\mu_{av}^N\coloneqq\tfrac{1}{N^{2/3}}\int_0^{N^{2/3}} \mu^N(t)\dee t$,
    \begin{align*}
        \mathrm{W}_1(\mu_{av}^N,\pi)^2 \lesssim C_\pi^2\bigg( \frac{\KL(p_0^N|\pi^{\otimes N})}{N^{\frac{5}{3}}} + \frac{c^*d}{N} + \frac{\beta_1 d + \beta_2 \mb{E}[\mb{E}_{y\sim \rho^N(0)} [ \| \nabla V(y) \|^2 ]^{\frac{1}{2}}] d^{\frac{1}{2}}}{N} +\frac{\nu^{-1}\beta_2^2 C_V^2 d}{N^{\frac{2}{3}}} \bigg).
    \end{align*}    
\end{theorem}

\begin{remark}\label{fisherconrem}
    Theorems \ref{thm:fisher convergence dynamcis} and \ref{thm:KL along RSVGF} highlight the main advantage of working with regularized SVGD. In contrast with the classical SVGD algorithm, we obtain convergence in (true) Fisher information for the annealed empirical measure $\mu^N$. In comparison, the rates in~\cite{balasubramanian2024improved}  are established only for the weaker $I_{\text{Stein}}$ metric. 
    
    Observe that the left hand side of \eqref{eq:regularizer condition dynamics} diverges as $\nu \rightarrow 1$. This indicates that $I_{\nu,\text{Stein}}(\mu^N(t)|\pi)$ and $I(\mu^N(t)|\pi)$ become more `singular' with respect to each other as $\nu \rightarrow 1$. Thus, although R-SVGD matches the performance of SVGD in this regime in $I_{\nu,\text{Stein}}(\mu^N(t)|\pi)$, it will have a possibly worse convergence rate with respect to the Fisher information and Wasserstein metric.

    In comparison to \cite[Theorem 5]{balasubramanian2024improved}, where dimension-dependent $W_2$ rates were obtained for a special (Mat\'ern) type of kernel for the true empirical measure, we obtain $W_1$ rates with dimension-independent exponents which are agnostic to the choice of the kernel, but for the annealed measure.
\end{remark}


\section{Convergence of R-SVGD Algorithm}\label{sec:rsvgdalgorithm}

\noindent In this section, we study the convergence of the actual R-SVGD algorithm \eqref{eq:algorithm}, in which the time-discretization with varying step-size and regularizers is introduced. We study the convergence by quantifying the decay rate of $\KL$ in every single step of the algorithm.

 In each step, we understand the R-SVGD algorithm as a transportation map on particles that push the empirical distribution of the particles towards the target distribution. For any $n\ge 1$, we define a transportation map $\Tm_{n}: \mb{R}^{dN}\to \mb{R}^{dN}$, as follows
\begin{align}\label{eq:algorithm map}
    {\Tm}_{n} (\underline{x})=\big( \frac{1-\nu_{n}}{N} K(\underline{x})+\nu_{n} I_N \big)^{-1}\big( \frac{1}{N} K(\underline{x}) {\nabla V}(\underline{x})-\frac{1}{N}\sum_{j=1}^N {\nabla_2 k}(\underline{x},x^j) \big), 
\end{align}
where $K(\underline{x})\coloneqq (k(x^i,x^j))_{i,j\in [N]}\in \mb{R}^{N\times N}$ is the Gram matrix and $\phi(\underline{x})\coloneqq (\phi(x_1),\cdots,\phi(x_N) )^\intercal$ for all $\phi$. Then the $n^{th}$ step of R-SVGD in \eqref{eq:algorithm} is exactly 
\begin{align}\label{eq:R-SVGD transportation map expression}
    \underline{X}_{n}= \Psi_n(h_{n}, \underline{X}_{n-1})\text{ where }\Psi_n(t,\underline{x})\coloneqq \underline{x}-t\Tm_{n}(\underline{x}), \quad \forall (t,\underline{x})\in [0,h_n]\times \mb{R}^{dN}.
\end{align}
Therefore, the single-step decaying property of $\KL$ along R-SVGD is equivalent to the decaying property $\KL$ after applying a sequence of maps $\{\Psi_n\}_{n\ge 0}$ in \eqref{eq:R-SVGD transportation map expression}.

Our strategy of analyzing the decay of $\KL$ after applying $\{\Psi_n\}_{n\ge 0}$ follows from \cite{korba2020non}. Denote ${p_n^N}=\text{Law}(\underline{X}_n)$ for all $n\ge 0$. We express $\{p_n^N\}_{n\ge 1}$ as a sequence of push-forward measures through $\{\Psi_{n}(h_{n},\cdot)\}_{n\ge 1}$ starting from $p_0^N$. For all $n\ge 0$ and $t\in [0,h_{n+1}]$, define the probability measure $\nu_{n,t}\coloneqq \Psi_{n+1}(t,\cdot)_{\#}p_n^N$, hence $\nu_{n,0}={p_n^N}$ and $\nu_{n,h_{n+1}}={p_{n+1}^N}$. Denote the density of $\nu_{n,t}$ by $q_{n,t}$. We study the decay of $\KL(q_{n,t}|\pi^{\otimes N})$ within the interval $[0,h_{n+1}]$ for a single iteration and then use telescoping property to analyze the decay property for multiple iterations.

Like many existing works studying SVGD~\citep{korba2020non, balasubramanian2024improved}, compared to continuous-time analysis, our discrete-time analysis requires extra (mild) assumptions on the kernel $k$ and target potential $V$.
\begin{assumption}\label{assump:convergence} We make the following assumptions; [(a)] Positivity: $\inf_{z\in \mb{R}^d} V(z)>0$, [(b)] Growth:  There exist $A>0$ and $\alpha\in [0,1/2]$ s.t. $\| \nabla V(x) \|\le AV(x)^\alpha$ for all $x\in \mb{R}^d$, and [(c)] Initial entropy bound: $\KL(p_0^N|\pi^{\otimes N})\le C_{\KL}Nd$ for some constant $C_{\KL}>0$.   
\end{assumption}
\begin{remark} Condition (b) in Assumption \ref{assump:convergence} is related to the tail-heaviness of the target density $\pi$. When the tails of $\pi$ decay slower than exponentially (including heavy tails), this corresponds to the choice $\alpha=0$, and $\alpha=1/2$ corresponds to $\pi$ having close to Gaussian tails. In our results, this tail decay rate will play a crucial role in tuning step sizes, regularization parameters and eventually in our convergence rates. 
    
\end{remark}
\begin{theorem}\label{thm:decay of regularized fisher svgd} Under Assumptions \ref{assump:kernel}, \ref{assump:target}, \ref{ass:cstarfin}, \ref{assump:convergence}, pick $\{(h_n, \nu_n)\}_{n=1}^T$ s.t. for all $n=1,2,\cdots, T$,
\begin{align}\label{eq:condition on step size and regularizer}
     \frac{h_{n}}{\nu_{n}}= \frac{\theta}{2}C_n^{-\frac{1}{2}} \le \frac{1}{16(1-\alpha)^2 C_V A^2 B^2 \sum_{l=1}^{n-1} h_l \nu_l^{-1}} \wedge \frac{N}{C_V B\nu_{n}+8Bd(1-\nu_{n})^2\nu_{n}^{-2}}   \wedge \frac{1}{B}
\end{align}
for some $\theta\in [0,1]$ and
\begin{align}\label{eq:choice of stepsize n+1}
   C_{n}=18 \max\{ \nu_{n}^{-4} A^2 B^4 N d M^{2\alpha} \big(d^{\frac{1}{1-\alpha}}+K\big)^{2\alpha}(\sum_{l=1}^{n-1} \frac{h_l}{\nu_l} \vee 1)^{\frac{2\alpha}{1-\alpha}}, \nu_{n}^{-4} B^4 N d^2 + B^2 C_V^2 d\}
\end{align} 
If the initial N-sample distribution $p_0^N$ satisfies $\int_{\mc{S}_K} {p^N_{0}}(\underline{z})\dee \underline{z}\ge 1/2 $ for some $K>0$ with
\begin{align}\label{eq:constraint initial density}
    {p^N_{0,K}}\coloneqq p_0^N|\mc{S}_K, \  \text{where } \ \mc{S}_K\coloneqq \left\{ \underline{x}\in \mb{R}^{dN}: N^{-1}\sum_{i=1}^N V(x^i)\le K \right\} ,
\end{align}
 and $\gamma\coloneqq 2C_{\KL}+\log 2$, then we have, writing $\rho^N_n := \frac{1}{N}\sum_{i=1}^N \delta_{X_n^i}$,   
\begin{align*}
 \mb{E}_{{p^N_{0,K}}} \left[\frac{1}{T} \sum_{n=1}^T h_n I_{\nu_{n},\text{Stein}}(\rho^N_{n-1}|\pi)  \right]&\le \frac{2\gamma d}{T} + \frac{2c^* d\sum_{n=1}^T h_n}{NT} + \frac{2 d \sum_{n=1}^T \beta_{1,n}(1-\nu_n) h_n}{NT} \\
 &\quad + \frac{\theta \sum_{n=1}^T \beta_{2,n}(1-\nu_n) \nu_{n}^3 }{B^2 N^{\frac{3}{2}} T } + \frac{\theta^2 \sum_{n=1}^T\nu_n^2 }{4NT},
\end{align*}
where $\beta_{1,n}$ and $\beta_{2,n}$ are given in \eqref{eq:constants alpha beta} with $\nu=\nu_n$.
\end{theorem}

To derive the convergence of Fisher information along the dynamics of R-SVGD, we follow the strategy in Section \ref{sec:FishWass}. First, instead of looking at $I(\rho_n^N|\pi)$ which is not well-defined, we look at $I({\mu^N_{n,K}}|\pi)$ where ${\mu^N_{n,K}}$ is defined as 
\begin{align}\label{eq:average population density}
    \mu^N_{n,K}\coloneqq \mb{E}_{\underline{X}_0\sim {p^N_{0,K}}} [ \rho_n^N ] =\frac{1}{N}\mb{E}_{\underline{X}_0\sim {p^N_{0,K}}} [\sum_{i=1}^N \delta_{X_n^i}].
\end{align} 
When the initial condition ${p^N_{0,K}}$ admits a continuous differentiable density, ${\mu^N_{n,K}}$ will also admit a continuous differentiable density. Hence, $I({\mu^N_{n,K}}|\pi)$ is well-defined. According to \cite[Proposition 3]{he2024regularized}, for any $n=0,1,\cdots, T$, assume that there exists $\gamma_n\in (0,1/2]$ such that 
\begin{align}\label{eq:L2 embedding}
     R_n\coloneqq \| \mc{J}({\mu^N_{n,K}},\pi)\|_{L_2^d({\mu^N_{n,K}})}<\infty ,\quad \text{for } \mc{J}({\mu^N_{n,K}},\pi) \coloneqq (\iota_{k,{\mu^N_{n,K}}}^*\iota_{k,{\mu^N_{n,K}}})^{-\gamma_n}\nabla\log \frac{{\mu^N_{n,K}}}{\pi}.
\end{align}
Then we can derive the convergence of the Fisher information from the convergence of $I_{\nu_n,\text{Stein}}$ in Theorem \ref{thm:decay of regularized fisher svgd} and the equivalence between Fisher information and regularized Stein Fisher information. We state the decay of Fisher information along the R-SVGD dynamics, and in $W_1$ under $\mathrm{W}_1\mathrm{I}$, in the following theorem.
\begin{theorem}\label{thm:decay of fisher} Under conditions in Theorem \ref{thm:decay of regularized fisher svgd}, for any $K>0$, suppose there exists $\{\gamma_n\}_{n=1}^T$ such that $R_n<\infty$ (defined in \eqref{eq:L2 embedding}) for all $n=1,2,\cdots, T$. Then if we pick $\{\nu_n\}_{n=1}^T$ small enough such that $\tfrac{\nu_n}{1-\nu_n}\le \left( \tfrac{I({\mu^N_{n-1,K}}|\pi)}{2R_{n-1}^2} \right)^{\frac{1}{2\gamma_{n-1}}}$ for all $n=1,2,\cdots,T$, we have   
\begin{align*}
\frac{1}{T}\sum_{n=1}^T \frac{h_n}{1-\nu_n} I({\mu^N_{n-1,K}} | \pi) &\le \frac{4\gamma d }{T} + \frac{4c^* d\sum_{n=1}^T h_n}{NT} +\frac{4 d \sum_{n=1}^T \beta_{1,n}(1-\nu_n) h_n}{NT} \\
&\quad + \frac{2\theta \sum_{n=1}^T \beta_{2,n}(1-\nu_n) \nu_{n}^3 }{B^2 N^{\frac{3}{2}} T } + \frac{\theta^2 \sum_{n=1}^T\nu_n^2 }{2NT} ,
\end{align*}
where $\gamma\coloneqq 2C_{\KL}+\log 2$ and $\beta_{1,n},\beta_{2,n}$ are given in \eqref{eq:constants alpha beta} with $\nu=\nu_n$. Furthermore, if the target distribution $\pi$ satisfies $\mathrm{W}_1\mathrm{I}$ with parameter $C_{\pi}$, then we have
\begin{align*}
    W_1(\mu^N_{T,av},\pi)^2 &\le \frac{4C_\pi^2\gamma d }{T} + \frac{4C_\pi^2c^* d\sum_{n=1}^T h_n}{NT} + \frac{4 C_\pi^2 d \sum_{n=1}^T \beta_{1,n}(1-\nu_n) h_n}{NT} \\
    &\quad + \frac{2C_\pi^2 \theta \sum_{n=1}^T \beta_{2,n}(1-\nu_n) \nu_{n}^3 }{B^2 N^{\frac{3}{2}} T } + \frac{C_\pi^2\theta^2 \sum_{n=1}^T\nu_n^2 }{2NT} ,
\end{align*}
where $\mu^N_{T,av}\coloneqq \tfrac{1}{T}\sum_{n=1}^T \frac{h_n}{1-\nu_n} {\mu^N_{n-1,K}}$.
\end{theorem}
Next, we investigate the optimal scaling in the bounds obtained in Theorems \ref{thm:decay of regularized fisher svgd} and \ref{thm:decay of fisher} when $h_n,\nu_n$ are constants, with proof deferred to Appendix \ref{append:optimal scaling}.

\begin{corollary}[Rates under constant $h,\nu$]\label{cor:optimal discussion} The optimal scalings in Theorems \ref{thm:decay of regularized fisher svgd} and \ref{thm:decay of fisher} under constant $h,\nu$ are as follows: 
\begin{itemize}
    \item [(1)] Picking $\nu\equiv 1-\tfrac{1}{N}$, $h= \Theta\big(\big(d^{\frac{1+\alpha}{2(1-\alpha)}} + d + K^{\alpha}\big)^{-(1-\alpha)} N^{-\frac{1+\alpha}{1-\alpha}}\big)$, $T=N^{\frac{2}{1-\alpha}}$ and $\theta=\sqrt{N/T}=N^{-\frac{1+\alpha}{2(1-\alpha)}}$, conditions in Theorem \ref{thm:decay of regularized fisher svgd} are satisfied, and we get
    \begin{align*}
     \mb{E}_{{p^N_{0,K}}} \left[\frac{1}{T} \sum_{n=1}^T  I_{\nu,\text{Stein}}(\rho^N_{n-1}|\pi)  \right] \lesssim \frac{d(d^{\frac{1+\alpha}{2}} + d^{1-\alpha} + K^{\alpha(1-\alpha)}d^{\frac{1-\alpha}{2}}) }{N}.
\end{align*}
   \item [(2)] Assume there exists $c\in [0,\tfrac{1-\alpha}{3-2\alpha})$\footnote{When $c=\tfrac{1-\alpha}{3-2\alpha}$, the estimation bound is $\mc{O}(1)$ in $N$. When $c>\tfrac{1-\alpha}{3-2\alpha}$, the estimation bound tends to infinity as $N\to\infty$.} such that $\nu_n\equiv \nu={\Theta}(N^{-c})$ satisfies $\tfrac{\nu}{1-\nu}\le \Big( \tfrac{I({\mu^N_{n-1,K}}|\pi)}{2R_{n-1}^2} \Big)^{\frac{1}{2\gamma_{n-1}}}$ for $n \le T$. Picking $h = \Theta \big(d^{-\frac{1-\alpha}{2}}(d^{\frac{1}{1-\alpha}}+K)^{-\alpha(1-\alpha)} N^{1-\alpha-c(3-2\alpha)-\frac{3}{2}-\max\{ 0, \frac{1}{2}-2c \}}\big)$, $T=N^{\frac{3}{2} + \max\{ 0, \frac{1}{2}-2c \} }$, $\theta=N^{\frac{3}{2}}/T= N^{-\max\{ 0, \frac{1}{2}-2c \}}$, we have
\begin{align*}
    \frac{1}{T}\sum_{n=1}^T  I({\mu^N_{n,K}} | \pi), W_1(\mu^N_{T,av},\pi)^2 \lesssim \frac{d(d^{\frac{1+\alpha}{2}} + d^{1-\alpha} + K^{\alpha(1-\alpha)}d^{\frac{1-\alpha}{2}}) }{N^{1-\alpha-c(3-2\alpha)}},
\end{align*}
\end{itemize}
where the notation $\Theta$ and $\lesssim$ hide parameters except for $N,d$.
\end{corollary}

\begin{remark}
Corollary~\ref{cor:optimal discussion} summarizes how to tune the regularization $\nu$, step size $h$, and time horizon $T$ to obtain explicit finite-$N$ accuracy guarantees. In regime~(1), taking $\nu\equiv 1-\frac{1}{N}$ keeps the method close to SVGD (strong kernelization) while still controlling the regularized Stein Fisher information; with the prescribed $h$ and $T$, the time-averaged discrepancy $\frac{1}{T}\sum_{n=1}^T I_{\nu,\mathrm{Stein}}(\rho^N_{n-1}\mid\pi)$ decays at the canonical $N^{-1}$ Monte Carlo rate up to a dimension-/kernel-dependent factor $d\big(d^{\frac{1+\alpha}{2}}+d^{1-\alpha}+K^{\alpha(1-\alpha)}d^{\frac{1-\alpha}{2}}\big)$. In regime~(2), $\nu$ is allowed to decrease with $N$ (i.e., progressively ``de-kernelizing'' the dynamics toward the Wasserstein gradient flow), and the corollary quantifies the resulting trade-off: smaller $\nu$ improves approximation to the WGF but introduces an estimation cost that slows the rate from $N^{-1}$ to $N^{-(1-\alpha-c(3-2\alpha))}$ for both the Fisher information and the squared $W_1$ error of the averaged iterate $\mu^N_{T,av}$. The condition $c<\tfrac{1-\alpha}{3-2\alpha}$ ensures this estimation term remains vanishing as $N\to\infty$, whereas $\nu$ that decays too quickly would prevent the finite-particle bound from converging to zero.
\end{remark}

\newpage


\bibliographystyle{plainnat}
\bibliography{cite}

\newpage 

\appendix
\section{Additional Related Works}

Rates of convergence to equilibrium in under various assumption for the WGF is relatively well-studied~\citep{ambrosio2005gradient}. Recently, several works established similar rates for the SVGF~\citep{liu2017stein, lu2019scaling,chewi2020svgd,duncan2019geometry,carrillo2024stein} and~R-SVGF~\citep{he2024regularized}. Convergence rates for time-discretizations of SVGF have also been studied by various authors including~\cite{korba2020non}, \cite{ salim2022convergence}, and \cite{sun2023convergence}.

\section{Notations}

\begin{table}[h!]
\centering
\begin{tabular}{lll}
\toprule
\textbf{Symbol} & \textbf{Meaning} & \textbf{Underlying Space} \\
\midrule
$X_n^i$ & the $i^{th}$ particle at iteration $n$ of R-SVGD & $\mathbb{R}^d$ \\
$\underline{X}_n$ & all particle at iteration $n$ of R-SVGD & $\mathbb{R}^{dN}$ \\
$x^i(t)$ & the $i^{th}$ particle at time $t$ of the continuous-time R-SVGD & $\mb{R}^d$ \\ 
$\underline{x}(t)$ & all particles at time $t$ of the continuous-time R-SVGD & $\mb{R}^d$ \\ 
$\rho^N(t)$ & empirical distribution at time $t$ along continuous-time R-SVGD & $\mathcal{P}(\mb{R}^d)$\\
$p^N(t)$ & joint distribution of all particles at time $t$ along continuous-time R-SVGD & $\mathcal{P}(\mb{R}^{dN})$\\
$\mu^N(t)$ & population expectation of the empirical distribution $\rho^N(t)$ & $\mc{P}(\mb{R}^d)$ \\
$\rho^N_n$ & empirical distribution at iteration $n$ in R-SVGD & $\mathcal{P}(\mb{R}^d)$\\
$p^N_n$ & joint distribution of all particles at iteration $n$ in R-SVGD & $\mathcal{P}(\mb{R}^{dN})$\\
$\mu^N_n$ & population expectation of the empirical distribution $\rho^N_n$ & $\mc{P}(\mb{R}^d)$ \\
\bottomrule
\end{tabular}
\end{table}

\section{Preliminaries on Regularized Stein Variational Gradient Descent}\label{sec:rsvgdprelim}

We now provide a brief description of the derivation of the \emph{Regularized Stein Variational Gradient Flow} (R-SVGF) by~\cite{he2024regularized}, and clarify how it connects to both the Stein Variational Gradient Flow (SVGF) and the Wasserstein Gradient Flow (WGF). 
The derivation proceeds from the standard variational characterization of steepest descent of $\mathrm{KL}(\rho|\pi)$ under infinitesimal transport maps. Consider a perturbation of the form $T(x)=x+h\phi(x)$, where $h>0$ is small and $\phi:\mathbb{R}^d\to\mathbb{R}^d$ is a smooth vector field. The key object is the \emph{Stein operator} associated with a density $p$, $\mathcal{A}_p\phi(x)=\phi(x)\otimes \nabla\log p(x)+\nabla\phi(x),
$ whose trace controls the first variation of the KL divergence: for $\rho\in\mathcal{P}(\mathbb{R}^d)$,~\cite{liu2016stein} showed that $\left.\nabla_h\,\mathrm{KL}(T_\#\rho\mid\pi)\right|_{h=0}
=-\mathbb{E}_{x\sim\rho}\!\left[\mathrm{trace}\big(\mathcal{A}_\pi\phi(x)\big)\right]$. Thus, choosing $\phi$ to maximize $\big(\mathbb{E}_{\rho}[\mathrm{trace}(\mathcal{A}_\pi\phi)]\big)^2$ over a unit ball yields the direction of steepest KL descent. If the unit ball is taken in $L_2^d(\rho)$, then an integration-by-parts argument identifies the maximizer with the Wasserstein gradient $\nabla\log(\rho/\pi)$, recovering the WGF. SVGD instead restricts $\phi$ to the unit ball of the vector-valued RKHS $\mathcal{H}_k^d$, which yields the kernelized direction $\mathcal{T}_{k,\rho}\nabla\log(\rho/\pi)$ and hence the SVGF. 

\cite{he2024regularized} chose to use a regularization that interpolates between these two geometries by maximizing the same Stein objective over an \emph{interpolated norm}: $\nu\|\phi\|_{\mathcal{H}_k^d}^2+(1-\nu)\|\phi\|_{L_2^d(\rho)}^2\le 1, \nu\in(0,1]$. This constraint trades off RKHS regularity and $L_2(\rho)$ fidelity, and leads to a closed-form optimal vector field expressed via the RKHS inclusion operator $\iota_{k,\rho}:\mathcal{H}_k^d\to L_2^d(\rho)$ and its adjoint $\iota_{k,\rho}^*$. Under mild assumptions (finite Fisher information and $\int k(x,x)\rho(x)\,dx<\infty$), the maximizing direction can be written equivalently as
\[
\phi_{\rho,\pi}^*
\ \propto\
-\Big((1-\nu)\,\iota_{k,\rho}^*\iota_{k,\rho}+\nu I_d\Big)^{-1}\,
\iota_{k,\rho}^*\!\left(\nabla\log\frac{\rho}{\pi}\right),
\]
which reduces to SVGD as $\nu\to 1$ and approaches the Wasserstein gradient as $\nu\downarrow 0$ (in the regimes quantified later). Plugging $\phi_{\rho_t,\pi}^*$ into the continuity equation yields the mean-field R-SVGF PDE:
\[
\partial_t \rho_t
=
\nabla\cdot\!\left(
\rho_t\,
\iota_{k,\rho_t}
\Big((1-\nu)\,\iota_{k,\rho_t}^*\iota_{k,\rho_t}+\nu I_d\Big)^{-1}
\iota_{k,\rho_t}^*\!\left(\nabla\log\frac{\rho_t}{\pi}\right)
\right).
\]
Finally, the operator appearing above admits an important \emph{alternative form}, making the connection to SVGF explicit: using $\iota_{k,\rho}\Big((1-\nu)\iota_{k,\rho}^*\iota_{k,\rho}+\nu I_d\Big)^{-1}\iota_{k,\rho}^*
=
\Big((1-\nu)\iota_{k,\rho}\iota_{k,\rho}^*+\nu I\Big)^{-1}\iota_{k,\rho}\iota_{k,\rho}^*,$ and recalling $\mathcal{T}_{k,\rho}=\iota_{k,\rho}\iota_{k,\rho}^*$, we obtain the compact expression introduced in~\eqref{eq:intro_RSVGF}. This equivalent formulation highlights R-SVGF as a resolvent-corrected version of SVGF: the flow applies $\big((1-\nu)\mathcal{T}_{k,\rho}+\nu I\big)^{-1}$ to ``undo'' the non-invertibility and constant-order bias induced by $\mathcal{T}_{k,\rho}$, while preserving the kernel-based structure that enables deterministic particle implementations.

\subsection{Particle-based spatial discretization (R-SVGD dynamics).}\label{sec:particledynamic}
To obtain an implementable algorithm, we discretize the mean-field R-SVGF by restricting the evolving law $\rho_t$ to the class of empirical measures of finitely many particles. Concretely, for $N\in\mathbb{N}$ we consider $\rho_t^N \;\coloneqq\; \frac{1}{N}\sum_{j=1}^N \delta_{x_j(t)},$ where $\{x_i(t)\}_{i=1}^N\subset\mathbb{R}^d$ are deterministic particles. This choice can be viewed as a spatial discretization of the continuity equation associated with the R-SVGF: the velocity field is obtained by substituting $\rho_t$ with $\rho_t^N$ in the \emph{optimal} steepest-descent direction derived from the regularized Stein variational problem. The key idea is to choose the transport direction that most rapidly decreases $\mathrm{KL}(\rho\mid\pi)$ while balancing smoothness (through the RKHS norm) and fidelity to the current distribution (through the $L_2(\rho)$ norm), which leads to a resolvent-preconditioned Stein update that admits equivalent formulations in terms of either RKHS embedding operators or the associated kernel integral operator; see \cite[Section 3]{he2024regularized}. As a result, each particle moves according to the same regularized Stein vector field evaluated at its current location, leading to the interacting ODE system
\vspace{-.1cm}
{\footnotesize
\begin{align*}
\left\{
    \begin{aligned}
    \dot{x}^i(t)&=-\left((1-\nu)\iota_{k,\rho^N(t)}^*\iota_{k,\rho^N(t)}+\nu I_d \right)^{-1}\bigg(\frac{1}{N}\sum_{j=1}^N -\nabla_2 k\left(x^i(t),x^j(t)\right)+k\left(x^i(t),x^j(t)\right)\nabla V(x^j(t)) \bigg)\\
     x^i(0)&=x^i_0\in \mb{R}^d,\quad i=1,2,\ldots,N
    \end{aligned}
    \right.,
\end{align*}
}
\noindent where $\nu \in (0,1]$, $\{x^i(t)\}_{i=1}^N$ is the set of $N$ particles, 
$\nabla_2 k(\cdot,\cdot)$ denotes the gradient of $k$ with respect to its second argument, and $\iota_{k,\rho_t^N}:\mathcal{H}_k^d\to L_2^d(\rho_t^N)$ is the RKHS embedding induced by the empirical measure (with adjoint $\iota_{k,\rho_t^N}^*$). The term inside the parentheses is the familiar SVGD-type ``Stein force'': it combines an \emph{attractive} drift term (through $\nabla V$) that pushes particles toward regions of high target density, and a \emph{repulsive} interaction term (through $\nabla_2 k$) that discourages particle collapse. The distinctive feature of R-SVGD is the preconditioning by the regularized inverse $\big((1-\nu)\iota_{k,\rho_t^N}^*\iota_{k,\rho_t^N}+\nu I_d\big)^{-1}$, which corrects the non-invertibility of the kernel operator and mitigates the constant-order bias present in SVGF. In particular, as $\nu\to 1$ the preconditioner reduces to $I_d$ and we recover the usual SVGD particle dynamics, whereas as $\nu\downarrow 0$ the dynamics increasingly ``de-kernelizes'' the gradient direction, bringing the particle flow closer to the Wasserstein gradient flow in the regimes quantified later.

\section{Proofs for Section~\ref{sec:rsvgddynamics}}\label{sec:proof thm 1}
\namedproof{Proof of Theorem \ref{thm:convergence}.}
    The $N$-body Liouville equation for the evolution of particle joint density $p^N(\cdot)$ is given by
\textcolor{black}{
\begin{align}\label{eq:liouville}
    \partial_t p^N(t)(\underline{x})+\frac{1}{N}\sum_{k,l=1}^N \text{div}_{{x^k}} \bigg( p^N(t)(\underline{x}) \Phi_{l}(\underline{x},{x^k}) \bigg) =0,
\end{align}
with $\Phi_{l}(\underline{x};\cdot)=\big( (1-\nu)\iota_{k,\rho^N}^*\iota_{k,\rho^N}+\nu I_d \big)^{-1}\big( \nabla_2 k(\cdot,x^l)-k(\cdot,x^l)\nabla V(x^l) \big)$.
} According to \eqref{eq:liouville}, we can compute the derivative of $\KL(p^N(t)|\pi^{\otimes N})$:
\begin{align*}
    \frac{\dee }{\dee t} \KL(p^N(t)|\pi^{\otimes N}) & = \int_{\mb{R}^{dN}} \log \frac{p^N(t)(\underline{x})}{\pi^{N}(\underline{x})} \partial_t p^N(t)(\underline{x}) + \partial_t p^N(t)(\underline{x}) \dee \underline{x} \\
    & = -\frac{1}{N} \sum_{k,l=1}^N \int_{\mb{R}^{dN}} \log \frac{p^N(t)(\underline{x})}{\pi^{N}(\underline{x})} \text{div}_{{x^k}} \bigg( p^N(t)(\underline{x}) \Phi_{l}(\underline{x};{x^k}) \bigg) \dee \underline{x} \\
    &=\frac{1}{N}\sum_{k,l=1}^N \int_{\mb{R}^{dN}}\nabla_{{x^k}}p^N(t)(\underline{x})\cdot \Phi_{l}(\underline{x};{x^k}) +\nabla V({x^k})\cdot \Phi_{l}(\underline{x};{x^k}) p^N(t)(\underline{x})\dee \underline{x}\\
     &=\frac{1}{N}\sum_{k,l=1}^N \int_{\mb{R}^{dN}} \bigg( -\text{div}_{{x^k}}\Phi_{l}(\underline{x};{x^k}) +\nabla V({x^k})\cdot \Phi_{l}(\underline{x};{x^k}) \bigg) p^N(t)(\underline{x})\dee \underline{x}.
\end{align*}
Next, we look at the terms of $\Phi_{l}$. According to the definition, we have for any $l\in [N]$,
\begin{align*}
    &\big( (1-\nu)\iota_{k,\rho^N}^*\iota_{k,\rho^N}+\nu I_d \big)\Phi_{l}(\underline{x};\cdot) = \nabla_2 k(\cdot,x^l)-k(\cdot,x^l)\nabla V(x^l) ,
\end{align*}
which implies that for any $k,l\in [N]$,
\begin{align*}
   (1-\nu)\frac{1}{N}\sum_{i=1}^N k({x^k},x^i)\Phi_{l}(\underline{x};x^i) + \nu \Phi_l(\underline{x};{x^k}) = \nabla_2 k({x^k},x^l)-k({x^k},x^l)\nabla V(x^l).
\end{align*}
If we further denote $\phi(\underline{x};\cdot):=\sum_{l=1}^N \Phi_l(\underline{x};\cdot)$ and $\underline{\phi}\coloneqq (\phi(\underline{x};x_1),\cdots, \phi(\underline{x};x_N))^\intercal$, then we have
\begin{align}\label{eq:matrix vector equation for the inverse operator}
    \big(\frac{1-\nu}{N}K + \nu I_{N} \big) \underline{\phi} = \Psi
\end{align}
where $\Psi\in \mb{R}^{N\times d}$ such that $\Psi_{k,\cdot}=\left(\sum_{l=1}^N \nabla_2 k({x^k},x^l)-k({x^k},x^l)\nabla V(x^l)\right)^\intercal$. Since \textcolor{black}{the kernel $k$ is positive semi-definite}, the matrix $K$ is positive semi-definite and $\underline{\phi}$ is uniquely defined as $\underline{\phi}=  \big(\frac{1-\nu}{N}K + \nu I_{N} \big)^{-1} \Psi$. As a consequence, we have
\begin{align}\label{eq:KL derivative}
     \frac{\dee }{\dee t} \KL(p^N(t)|\pi^{\otimes N}) & = \frac{1}{N}\sum_{k=1}^N \mb{E}_{\underline{x}\sim p^N(t)} \big[ -\text{div}_{{x^k}}\phi(\underline{x};{x^k})+\nabla V({x^k})\cdot \phi(\underline{x};{x^k}) \big],
\end{align}
where $\phi$ satisfies \eqref{eq:matrix vector equation for the inverse operator}.

\noindent On the other hand, the regularized Stein Fisher information is given by
\begin{align*}
    &\quad I_{\nu,\text{Stein}}(\mu|\pi) \\
    &= \langle \int_{\mb{R}^d} -\nabla_2 k(\cdot,y)+k(\cdot,y)\nabla V(y) \mu(y)\dee y, \big( (1-\nu)\iota_{k,\mu}^*\iota_{k,\mu}+\nu I \big)^{-1} \\
    &\qquad \int_{\mb{R}^d} -\nabla_2 k(\cdot,y)+k(\cdot,y)\nabla V(y) \mu(y)\dee y  \rangle_{\mc{H}_k^d}. 
\end{align*}
Therefore, when $\mu=\rho^N\coloneqq \frac{1}{N}\sum_{i=1}^N \delta_{x^i} $, we have
\begin{align*}
     &\quad I_{\nu,\text{Stein}}(\rho^N|\pi) \\
     &= \langle \frac{1}{N}\sum_{i=1}^N -\nabla_2 k(\cdot,x^i)+k(\cdot,x^i)\nabla V(x^i),  \big( (1-\nu)\iota_{k,\rho^N}^*\iota_{k,\rho^N}+\nu I \big)^{-1} \\
     &\qquad \big(\frac{1}{N}\sum_{i=1}^N -\nabla_2 k(\cdot,x^i)+k(\cdot,x^i)\nabla V(x^i)\big) \rangle_{\mc{H}_k^d}.
\end{align*}
If we denote $\varphi(\underline{x};\cdot)\coloneqq \big( (1-\nu)\iota_{k,\rho^N}^*\iota_{k,\rho^N}+\nu I \big)^{-1} \big(\frac{1}{N}\sum_{i=1}^N -\nabla_2 k(\cdot,x^i)+k(\cdot,x^i)\nabla V(x^i)\big)  $ and $\underline{\varphi}\coloneqq (\varphi(\underline{x};x_1),\cdots,\varphi(\underline{x};x_N))^\intercal$, then for all $i\in [N]$,
\begin{align*}
    &(1-\nu)\frac{1}{N}\sum_{j=1}^N k(x^i,x^j)\varphi(\underline{x};x^j)+\nu \varphi(\underline{x};x^i)= \frac{1}{N}\sum_{j=1}^N -\nabla_2 k(x^i,x^j)+k(x^i,x^j)\nabla V(x^j)\\
    \implies & N\big( \frac{1-\nu}{N}K+\nu I_N \big) \underline{\varphi} = -\Psi.
\end{align*}
According to \eqref{eq:matrix vector equation for the inverse operator} and the uniqueness of $\underline{\phi}$, we know that $\underline{\phi}=-N\underline{\varphi}$. Therefore, writing $\nabla \cdot \varphi(\underline{x};x^i)$ to denote the divergence of the map $z \mapsto \varphi(\underline{x};z)$ at $z=x_i$, we have
\begin{align}
    I_{\nu,\text{Stein}}(\rho^N|\pi)& = \langle \frac{1}{N}\sum_{i=1}^N -\nabla_2 k(\cdot,x^i)+k(\cdot,x^i)\nabla V(x^i), \varphi(\underline{x};\cdot) \rangle_{\mc{H}_k^d} \nonumber \\
    &= \frac{1}{N}\sum_{i=1}^N -\nabla \cdot \varphi(\underline{x};x^i)+\nabla V(x^i)\cdot \varphi(\underline{x};x^i) \nonumber\\
    &= -\frac{1}{N^2}\sum_{i=1}^N -\nabla \cdot \phi(\underline{x};x^i)+\nabla V(x^i)\cdot \phi(\underline{x};x^i). \label{eq:regularized Stein fisher}
\end{align}
Combine \eqref{eq:KL derivative} and \eqref{eq:regularized Stein fisher}, we have
\begin{align*}
    \frac{\dee }{\dee t} \KL(p^N(t)|\pi^{\otimes N}) & = -N\mb{E}_{\underline{x}\sim p^N(t)} \big[ I_{\nu,\text{Stein}}(\rho^N|\pi) \big] + \mb{E}\big[ \frac{1}{N}\sum_{i=1}^N \nabla\cdot \phi(\underline{x};x^i)-\text{div}_{x^i}\phi(\underline{x};x^i) \big].
\end{align*}
Note that the last term is not zero as $\nabla\cdot$ is applied to $\phi(\underline{x};\cdot)$ but $\text{div}_{x^i}$ is applied to $\phi(\underline{x};x^i)$ where the first variable $\underline{x}$ also depends on $x^i$. We denote $$C^*(\underline{x})\coloneqq \frac{1}{N}\sum_{i=1}^N \nabla\cdot \phi(\underline{x};x^i)-\text{div}_{x^i}\phi(\underline{x};x^i) = \frac{1}{N}\sum_{i=1}^NC^*_i(\underline{x}),$$ where $C^*_i(\underline{x}) = \nabla\cdot \phi(\underline{x};x^i)-\text{div}_{x^i}\phi(\underline{x};x^i)$. Then after integrating both sides and taking the average on $[0,T]$, we get
\begin{align*}
   \frac{\KL(p^N(T)|\pi^{\otimes N})-\KL(p_0^N|\pi^{\otimes N})}{T}=-\frac{N}{T} \int_0^T \mb{E}_{\underline{x}\sim p^N(t)} \big[ I_{\nu,\text{Stein}}(\rho^N|\pi) \big]\dee t +\frac{1}{T}\int_0^T \mb{E}_{\underline{x}\sim p^N(t)}[C^*(\underline{x})] \dee t.
\end{align*}
Equation \eqref{eq:convergence of regularized Stein fisher} and Equation \eqref{eq:KL decay} then follow from the non-negativity of $\KL$ and $I_{\nu,\text{Stein}}$ respectively. 

\noindent Next, we derive the explicit formula for $C^*(\underline{x})$ by characterizing the difference between $\nabla\cdot \phi(\underline{x};x^i)$ and $\text{div}_{x^i}\phi(\underline{x};x^i)$. According to the definition of $\phi$, we have
    \begin{align*}
    &(1-\nu)\frac{1}{N}\sum_{j=1}^N k(\cdot,x^j)\phi(\underline{x};x^j)+\nu \phi(\underline{x};\cdot) = \sum_{j=1}^N \nabla_2 k(\cdot,x^j)-k(\cdot,x^j)\nabla V(x^j) \\
    \implies & \frac{1-\nu}{N}\sum_{j=1}^N \nabla_1 k(\cdot,x^j)\cdot\phi(\underline{x};x^j)+\nu \nabla\cdot\phi(\underline{x};\cdot)= \sum_{j=1}^N \nabla_1\cdot\nabla_2 k(\cdot,x^j)-\nabla_1 k(\cdot,x^j)\cdot\nabla V(x^j)\\
    \implies & \nabla\cdot \phi(\underline{x};x^i)=\frac{1}{\nu}\sum_{j=1}^N \nabla_1\cdot\nabla_2 k(x^i,x^j)-\nabla_1 k(x^i,x^j)\cdot\nabla V(x^j)-\frac{1-\nu}{N}\nabla_1 k(x^i,x^j)\cdot\phi(\underline{x};x^j).
\end{align*}
On the other hand, if we denote $\underline{\text{div}}\phi=\big( \text{div}_{{x^k}}\phi(\underline{x};x^j)\big)_{j,k\in [N]} $, we have for any $k \in [N]$,
\small{
    \begin{align*}
    &(1-\nu)\frac{1}{N}\sum_{j=1}^N k(x^i,x^j)\phi(\underline{x};x^j)+\nu \phi(\underline{x};x^i)= \sum_{j=1}^N \nabla_2 k(x^i,x^j)-k(x^i,x^j)\nabla V(x^j) \\
    \implies & \frac{1-\nu}{N}\sum_{j=1}^N \left(\nabla_1 k(x^i,x^j)\cdot\phi(\underline{x};x^j)1_{k=i}+k(x^i,x^j)\text{div}_{{x^k}}\phi(\underline{x};x^j)\right)  + \frac{1-\nu}{N}\nabla_2 k(x^i,{x^k})\cdot\phi(\underline{x};{x^k}) +\nu \text{div}_{{x^k}}\phi(\underline{x};x^i)\\
    = & 1_{k=i}\sum_{j=1}^N \left(\nabla_1\cdot\nabla_2 k(x^i,x^j)-\nabla_1 k(x^i,x^j)\cdot\nabla V(x^j)\right) +\Delta_2 k(x^i,{x^k})-\nabla_2 k(x^i,{x^k})\cdot\nabla V({x^k}) -k(x^i,{x^k})\Delta V({x^k}) \\
    \implies & \underline{\text{div}}\phi= \big( \frac{1-\nu}{N} K + \nu I \big)^{-1}\Lambda ,
\end{align*}
}
where $\Lambda\in \mb{R}^{N\times N}$ and 
\begin{align*}
    \Lambda_{i,k}&= \Delta_2 k(x^i,{x^k})-\nabla_2 k(x^i,{x^k})\cdot\nabla V({x^k}) -k(x^i,{x^k})\Delta V({x^k}) -\frac{1-\nu}{N}\nabla_2 k(x^i,{x^k})\cdot\phi(\underline{x};{x^k})  \\
    &\quad +  1_{k=i} \sum_{j=1}^N \left(\nabla_1\cdot\nabla_2 k(x^i,x^j)-\nabla_1 k(x^i,x^j)\cdot\nabla V(x^j)-\frac{1-\nu}{N}\nabla_1 k(x^i,x^j)\cdot\phi(\underline{x};x^j)\right).
\end{align*}

Therefore, we get
\begin{align*}
     \nabla\cdot \phi(\underline{x};x^i)&=\frac{1}{\nu}\sum_{j=1}^N \nabla_1\cdot\nabla_2 k(x^i,x^j)-\nabla_1 k(x^i,x^j)\cdot\nabla V(x^j)-\frac{1-\nu}{N}\nabla_1 k(x^i,x^j)\cdot\phi(\underline{x};x^j) \\
     \text{div}_{x^i}\phi(\underline{x};x^i)&=(\underline{\text{div}}\phi)_{i,i}= \bigg(\big( \frac{1-\nu}{N} K + \nu I \big)^{-1}\Lambda \bigg)_{i,i}.    
\end{align*}
\vspace{.2cm}
For simplicity, we denote $\Knu\coloneqq\big( \frac{1-\nu}{N} K + \nu I \big)^{-1}$. Then we can express $C_i^*(\underline{x})$ as follows,
\begin{align*}
    &\quad C_i^*(\underline{x}) = \nabla\cdot \phi(\underline{x};x^i)-\text{div}_{x^i}\phi(\underline{x};x^i)\\
    =& \frac{1}{\nu}\sum_{j=1}^N \nabla_1\cdot\nabla_2 k(x^i,x^j)-\nabla_1 k(x^i,x^j)\cdot\nabla V(x^j)-\frac{1-\nu}{N}\nabla_1 k(x^i,x^j)\cdot\phi(\underline{x};x^j)\\
    &- \sum_{j=1}^N \Knu_{ij}\big( \Delta_2 k(x^j,x^i)-\nabla_2 k(x^j,x^i)\cdot\nabla V(x^i) -k(x^j,x^i)\Delta V(x^i) -\frac{1-\nu}{N}\nabla_2 k(x^j,x^i)\cdot\phi(\underline{x};x^i) \big) \\
    &-\sum_{j=1}^N \Knu_{ii}\big(\nabla_1\cdot\nabla_2 k(x^i,x^j)-\nabla_1 k(x^i,x^j)\cdot\nabla V(x^j)-\frac{1-\nu}{N}\nabla_1 k(x^i,x^j)\cdot\phi(\underline{x};x^j)\big) \\
    =&\sum_{j=1}^N (\nu^{-1}I-\Knu)_{ii}\big(\nabla_1\cdot\nabla_2 k(x^i,x^j)-\nabla_1 k(x^i,x^j)\cdot\nabla V(x^j)\big)\\
    &-\sum_{j=1}^N \Knu_{ij}\big( \Delta_2 k(x^j,x^i)-\nabla_2 k(x^j,x^i)\cdot\nabla V(x^i) -k(x^j,x^i)\Delta V(x^i)  \big)\\
    &+\frac{1-\nu}{N}\sum_{j=1}^N \Knu_{ij}\nabla_2 k(x^j,x^i)\cdot\phi(\underline{x};x^i)-(\nu^{-1}I-\Knu)_{ii}\nabla_1 k(x^i,x^j)\cdot\phi(\underline{x};x^j) .
\end{align*}
Notice that $\phi(\underline{x};x^i)=(\Knu \Psi)_{\cdot,i}\in \mb{R}^d$, we have $\phi(\underline{x};x^i)^\intercal=(\Psi^\intercal \Knu)_{i,\cdot}=\Psi^\intercal \Knu_{\cdot,i}=(\sum_{k=1}^N \Knu_{i,k}\Psi_{k,\cdot})^\intercal$. Hence
\begin{align*}
    \phi(\underline{x};x^i)=\sum_{k,l=1}^N \Knu_{i,k} \big(\nabla_2 k({x^k},x^l)-k({x^k},x^l)\nabla V(x^l)\big).
\end{align*}
Therefore,
\begin{align*}
    C_i^*(\underline{x}) = &  \nabla\cdot \phi(\underline{x};x^i)-\text{div}_{x^i}\phi(\underline{x};x^i)\\
    =&\sum_{j=1}^N (\nu^{-1}I-\Knu)_{ii}\big(\nabla_1\cdot\nabla_2 k(x^i,x^j)-\nabla_1 k(x^i,x^j)\cdot\nabla V(x^j)\big)\\
    &-\sum_{j=1}^N \Knu_{ij}\big( \Delta_2 k(x^j,x^i)-\nabla_2 k(x^j,x^i)\cdot\nabla V(x^i) -k(x^j,x^i)\Delta V(x^i)  \big)\\
    &+\frac{1-\nu}{N}\sum_{j,k,l=1}^N \Knu_{ij}\Knu_{ik}\nabla_2 k(x^j,x^i)\cdot \big(\nabla_2 k({x^k},x^l)-k({x^k},x^l)\nabla V(x^l)\big)\\
    &-\frac{1-\nu}{N}\sum_{j,k,l=1}^N (\nu^{-1}I-\Knu)_{ii}\Knu_{jk} \nabla_1 k(x^i,x^j)\cdot\big(\nabla_2 k({x^k},x^l)-k({x^k},x^l)\nabla V(x^l)\big) 
\end{align*}
Since $\nu^{-1}I-\Knu=(\nu I)^{-1}-\big( \tfrac{1-\nu}{N} K + \nu I \big)^{-1}=\tfrac{1}{\nu}\tfrac{1-\nu}{N}K\Knu$, we can further write the above quantity as
\begin{align*}
    C_i^*(\underline{x})=&\frac{1-\nu}{N}\frac{1}{\nu}\sum_{j,l=1}^N k(x^i,x^l)\Knu_{li}\big(\nabla_1\cdot\nabla_2 k(x^i,x^j)-\nabla_1 k(x^i,x^j)\cdot\nabla V(x^j)\big)\\
    &-\sum_{j=1}^N \Knu_{ij}\big( \Delta_2 k(x^j,x^i)-\nabla_2 k(x^j,x^i)\cdot\nabla V(x^i) -k(x^j,x^i)\Delta V(x^i)  \big)\\
    &+\frac{1-\nu}{N}\sum_{j,k,l=1}^N \Knu_{ij}\Knu_{ik}\nabla_2 k(x^j,x^i)\cdot \big(\nabla_2 k({x^k},x^l)-k({x^k},x^l)\nabla V(x^l)\big)\\
    &-\frac{(1-\nu)^2}{N^2}\frac{1}{\nu}\sum_{j,k,l,r=1}^N k(x^i,x_r)\Knu_{ri} \Knu_{jk} \nabla_1 k(x^i,x^j)\cdot\big(\nabla_2 k({x^k},x^l)-k({x^k},x^l)\nabla V(x^l)\big) \\
    =&-\sum_{j=1}^N \Knu_{ij}\big( \Delta_2 k(x^j,x^i)-\nabla_1 k(x^i,x^j)\cdot\nabla V(x^i) -k(x^j,x^i)\Delta V(x^i)  \big) \\
    &+\frac{1-\nu}{\nu}\frac{1}{N}\sum_{j,l=1}^N k(x^i,x^l)\Knu_{li}\big(\nabla_1\cdot\nabla_2 k(x^i,x^j)-\nabla_1 k(x^i,x^j)\cdot\nabla V(x^j)\big)\\
    &+\frac{1-\nu}{N}\sum_{j,k,l=1}^N \Knu_{ij}\Knu_{ik}\nabla_1 k(x^i,x^j)\cdot \big(\nabla_2 k({x^k},x^l)-k({x^k},x^l)\nabla V(x^l)\big)\\
     &-\frac{(1-\nu)^2}{\nu}\frac{1}{N^2}\sum_{j,k,l,r=1}^N \Knu_{ri} \Knu_{jk} k(x^i,x_r)\nabla_1 k(x^i,x^j)\cdot\big(\nabla_2 k({x^k},x^l)-k({x^k},x^l)\nabla V(x^l)\big), 
\end{align*}
which recovers \eqref{eq:C*}.

\noindent Lastly, under the additional assumptions, we utilize the bound of $C^*$ in Lemma \ref{lem:bound C*} (see below) to prove \eqref{eq:improved convergence regularized fisher}. Plug the bounds in Lemma \ref{lem:bound C*} into \eqref{eq:convergence of regularized Stein fisher} and we get that for any $t\in [0,T]$,
\begin{align*}
    & \quad \int_0^t \mb{E} \big[ I_{\nu,\text{Stein}}(\rho^N(s)|\pi) \big] \dee s \\
    &\le \frac{\KL(p_0^N|\pi^{\otimes N})}{N} + \frac{c^*dT}{N}\\
    & \quad + \frac{1-\nu}{N}\int_0^t \left(\beta_1 d + \beta_2 d^{\frac{1}{2}}\mb{E}\left[\mb{E}_{y\sim \rho^N(0)} [ \| \nabla V(y) \|^2 ]^{\frac{1}{2}}\right] + \nu^{-\frac{1}{2}}\beta_2 d^{\frac{1}{2}} C_V  \int_0^s \mb{E}[I_{\nu,\text{Stein}}(\rho^N(u)|\pi)^{\frac{1}{2}}]\dee u\right)  \dee s \\
    &\le \frac{\KL(p_0^N|\pi^{\otimes N})}{N} + \frac{c^* dT}{N}\\
    & \quad + (1-\nu)\left[\frac{\beta_1 d + \beta_2 d^{\frac{1}{2}}\mb{E}\left[\mb{E}_{y\sim \rho^N(0)} [ \| \nabla V(y) \|^2 ]^{\frac{1}{2}}\right]}{N} t + \frac{\beta_2 d^{\frac{1}{2}} C_V}{\nu^{\frac{1}{2}}N}\int_0^t  \left( s \int_0^s \mb{E}[I_{\nu,\text{Stein}}(\rho^N(u)|\pi)]\dee u \right)^{\frac{1}{2}}  \dee s\right],
\end{align*}
where the last inequality follows from the Jensen's inequality. Now denote $G(t)=\int_0^t \mb{E} \big[ I_{\nu,\text{Stein}}(\rho^N(s)|\pi) \big] \dee s$. After applying the fact that $G(t)$ is non-decreasing, the above inequality gives
\begin{align*}
    G(t) \le \frac{\KL(p_0^N|\pi^{\otimes N})}{N} + \frac{c^* dT}{N}  + (1-\nu)\left[\frac{\beta_1 d + \beta_2 d^{\frac{1}{2}}\mb{E}\left[\mb{E}_{y\sim \rho^N(0)} [ \| \nabla V(y) \|^2 ]^{\frac{1}{2}}\right]}{N} t + \frac{\beta_2 d^{\frac{1}{2}} C_V}{\nu^{\frac{1}{2}}N} t^{\frac{3}{2}} \sqrt{G(t)}\right].
\end{align*}
Solve the above quadratic inequality for $\sqrt{G(t)}>0$ and we get,
\begin{align*}
    &\sqrt{G(t)} \le (1-\nu)\frac{\beta_2d^{\frac{1}{2}} C_V}{2\nu^{\frac{1}{2}}N} t^{\frac{3}{2}}\\
    &\quad + \frac{1}{2}\sqrt{ (1-\nu)^2\frac{\beta_2^2 d C_V^2}{\nu N^2} t^{3} + \frac{4\KL(p_0^N|\pi^{\otimes N})}{N}  +\frac{4c^* dt}{N} +4(1-\nu)\frac{\beta_1 d + \beta_2 d^{\frac{1}{2}} \mb{E}\left[\mb{E}_{y\sim \rho^N(0)} [ \| \nabla V(y) \|^2 ]^{\frac{1}{2}}\right]}{N} t},
\end{align*}
which implies that
\begin{align*}
    G(t) &\le \frac{2\KL(p_0^N|\pi^{\otimes N}) + 2 c^* dt}{N} + 2(1-\nu)\frac{\beta_1 d + \beta_2 d^{\frac{1}{2}}\mb{E}\left[\mb{E}_{y\sim \rho^N(0)} [ \| \nabla V(y) \|^2 ]^{\frac{1}{2}}\right]}{N} t + (1-\nu)\frac{\beta_2^2 d C_V^2}{\nu N^2} t^{3}.
\end{align*}
Hence 
\begin{align*}
    \frac{1}{T}\int_0^T \mb{E}_{\underline{x}\sim p^N(t)} \big[ I_{\nu,\text{Stein}}(\rho^N(t)|\pi) \big] \dee t &\lesssim \frac{\KL(p_0^N|\pi^{\otimes N})}{NT} + \frac{c^* d}{N}\\
    &\quad + (1-\nu)\left[\frac{\beta_1 d + \beta_2 d^{\frac{1}{2}}\mb{E}\left[\mb{E}_{y\sim \rho^N(0)} [ \| \nabla V(y) \|^2 ]^{\frac{1}{2}}\right]}{N} + \frac{\beta_2^2 d C_V^2 T^2} {\nu N^2 }\right].
\end{align*}
At the end, plug the estimations of $C^*$ into \eqref{eq:KL decay} and we get for any $\epsilon>0$,
\begin{align*}
    &\quad\KL(p^N(T)|\pi^{\otimes N})\\
    & \le \KL(p^N_0|\pi^{\otimes N}) + c^* dT +(1-\nu)\left[\beta_1  d T + \beta_2 d^{\frac{1}{2}}\mb{E}\left[\mb{E}_{y\sim \rho^N(0)} [ \| \nabla V(y) \|^2 ]^{\frac{1}{2}}\right] T + \beta_2 d^{\frac{1}{2}} C_V \nu^{-\frac{1}{2}}\int_0^T t^{\frac{1}{2}}G(t)^{\frac{1}{2}}\dee t\right]\\
    &\le \KL(p^N_0|\pi^{\otimes N}) + c^* dT + (1-\nu)\left[\beta_1  d T + \beta_2 d^{\frac{1}{2}} \mb{E}\left[\mb{E}_{y\sim \rho^N(0)} [ \| \nabla V(y) \|^2 ]^{\frac{1}{2}}\right] T + \beta_2 d^{\frac{1}{2}} C_V \nu^{-\frac{1}{2}} \int_0^T \left(t\epsilon +  \frac{G(t)}{\epsilon} \right) \dee t\right]\\
    & \lesssim \KL(p^N_0|\pi^{\otimes N}) + c^* dT + (1-\nu)\left[\beta_1  d T + \beta_2 d^{\frac{1}{2}}\mb{E}\left[\mb{E}_{y\sim \rho^N(0)} [ \| \nabla V(y) \|^2 ]^{\frac{1}{2}}\right] T + \beta_2d^{\frac{1}{2}} C_V \nu^{-\frac{1}{2}}\epsilon T^2\right. \\
    &\left.\quad + \epsilon^{-1} \beta_2 d^{\frac{1}{2}}C_V\nu^{\frac{1}{2}} \left( \frac{\KL(p_0^N|\pi^{\otimes N})}{N} + \frac{\beta_1 d + \beta_2 d^{\frac{1}{2}}\mb{E}\left[\mb{E}_{y\sim \rho^N(0)} [ \| \nabla V(y) \|^2 ]^{\frac{1}{2}}\right]}{N} T + \frac{\beta_2^2 d C_V^2}{\nu N^2} T^{3} \right) T\right].
\end{align*}
Picking $\epsilon= \beta_2 d^{\frac{1}{2}} C_V\nu^{\frac{1}{2}}T/N$, we get
\begin{align*}
    \KL(p^N(T)|\pi^{\otimes N})& \lesssim \KL(p^N_0|\pi^{\otimes N}) + c^* dT + (1-\nu)\left[\beta_1  d T + \beta_2 d^{\frac{1}{2}}\mb{E}\left[\mb{E}_{y\sim \rho^N(0)} [ \| \nabla V(y) \|^2 ]^{\frac{1}{2}}\right] T + \frac{\beta_2^2 d C_V^2 T^3} {\nu N }\right].
\end{align*}
\end{proof}

\begin{lemma}[Expectation bound for $C^*(\underline{x})$]\label{lem:bound C*}
Let Assumptions \ref{assump:kernel}-\ref{ass:cstarfin} hold.
For any fixed $t\ge 0$, define
\begin{align*}
    Y(t):=\int \| \nabla V(y)\|\,\rho^N(t)(dy)
= \frac{1}{N}\sum_{i=1}^N \|\nabla V(x^i(t))\|.
\end{align*}
We have
\begin{align}\label{eq:C* bound intermediate}
    \mathbb{E}_{\underline{x}\sim p^N(t)}[ C^*(\underline{x}) ] \le  
c^* d + (1-\nu)\left(\beta_1 d + \beta_2 \sqrt{d}\mb{E}[Y(t)]\right),
\end{align}
where
\begin{align*}
\beta_1
&= B(1+C_V)
+ \frac{2B^2}{\nu^2}
+ \frac{B^3}{\nu^3}(1-\nu),\  \beta_2= B
+ \frac{2B^2}{\nu^2}
+ \frac{B^3}{\nu^3}(1-\nu).
\end{align*}
Furthermore, then for any $\nu\in(0,1]$, we have
\begin{align*}
    \mb{E}[Y(t)]\le \mb{E}[\mb{E}_{y\sim \rho^N(0)} [ \| \nabla V(y) \|^2 ]^{\frac{1}{2}}] + \nu^{-\frac{1}{2}}C_V \int_0^t \mb{E}[I_{\nu,\text{Stein}}(\rho^N(s)|\pi)^{\frac{1}{2}}]\dee s .
\end{align*}
\end{lemma}
\begin{remark}\label{eq:C* bound generalization} The first part of the Lemma, i.e., \eqref{eq:C* bound intermediate} doesn't rely on the (time) continuity of the dynamics of $\underline{x}(t)$. As a consequence, the same result holds if we replace $\underline{x}(t)$ by the discrete-time R-SVGD $X_n$, which we will use in the proof of Theorem \ref{thm:decay of regularized fisher svgd}.  
\end{remark}
{\namedproof{Proof of Lemma \ref{lem:bound C*}.}
Recall the expression of $C^*(\underline{x})$ in \eqref{eq:C*}:
{\small
\begin{align*}
    C^*(\underline{x})=&\underbrace{-\frac{1}{N}\sum_{i,j=1}^N \Knu_{ij}\big( \Delta_2 k(x^j,x^i)-\nabla_1 k(x^i,x^j)\cdot\nabla V(x^i) -k(x^j,x^i)\Delta V(x^i)  \big)}_{\coloneqq A_1}\\
    &\underbrace{+\frac{1-\nu}{\nu}\frac{1}{N^2}\sum_{i,j,l=1}^N k(x^i,x^l)\Knu_{li}\big(\nabla_1\cdot\nabla_2 k(x^i,x^j)-\nabla_1 k(x^i,x^j)\cdot\nabla V(x^j)\big) }_{\coloneqq A_2}\\
    &\underbrace{+\frac{1-\nu}{N^2}\sum_{i,j,k,l=1}^N \Knu_{ij}\Knu_{ik}\nabla_1 k(x^i,x^j)\cdot \big(\nabla_2 k({x^k},x^l)-k({x^k},x^l)\nabla V(x^l)\big)}_{A_3}\\
     &\underbrace{-\frac{(1-\nu)^2}{\nu}\frac{1}{N^3}\sum_{i,j,k,l,r=1}^N \Knu_{ri} \Knu_{jk} k(x^i,x_r)\nabla_1 k(x^i,x^j)\cdot\big(\nabla_2 k({x^k},x^l)-k({x^k},x^l)\nabla V(x^l)\big)}_{\coloneqq A_4}. 
\end{align*}}
We bound the 4 terms $A_1, A_2, A_3, A_4$ in the decomposition respectively. According to Assumptions~\ref{assump:kernel} and \ref{assump:target}:
\begin{align*}
    |k|\le B,\quad 
\|\nabla_1 k\|\le \sqrt{d} B,\quad 
|\nabla_1 \cdot\nabla_2 k|\le dB,\quad 
|\Delta_1 k|,|\Delta_2 k|\le dB,\quad 
\|\nabla^2 V\|_{\mathrm{op}} \le C_V,
\end{align*}
and the last bound also implies $|\Delta V|\le dC_V$.
Since $K\succeq 0$, we have the following bound for the $i$th row for any $i \in [N]$:
\begin{align*}
\big\|K^{\nu}(\underline{x})_{i,\cdot}\big\|_2
=\big\|\left(\big((1-\nu)K/N+\nu I\big)^{-1}\right)_{i,\cdot}\big\|_2
\le\nu^{-1}.    
\end{align*}
Moreover, observe that
\begin{align*}
    \Knu(\underline{x}) = \left(I_N - (1-\nu)\left(I_N - \frac{K}{N}\right)\right)^{-1} = I + (1-\nu)R,
\end{align*}
where
$
\|R_{i,\cdot}\|_2 \le 1
$
for any $i \in [N]$.
Write
\begin{align*}
    A_1 = -\frac{1}{N}\sum_{i=1}^N \big( \Delta_2 k(x^i,x^i)-\nabla_1 k(x^i,x^i)\cdot\nabla V(x^i) -k(x^i,x^i)\Delta V(x^i)  \big) + \hat{A}_1
\end{align*}
where
\begin{align*}
    \hat{A}_1 := -\frac{1-\nu}{N}\sum_{i,j=1}^N R_{ij}\big( \Delta_2 k(x^j,x^i)-\nabla_1 k(x^i,x^j)\cdot\nabla V(x^i) -k(x^j,x^i)\Delta V(x^i)  \big).
\end{align*}

 For each term of $\hat{A}_1,A_2,A_3,A_4$, we count the numbers of occurrences of factors $K^{\nu}$, $R$, $k$ and its derivatives, and track their contributions to the magnitude using the bounds above.

\noindent Taking $\hat{A}_1$ for example, we express $A_1$ as average over indices $i$, i.e., $\hat{A}_1=\tfrac{1}{N}\sum_{i=1}^N \hat{A}_{1,i}$. For each $i$, the block $\hat{A}_{1,i}$ has one row of $R$ multiplying a column that is bounded in $L^2$-norm by $dB+B\sqrt{d}\|\nabla V(x^i)\|+dBC_V$. 
Therefore, Cauchy--Schwarz in the $j$-index and $\| R_{i,\cdot}\|_2\le 1$ give
\begin{align*}
    |\hat{A}_{1,i}|\le (1-\nu)\left(dB+B \sqrt{d}\|\nabla V(x^i)\|+dBC_V\right).
\end{align*}
Averaging over $i$ and taking expectation,
\begin{align*}
    \mathbb{E}|\hat{A}_1|\le B(1-\nu)\left((1+C_V)d
+ \sqrt{d}\mb{E}[Y(t)]\right).
\end{align*}
Similarly, for $A_2,A_3,A_4$, using $\big\|K^{\nu}(\underline{x})_{i,\cdot}\big\|_2 \le \nu^{-1}$ instead of $\| R_{i,\cdot}\|_2\le 1$, we have
\begin{align*}
    \mathbb{E}|A_2| &\le
\frac{1-\nu}{\nu}\frac{1}{\nu}\big(dB^2+B^2\sqrt{d}\mb{E}[Y(t)]\big),\\
\mathbb{E}|A_3|&\le
(1-\nu)\frac{1}{\nu^2}\big(dB^2+B^2\sqrt{d}\mb{E}[Y(t)]\big),\\
\mathbb{E}|A_4|&\le
\frac{(1-\nu)^2}{\nu}\frac{1}{\nu^2}\big(dB^3+B^3\sqrt{d}\mb{E}[Y(t)]\big).
\end{align*}
Combining the 4 estimations, and using Assumption \ref{ass:cstarfin}, we obtain 
\eqref{eq:C* bound intermediate}.

\noindent The second part of the proof relies on bounding $\mb{E}[Y(t)]$ based on the $C_V$-smoothness of $V$ and the fact that $x^i(t)$ follows the R-SVGD dynamics. Denote $F(t)\coloneqq \mb{E}_{y\sim \rho^N(t)}[\| \nabla  V(y) \|^2]$. Then we have
\begin{align*}
    F'(t) & =\frac{1}{N}  \sum_{i=1}^N \frac{\dee}{\dee t} \| \nabla V(x^i(t)) \|^2 \\
    & = \frac{2}{N}  \sum_{i=1}^N \langle \nabla^2 V(x^i(t)) \nabla V(x^i(t)), \frac{\dee}{\dee t}x^i(t)\rangle \\
    & \le \frac{2C_V}{N}  \sum_{i=1}^N \| \nabla V(x^i(t))\| \| \frac{\dee}{\dee t}x^i(t)  \| \\
    & \le 2C_V F(t)^{\frac{1}{2}} N^{-\frac{1}{2}} \| \left((1-\nu)\iota_{k,\rho^N(t)}^*\iota_{k,\rho^N(t)}+\nu I_d \right)^{-1} \iota_{k,\rho^N(t)}^*\iota_{k,\rho^N(t)}\nabla\log \frac{\rho^N(t)}{\pi} \|_{L^2(\rho^N(t))} \\
    & \le 2\nu^{-\frac{1}{2}} C_V F(t)^{\frac{1}{2}} I_{\nu,\text{Stein}}(\rho^N(t)|\pi)^{\frac{1}{2}},
\end{align*}
where in the second last inequality, the notation $\iota_{k,\rho^N(t)}^*\iota_{k,\rho^N(t)}\nabla\log \tfrac{\rho^N(t)}{\pi}$ represents a term as a whole, which is well-defined through integration by parts even though $\nabla\log \tfrac{\rho^N(t)}{\pi}$ itself is not well-defined, i.e., $\iota_{k,\rho^N(t)}^*\iota_{k,\rho^N(t)}\nabla\log \tfrac{\rho^N(t)}{\pi}=\sum_{j=1}^N k(x^l(t),x^j(t))\nabla V(x^j(t))-\nabla_2 k(x^l(t),x^j(t)) $.
In the last identity, we used the spectrum expression of $I_{\nu,\text{Stein}}$ introduced in \cite{he2024regularized}. From the above differential inequality, we obtain
$$
\sqrt{F(t)} \le \mb{E}_{y\sim \rho^N(0)} [ \| V(y) \|^2 ]^{\frac{1}{2}} + \nu^{-\frac{1}{2}} C_V \int_0^t I_{\nu,\text{Stein}}(\rho^N(s)|\pi)^{\frac{1}{2}} \dee s.
$$
With the above bound and Jensen's inequality, we obtain
\begin{align*}
    \mb{E}[Y(t)] &= \mb{E}[\frac{1}{N}\sum_{i=1}^N \|\nabla V(x^i(t))\|]  \le \mb{E}[\big(\mb{E}_{y\sim \rho^N(t)}[\| \nabla  V(y) \|^2]\big)^{\frac{1}{2}}]=\mb{E}[F(t)^{\frac{1}{2}}]\\
    &\le \mb{E}[\mb{E}_{y\sim \rho^N(0)} [ \| V(y) \|^2 ]^{\frac{1}{2}}] + \nu^{-\frac{1}{2}} C_V \mb{E}[\int_0^t I_{\nu,\text{Stein}}(\rho^N(s)|\pi)^{\frac{1}{2}} \dee s].
\end{align*}
\end{proof}}

\namedproof{Proof of Corollary \ref{cor:average regularize fisher}.}
According to Theorem \ref{thm:convergence} and the convexity of $I_{\nu_N,\text{Stein}}(\cdot|\pi)$, as proved in Lemma \ref{lem:convexity of regularized fisher}, we have
\begin{align*}
    &\mb{E}[I_{\nu_N,\text{Stein}}(\rho_{av}^N|\pi)]\le \frac{1}{N}\int_0^{N} \mb{E} \big[ I_{\nu_N,\text{Stein}}(\rho^N(t)|\pi) \big] \dee t \\
    &\quad\lesssim \frac{\KL(p_0^N|\pi^{\otimes N})}{N^2} + \frac{c^* d}{N} + \frac{\beta_{1,N} d + \beta_{2,N} \mb{E}_{y\sim \rho^N(0)} [ \| \nabla V(y) \|^2 ]^{\frac{1}{2}}d^{\frac{1}{2}}}{N^2} +\frac{\nu_N^{-1}\beta_2^2 C_V^2 d}{N}.
\end{align*}
\end{proof}

\namedproof{Proof of Theorem \ref{thm:weak convergence under exchangeability}.} Under conditions in Theorem \ref{thm:convergence}, suppose that the law $p_0^N$ of the initial particles $x^1(0),x^2(0), \cdots, x^N(0)$ is exchangeable for each $N\in \mb{N}$. By the subadditivity of relative entropy \cite[Lemma 2.4 b and Theorem 2.6]{budhiraja2019analysis}, 
$$
\KL(\mathrm{Law}(x^1(T)) | \pi ) \le \frac{1}{N}\KL(p^N(T)|\pi^{\otimes N}),
$$
and we have the following from \eqref{eq:KL bound} for $T>0$:
 \begin{align*}
        &\KL(\mathrm{Law}(x^1(T)) | \pi )\\
        &\quad \lesssim \frac{\KL(p^N_0|\pi^{\otimes N})}{N} + \frac{c^* dT}{N} + \frac{1-\nu}{N}\left(\beta_1  d T + \beta_2  \mb{E}[\mb{E}_{y\sim \rho^N(0)} [ \| \nabla V(y) \|^2 ]^{\frac{1}{2}}] d^{\frac{1}{2}}T + \frac{\beta_2^2 C_V^2 d T^3} {\nu N }\right).
    \end{align*} 
By the variational representation of relative entropy \cite[Proposition 2.3]{budhiraja2019analysis}, this gives tightness of $\{\bar{\rho}^N\}_{N\ge 1}$ when $\limsup_{N \rightarrow \infty} \frac{\KL(p_0^N|\pi^{\otimes N})}{N} < \infty$. Furthermore, following the idea in \cite[Theorem 2]{balasubramanian2024improved}, we have
\begin{align*}
    &I_{\nu, \text{Stein}}(\bar{\rho}^N|\pi) =  I_{\nu, \text{Stein}}(\frac{1}{(1-\nu)^{-\frac{1}{3}}N^{\frac{2}{3}}}\int_0^{(1-\nu)^{-\frac{1}{3}}N^{\frac{2}{3}}} \mathrm{Law}(x^1(t))\dee t|\pi) \\
    &\quad\le \frac{1}{(1-\nu)^{-\frac{1}{3}}N^{\frac{2}{3}}}\int_0^{(1-\nu)^{-\frac{1}{3}}N^{\frac{2}{3}}} I_{\nu, \text{Stein}}(\mathrm{Law}(x^1(t))|\pi) \dee t,\quad \quad \quad \text{using convexity of }I_{\nu,\text{Stein}}(\cdot|\pi) \\
    &\quad = \frac{1}{(1-\nu)^{-\frac{1}{3}}N^{\frac{2}{3}}}\int_0^{(1-\nu)^{-\frac{1}{3}}N^{\frac{2}{3}}} I_{\nu, \text{Stein}}(\mb{E}[\rho^N(t)]|\pi)\dee t,\quad \quad \quad \quad \text{using exchangeability} \\
    &\quad\le \frac{1}{(1-\nu)^{-\frac{1}{3}}N^{\frac{2}{3}}}\int_0^{(1-\nu)^{-\frac{1}{3}}N^{\frac{2}{3}}} \mb{E}[I_{\nu, \text{Stein}}(\rho^N(t)|\pi)] \dee t \to 0, \quad \text{using convexity of }I_{\nu,\text{Stein}}(\cdot|\pi)\ \text{and Theorem \ref{thm:convergence}}
\end{align*}
where the convexity of $I_{\nu,\text{Stein}}(\cdot|\pi)$ is proved in Lemma \ref{lem:convexity of regularized fisher}, and $\mb{E}[\rho^N(t)](\dee x) \coloneqq \mb{E}[\rho^N(t,\dee x)]$.
 According to Appendix \ref{append:KSD and regularized Fisher}, the above implies $\mathrm{KSD}(\bar{\rho}^N|\pi)\to 0$ as $N\to\infty$. Therefore, applying \cite[Theorem 7]{gorham2017measuring}, $\bar{\rho}^N \rightharpoonup \pi$ as $N\to\infty$. The above calculations all go through for $\nu = \nu_N = 1-\frac{1}{N}$.
\end{proof}

\namedproof{Proof of Theorem \ref{thm:fisher convergence dynamcis}.}
    First, we can bound $I_{\nu,\text{Stein}}(\mu^N(t)|\pi)$. According to \eqref{eq:average population density0}, we have 
\begin{align*}
    I_{\nu,\text{Stein}}(\mu^N(t)|\pi)&=I_{\nu,\text{Stein}}\bigg(\mb{E}_{\underline{x}(0) \sim p_0^N} [ \rho^N(t) ] | \pi\bigg) \le  \mb{E}_{\underline{x}(0) \sim p_0^N} [I_{\nu,\text{Stein}}(\rho^N(t)|\pi)] ,
\end{align*}
where the inequality follows from the convexity of $I_{\nu,\text{Stein}}(\cdot|\pi)$ (Lemma \ref{lem:convexity of regularized fisher}). Then it follows from \eqref{eq:fisher equivalence} that
\begin{align*}
    & \frac{1}{T}\int_0^T I(\mu^N(t)|\pi) \dee t \\
    & \quad\le \frac{2}{T}\int_0^T (1-\nu)I_{\nu,\text{Stein}}(\mu^N(t)|\pi) \dee t \\
    & \quad\lesssim \frac{1-\nu}{T}\int_0^T   \mb{E} [I_{\nu,\text{Stein}}(\rho^N(t)|\pi)] \dee t \\
    & \quad\lesssim \frac{\KL(p_0^N|\pi^{\otimes N})}{NT} + \frac{c^* d}{N} + \frac{\beta_1 d + \beta_2 \mb{E}[\mb{E}_{y\sim \rho^N(0)} [ \| \nabla V(y) \|^2 ]^{\frac{1}{2}}]d^{\frac{1}{2}}}{N} + \frac{\beta_2^2 C_V^2 d T^2} {\nu N^2 },
\end{align*}
where the last inequality follows from Theorem \ref{thm:convergence}. Last, \eqref{eq:average fisher convergence} follows from the fact that $I(\mu_{av}^N|\pi)\le \tfrac{1}{N^{2/3}}\int_0^{N^{2/3}} I(\mu^N(t)|\pi)\dee t$ which is true due to the definition of $\mu_{av}^N$ and convexity of $I(\cdot|\pi)$.
\end{proof}

\section{Proofs for Section~\ref{sec:rsvgdalgorithm}}
First we prove our main results assuming some technical lemmas, which will be proved subsequently.
\namedproof{Proof of Theorem \ref{thm:decay of regularized fisher svgd}.} For simplicity, we denote $\En_n(t)\coloneqq \KL(q_{n,t}|\pi^{\otimes N})$ for all $t\in [0,h_{n+1}]$ in our proof. First, for all $n\ge 0$, using the integral property of $\En_n$, we have
\begin{align}\label{eq:single step KL decay}
\En_n(h_{n+1})= \En_n(0) + h_{n+1}\En'_n(0) +\int_0^{h_{n+1}} (h_{n+1}-t)\En''_n(t) \dee t.
\end{align}
The $\KL$-decaying magnitude can be upper bounded via bounding $\En'_n(0)$ and $\En''_n(t)$ for all $t\in [0,h_{n+1}]$. Therefore, we begin with expressing $\En'_n(0)$ and $\En''_n(t)$. 

\noindent Since $q_{n,t}$ is the push-forward density from ${p_n^N}$ via $\Psi_{n}(t,\cdot)$, it can be expressed as
\begin{align*}
    q_{n,t}(\underline{x}) = \det \big( J\Psi_n(t,\cdot) \circ \Psi_n^{-1}(t,\underline{x}) \big)^{-1} {p_n^N}(\Psi_n^{-1}(t,\underline{x})),
\end{align*}
where $J\Psi$ denotes the Jacobian of $\Psi$. Following the same computations in \cite{korba2020non,balasubramanian2024improved}, we obtain
\begin{align}\label{eq:1st derivative expression}
    \En_n'(0) & = \int \big( \text{div} (\Tm_{n+1}(\underline{x})) - \langle {\nabla V}(\underline{x}), \Tm_{n+1}(\underline{x}) \rangle \big) {p_n^N}(\underline{x}) \dee \underline{x}
\end{align}
and 
\begin{align*}
    \En_n''(t) & = \underbrace{\int \langle \Tm_{n+1}(\underline{x}), H_V(\Psi_n(t,\underline{x})) \Tm_{n+1}(\underline{x}) \rangle {p_n^N}(\underline{x}) \dee \underline{x}}_{\coloneqq \psi_1(t)} \\
    &\quad + \underbrace{\int \| J\Tm_{n+1}(\underline{x})(\id - t J\Tm_{n+1}(\underline{x}) )^{-1} \|_{\HS}^2 {p_n^N}(\underline{x})\dee \underline{x}}_{\coloneqq \psi_2(t)}.
\end{align*}
To bound $\En_n'(0)$, notice that 
\begin{align*}
    \text{div} (\Tm_{n+1}(\underline{x})) - \langle {\nabla V}(\underline{x}), \Tm_{n+1}(\underline{x}) \rangle & = -\frac{1}{N} \sum_{k=1}^N \text{div}_{x^j} \phi(\underline{x};x^j) +\langle \nabla V(x^j), \phi(\underline{x};x^j) \rangle,
\end{align*}
where $\phi$ is defined in \eqref{eq:matrix vector equation for the inverse operator} with $\nu=\nu_{n+1}$. Then as shown in the proof of Theorem \ref{thm:convergence}, 
\begin{align*}
    \En_n'(0) & = -N \mb{E}_{\underline{x}\sim {p_n^N}} [ I_{\nu_{n+1},\text{Stein}}(\rho^N|\pi) ] + \mb{E}_{\underline{x}\sim {p_n^N}} [ C^*(\underline{x}) ].
\end{align*}
where $\rho^N=\frac{1}{N}\sum_{i=1}^N \delta_{x^i}$. 

\noindent Under the initial condition of $p_0^N$ in Theorem \ref{thm:decay of regularized fisher svgd},
denote the law of $\underline{X}_n$ starting from $\underline{x}(0)\sim {p^N_{0,K}}$ by ${p^N_{n,K}}$. We can apply the above computations to ${p^N_{n,K}}$ rather than ${p_n^N}$. We work on the $\underline{X}_n$ with $\underline{x}(0)\sim {p^N_{0,K}}$ in the following proof.

\noindent According to Lemma \ref{lem:bound C*} and Remark \ref{rem:c^* bound}, we can upper bound $\mb{E}_{\underline{x}\sim {p_{n,K}^N}} [ C^*(\underline{x}) ]$ as follows:
\begin{align*}
    &\quad \mb{E}_{\underline{x}\sim p_{n,K}^N}[C^*(\underline{x})] \\
    &\le c^* d + (1-\nu_{n+1})\big(\beta_1 d + \beta_2 d^{\frac{1}{2}} \mb{E}[\frac{1}{N}\sum_{i=1}^N \| \nabla V(X_n^i) \|]\big) \\ 
     &\le c^* d + (1-\nu_{n+1}) \bigg(\beta_1 d + \beta_2 d^{\frac{1}{2}} A \big( \frac{1}{N}\sum_{i=1}^N \mb{E}[V(X_n^i)] \big)^\alpha\bigg) \\
     & \le c^* d + (1-\nu_{n+1}) \bigg(\beta_1 d + \beta_2 d^{\frac{1}{2}} A \bigg[ M \big(d^{\frac{1}{1-\alpha}}+\frac{1}{N}\sum_{i=1}^N V(X^i_0)\big) \big[ (\sum_{l=1}^n h_l\nu_l^{-1})^{\frac{1}{1-\alpha}} \vee 1\big] \bigg]^\alpha\bigg)\\
     & \le c^* d + (1-\nu_{n+1})\bigg(\beta_1 d + \beta_2 d^{\frac{1}{2}} A \bigg[ M \big(d^{\frac{1}{1-\alpha}}+K \big) \big[ (\sum_{l=1}^n h_l\nu_l^{-1})^{\frac{1}{1-\alpha}} \vee 1\big] \bigg]^\alpha\bigg)\\
     &\le c^* d+ (1-\nu_{n+1})\big(\beta_1 d + \beta_2 B^{-2}N^{-\frac{1}{2}}C_{n+1}^{\frac{1}{2}}\nu_{n+1}^2\big),
\end{align*}
where the second inequality follows from Assumption \ref{assump:convergence} and Jensen's inequality. The third inequality follows from Lemma \ref{lem:trajectory bound of V empirical expectation}. The second last inequality follows from the definition of $p_{0,K}^N$. The last identity follows from the definition of $C_n$ in \eqref{eq:choice of stepsize n+1}. Note that $\beta_1$ and $\beta_2$ are defined in \eqref{eq:constants alpha beta} with $\nu=\nu_{n+1}$. So $\beta_1=\beta_{1,n+1}$ and $\beta_2=\beta_{2,n+1}$. For simplicity, we will drop the $n$ in the sub-indices for now and add it back at the end of the proof. 

\noindent Based on the bound of $\mb{E}_{\underline{x}\sim p_{n,K}^N}[C^*(\underline{x})]$, we get
\begin{align}\label{eq:E'(0) bound}
    E_n'(0) \le -N \mb{E}_{\underline{x}\sim {p_{n,K}^N}} [ I_{\nu_{n+1},\text{Stein}}(\rho^N|\pi) ]  + c^* d + (1-\nu_{n+1})\big(\beta_1 d + \beta_2 B^{-2}N^{-\frac{1}{2}}C_{n+1}^{\frac{1}{2}}\nu_{n+1}^2\big).
\end{align}
\vspace{.2cm}

\noindent Next, we bound $\En_n''(t)$, by estimating the two terms $\psi_1(t)$ and $\psi_2(t)$ respectively. For $\psi_1(t)$, according to Lemma \ref{lem:upper bound psi 1} and Assumption \ref{assump:kernel}, we have
\begin{align}\label{eq:upper bound psi 1}
    \psi_1(t) & =\int \langle \Tm_{n+1}(\underline{x}), H_V(\Psi_n(t,\underline{x})) \Tm_{n+1}(\underline{x}) \rangle {p_{n,K}^N}(\underline{x}) \dee \underline{x} \le C_V \int \| \Tm_{n+1}(\underline{x}) \|^2 {p_{n,K}^N}(\underline{x}) \dee \underline{x} \nonumber \\
    &\le C_V\frac{B}{(1-\nu_{n+1})B+\nu_{n+1}} \mb{E}_{\underline{x}\sim {p_{n,K}^N}}[I_{\nu_{n+1},\text{Stein}}(\rho^N|\pi)].
\end{align}
For $\psi_2(t)$, we have
\begin{align}\label{eq:term psi 2}
    \psi_2(t) & = \int \| J\Tm_{n+1}(\underline{x})(\id - t J\Tm_{n+1}(\underline{x}) )^{-1} \|_{\HS}^2 {p_{n,K}^N}(\underline{x})\dee \underline{x} \nonumber\\
    &\le \int \| J\Tm_{n+1}(\underline{x})\|_{\HS}^2\| (\id - t J\Tm_{n+1}(\underline{x}) )^{-1} \|_{2}^2 {p_{n,K}^N}(\underline{x})\dee \underline{x}.
\end{align}
Notice that
\begin{align*}
    \| (\id - t J\Tm_{n+1}(\underline{x}) )^{-1} \|_{2} \le \sum_{m=0}^\infty \| t J\Tm_{n+1}(\underline{x}) \|_2^m\le \sum_{m=0}^\infty \| J\Tm_{n+1}(\underline{x}) \|_{\HS}^m t^m.
\end{align*}
Therefore, we can bound $\psi_2(t)$ via bounding $\| J\Tm_{n+1}(\underline{x}) \|_{\HS}$. 

\vspace{.2cm}

\noindent Under the initial condition of $p_{0,K}^N$, according to Lemma \ref{lem:Jacobian pointwise bound}, we have
\begin{align*}
    \| J\Tm_{n+1}(\underline{X}_n) \|_{\HS}^2 &\le 9\nu_{n+1}^{-4}A^2B^4 N d \left[ M^{2\alpha} \left(d^{\frac{1}{1-\alpha}}+K\right)^{2\alpha} \big[ (\sum_{l=1}^n h_l\nu_l^{-1})^{\frac{2\alpha}{1-\alpha}} \vee 1\big] \right]  \\
        & +9\nu_{n+1}^{-4} B^4 N d^2+ 3\nu_{n+1}^{-2}B^2 C_V^2 d.
\end{align*}
By picking $ h_{n+1}/\nu_{n+1} = C_{n+1}^{-1/2} \theta/2$ with $\theta\in [0,1]$ and $C_{n+1}$ in \eqref{eq:choice of stepsize n+1}, i.e.,
\begin{align*}
   C_{n+1}=18 \max\{ \nu_{n+1}^{-4} A^2 B^4 N d M^{2\alpha} \left(d^{\frac{1}{1-\alpha}}+K\right)^{2\alpha}(\sum_{l=1}^n h_l\nu_l^{-1} \vee 1)^{\frac{2\alpha}{1-\alpha}}, \nu_{n+1}^{-4}B^4 N d^2 + B^2 C_V^2 d\},
\end{align*}
for any $t\in [0,h_{n+1}]$, we have $\|  t J\Tm_{n+1}(\underline{x}) \|_{\HS}\le\nu_{n+1}C_{n+1}^{-1/2}\|  J\Tm_{n+1}(\underline{x}) \|_{\HS}\theta/2\le \nu_{n+1}/2\le  1/2$,
 which implies that
\[\| (\id - t J\Tm_{n+1}(\underline{x}) )^{-1} \|_{2} \le \sum_{m=0}^\infty (1/2)^m\le 2.\]
Hence, we can further bound $\psi_2(t)$ in \eqref{eq:term psi 2} using Lemma \ref{lem:upper bound psi 2} and Lemma \ref{lem:trajectory bound of V empirical expectation}.
\begin{align*}
    \psi_2(t)&\le 4 \int \| J\Tm_{n+1}(\underline{x})\|_{\HS}^2 {p_{n,K}^N}(\underline{x})\dee \underline{x} \\
         &\le 8Bd(1-\nu_{n+1})^2\nu_{n+1}^{-3}\mb{E}_{\underline{x}\sim {p_{n,K}^N}}[I_{\nu_{n+1},\text{Stein}}(\rho^N|\pi)]+ 4\nu_{n+1}^{-2}B^2 d^2 (2N+3) + 12\nu_{n+1}^{-2} B^2 C_V^2 d\\
        &\quad + 4\nu_{n+1}^{-2}A^2B^2 d (2N+3)\sup_{\underline{x}\in \mc{S}_{n,K}}\big( \frac{1}{N} \sum_{i=1}^N V(x^i)\big)^{2\alpha}\\
        &\le 8Bd(1-\nu_{n+1})^2\nu_{n+1}^{-3}\mb{E}_{\underline{x}\sim {p_{n,K}^N}}[I_{\nu_{n+1},\text{Stein}}(\rho^N|\pi)]+ 4\nu_{n+1}^{-2}B^2 d^2 (2N+3) + 12\nu_{n+1}^{-2} B^2 C_V^2 d\\
        &\quad + 4\nu_{n+1}^{-2}A^2 B^2 d (2N+3)  \left[ M^{2\alpha} \left(d^{\frac{1}{1-\alpha}}+K\right)^{2\alpha} \big[ (\sum_{l=1}^n h_l\nu_l^{-1})^{\frac{2\alpha}{1-\alpha}} \vee 1\big] \right]\\
        &\le 8Bd(1-\nu_{n+1})^2\nu_{n+1}^{-3}\mb{E}_{\underline{x}\sim {p_{n,K}^N}}[I_{\nu_{n+1},\text{Stein}}(\rho^N|\pi)]+ C_{n+1} \\
        &\le 8Bd(1-\nu_{n+1})^2\nu_{n+1}^{-3}\mb{E}_{\underline{x}\sim {p_{n,K}^N}}[I_{\nu_{n+1},\text{Stein}}(\rho^N|\pi)]+\nu_{n+1}^2 \theta^2 h_{n+1}^{-2}/4.
\end{align*}
Therefore, combining estimations of $\psi_1$ and $\psi_2$, we have
\begin{align}\label{eq:E''(t)}
    E''_n(t) &\le \big( C_V \big((1-\nu_{n+1})B +\nu_{n+1} \big)^{-1}B+8Bd(1-\nu_{n+1})^2\nu_{n+1}^{-3} \big)\mb{E}_{\underline{x}\sim {p_{n,K}^N}}[I_{\nu_{n+1},\text{Stein}}(\rho^N|\pi)] \nonumber\\
    &\quad +\nu_{n+1}^2 \theta^2 h_{n+1}^{-2}/4 .
\end{align}
Plugging \eqref{eq:E'(0) bound} and \eqref{eq:E''(t)} into \eqref{eq:single step KL decay}, and noting from \eqref{eq:condition on step size and regularizer} that
\begin{align*}
    h_{n+1}\le \frac{N}{C_V \big((1-\nu_{n+1})B +\nu_{n+1} \big)^{-1}B+8Bd(1-\nu_{n+1})^2\nu_{n+1}^{-3}},    
\end{align*}
we get (adding back the $n$-dependency in $\beta_1$ and $\beta_2$)
\begin{align*}
    &\quad \KL({p^N_{n+1,K}}|\pi^{\otimes N})\\
    &\le \KL({p^N_{n,K}}|\pi^{\otimes N}) \\
    &\quad + h_{n+1}\big(-N \mb{E}_{\underline{x}\sim {p^N_{n,K}}} [ I_{\nu_{n+1},\text{Stein}}(\rho^N|\pi) ]+ c^*d + (1-\nu_{n+1})\big(\beta_{1,n+1} d + \beta_{2,n+1} B^{-2}N^{-\frac{1}{2}}C_{n+1}^{\frac{1}{2}}\nu_{n+1}^2\big)\big) \\
    & \quad + \frac{1}{2}\big( C_V ((1-\nu_{n+1})B+\nu_{n+1})^{-1}+4 d (1-\nu_{n+1})^2\nu_{n+1}^{-3} \big) B h_{n+1}^2\mb{E}_{\underline{x}\sim {p^N_{n,K}}} [ I_{\nu_{n+1},\text{Stein}}(\rho^N|\pi) ]  \\
    &\quad + \frac{1}{8}\nu_{n+1}^{2} \theta^2\\
    &\le \KL({p^N_{n,K}}|\pi^{\otimes N}) - \frac{1}{2}h_{n+1} N \mb{E}_{\underline{x}\sim {p^N_{n,K}}} [ I_{\nu_{n+1},\text{Stein}}(\rho^N|\pi) ] + c^* d h_{n+1}+ (1-\nu_{n+1}) h_{n+1}\beta_{1,n+1} d \\
    &\quad + \frac{1}{2}(1-\nu_{n+1})\beta_{2,n+1} B^{-2}N^{-\frac{1}{2}}\theta \nu_{n+1}^3+\frac{1}{8}\nu_{n+1}^{2} \theta^2,
\end{align*}
where the last inequality follows from the upper bound of $h_{n+1}$ and the fact that $h_{n+1}/\nu_{n+1}=C_{n+1}^{-\frac{1}{2}}\theta/2$.
Then using the telescoping sum, we get for all $T\ge 2$,
\begin{align*}
    \sum_{n=1}^{T} h_{n}\mb{E}_{\underline{x}\sim {p^N_{n-1,K}}} [ I_{\nu_{n},\text{Stein}}(\rho^N|\pi) ]&\le \frac{2\KL({p^N_{0,K}}|\pi^{\otimes N})}{N} + \frac{2c^* d\sum_{n=1}^T h_n}{N} + \frac{2 d \sum_{n=1}^T \beta_{1,n} (1-\nu_n) h_n}{N} \\
    &\quad+ \frac{\theta \sum_{n=1}^T \beta_{2,n} (1-\nu_n) \nu_{n}^3 }{B^2 N^{\frac{3}{2}}} + \frac{\theta^2 \sum_{n=1}^T\nu_n^2 }{4N} ,
\end{align*}
i.e.
\begin{align*}
      \mb{E}_{{p^N_{0,K}}} \left[\frac{1}{T} \sum_{n=1}^T h_n I_{\nu_{n},\text{Stein}}(\rho^N_{n-1}|\pi)  \right]&\le \frac{2\KL({p^N_{0,K}}|\pi^{\otimes N})}{NT} + \frac{2c^*d\sum_{n=1}^T h_n}{NT} +  \frac{2 d \sum_{n=1}^T \beta_{1,n}(1-\nu_n) h_n}{NT} \\
      &\quad+ \frac{\theta \sum_{n=1}^T \beta_{2,n}(1-\nu_n) \nu_{n}^3 }{B^2 N^{\frac{3}{2}} T } + \frac{\theta^2 \sum_{n=1}^T\nu_n^2 }{4NT}.
\end{align*}
According to \cite{balasubramanian2024improved}, $\KL({p^N_{0,K}}|\pi^{\otimes K})\le 2 \KL(p_0^N|\pi^{\otimes N})+\log 2 \le \gamma d N$ where $\gamma\coloneqq 2C_{\KL}+\log 2$. Therefore,
\begin{align*}
     \mb{E}_{{p^N_{0,K}}} \left[\frac{1}{T} \sum_{n=1}^T h_n I_{\nu_{n},\text{Stein}}(\rho^N_{n-1}|\pi)  \right]&\le \frac{2\gamma d }{T} +\frac{2c^*d\sum_{n=1}^T h_n}{NT} + \frac{2 d \sum_{n=1}^T \beta_{1,n}(1-\nu_n) h_n}{NT} \\
     &\quad + \frac{\theta \sum_{n=1}^T \beta_{2,n}(1-\nu_n) \nu_{n}^3 }{B^2 N^{\frac{3}{2}} T } + \frac{\theta^2 \sum_{n=1}^T\nu_n^2 }{4NT}.
\end{align*}
\end{proof}
\namedproof{Proof of Theorem \ref{thm:decay of fisher}.} First, following the proof of Theorem \ref{thm:decay of regularized fisher svgd}, we have
    \begin{align*}
        \KL({p^N_{n+1,K}}|\pi^{\otimes N})&\le \KL({p^N_{n,K}}|\pi^{\otimes N}) - \frac{1}{2}h_{n+1} N \mb{E}_{\underline{x}\sim {p^N_{n,K}}} [ I_{\nu_{n+1},\text{Stein}}(\rho^N|\pi) ] + c^* d h_{n+1}  \\
    &\quad + h_{n+1}\beta_{1,n+1}(1-\nu_{n+1}) d + \frac{1}{2}\beta_{2,n+1}(1-\nu_{n+1}) B^{-2}N^{-\frac{1}{2}}\theta \nu_{n+1}^3 +\frac{1}{8}\nu_{n+1}^{2} \theta^2.
    \end{align*}
Note that according to Jensen's inequality
\begin{align*}
    \mb{E}_{\underline{x}\sim {p^N_{n,K}}} [ I_{\nu_{n+1},\text{Stein}}(\rho^N|\pi) ] & = \mb{E}_{\underline{x}\sim {p^N_{0,K}}} [ I_{\nu_{n+1},\text{Stein}}(\rho_n^N|\pi) ] \\
    &\ge I_{\nu_{n+1},\text{Stein}}( \mb{E}_{\underline{x}\sim {p^N_{0,K}}}[\rho_n^N]|\pi)  = I_{\nu_{n+1},\text{Stein}}( {\mu^N_{n,K}}|\pi) .
\end{align*}
Therefore,
\begin{align*}
     \KL({p^N_{n+1,K}}|\pi^{\otimes N})&\le \KL({p^N_{n,K}}|\pi^{\otimes N}) - \frac{1}{2}h_{n+1} N I_{\nu_{n+1},\text{Stein}}( {\mu^N_{n,K}}|\pi) + c^* d h_{n+1}  \\
    &\quad +h_{n+1}\beta_{1,n+1}(1-\nu_{n+1}) d + \frac{1}{2}\beta_{2,n+1}(1-\nu_{n+1}) B^{-2}N^{-\frac{1}{2}}\theta \nu_{n+1}^3 +\frac{1}{8}\nu_{n+1}^{2} \theta^2.
\end{align*}
Since ${\mu^N_{n,K}}$ is smooth, $I({\mu^N_{n,K}} | \pi)$ is well-defined and we have by \cite[Equation (12)]{he2024regularized},
\begin{align*}
  I({\mu^N_{n,K}} | \pi) - (1-\nu_{n+1}) I_{\nu_{n+1},\text{Stein}}( {\mu^N_{n,K}}|\pi) \le \nu_{n+1}^{2\gamma_n} (1-\nu_{n+1})^{-2\gamma_n} R_n^2 ,  
\end{align*}
which implies that
\begin{align*}
     \KL({p^N_{n+1,K}}|\pi^{\otimes N})&\le \KL({p^N_{n,K}}|\pi^{\otimes N}) - \frac{h_{n+1}}{2(1-\nu_{n+1})} N I({\mu^N_{n,K}} | \pi) + \frac{1}{2}h_{n+1}N \nu_{n+1}^{2\gamma_n} (1-\nu_{n+1})^{-2\gamma_n-1} R_n^2\\
     &\quad + h_{n+1} c^* d + h_{n+1}\beta_{1,n+1}(1-\nu_{n+1}) d  + \frac{1}{2}\beta_{2,n+1}(1-\nu_{n+1}) B^{-2}N^{-\frac{1}{2}}\theta \nu_{n+1}^3 +\frac{1}{8}\nu_{n+1}^{2} \theta^2.
\end{align*}
Using the telescope sum and we get
\begin{align*}
    \frac{1}{T}\sum_{n=1}^T \frac{h_n}{1-\nu_n} I({\mu^N_{n-1,K}} | \pi) &\le \frac{2\KL({p^N_{0,K}}|\pi^{\otimes N})}{NT} +\frac{2 c^* d \sum_{n=1}^T  h_n}{NT} + \frac{2 d \sum_{n=1}^T \beta_{1,n}(1-\nu_{n}) h_n}{NT}  \\
    & \quad + \frac{\theta \sum_{n=1}^T \beta_{2,n} (1-\nu_{n}) \nu_{n}^3 }{B^2 N^{\frac{3}{2}} T } + \frac{\theta^2 \sum_{n=1}^T\nu_n^2 }{4NT} + \frac{1}{T}\sum_{n=1}^T \frac{h_n}{1-\nu_n} (\frac{\nu_n}{1-\nu_n})^{2\gamma_{n-1}}R_{n-1}^2 .
\end{align*}
Last, according to \cite{balasubramanian2024improved}, $\KL({p^N_{0,K}}|\pi^{\otimes K})\le 2 \KL(p_0^N|\pi^{\otimes N})+\log 2 \le \gamma d N$ where $\gamma\coloneqq 2C_{\KL}+\log 2$. Therefore,
\begin{align*}
      \frac{1}{T}\sum_{n=1}^T \frac{h_n}{1-\nu_n} I({\mu^N_{n-1,K}} | \pi) &\le \frac{2\gamma d }{T} + \frac{2 c^* d \sum_{n=1}^T  h_n}{NT} +\frac{2 d \sum_{n=1}^T \beta_{1,n}(1-\nu_{n}) h_n}{NT} + \frac{\theta \sum_{n=1}^T \beta_{2,n} (1-\nu_{n})\nu_{n}^3 }{B^2 N^{\frac{3}{2}} T } \\
      &\quad + \frac{\theta^2 \sum_{n=1}^T\nu_n^2 }{4NT}+  \frac{1}{T}\sum_{n=1}^T \frac{h_n}{1-\nu_n} (\frac{\nu_n}{1-\nu_n})^{2\gamma_{n-1}}R_{n-1}^2.
\end{align*}
The first bound in Theorem \ref{thm:decay of fisher} follows from the assumptions on $\{\nu_n\}_{n\ge 1}$. Last, if $\pi$ satisfies $\mathrm{W}_1\mathrm{I}$ with parameter $C_\pi$, according to convexity of $I(\cdot|\pi)$, we have
\begin{align*}
    \mathrm{W}_1 (\mu_{T,av}^N,\pi)^2 \le C_\pi^2 I(\mu_{T, av}^N|\pi) \le \frac{C_\pi^2}{T}\sum_{n=1}^T \frac{h_n}{1-\nu_n} I(\mu^N_{n-1,K}|\pi) .
\end{align*}
\end{proof}

\begin{lemma}\label{lem:upper bound psi 1} Under Assumption \ref{assump:convergence}, we have $$\int \| \Tm_{n+1}(\underline{x}) \|^2 {p_n^N}(\underline{x})\dee x \le \frac{B}{(1-\nu_{n+1})B+\nu_{n+1}}\mb{E}_{\underline{x}\sim {p_n^N}}[I_{\nu_{n+1},\text{Stein}}(\rho^N|\pi)].$$    
\end{lemma}
\namedproof{Proof of Lemma \ref{lem:upper bound psi 1}.} According to the definition of $\Tm_{n+1}$,
\begin{align*}
    &\quad \int \| \Tm_{n+1}(\underline{x}) \|^2 {p_n^N}(\underline{x})\dee x \\
    &= \frac{1}{N^2}\int \sum_{i=1}^N \| \sum_{l=1}^N \big( \frac{1-\nu_{n+1}}{N}K(\underline{x})+\nu_{n+1}I_N \big)^{-1}_{i,l} \big( \sum_{j=1}^N k(x^l,x^j)\nabla V(x^j)-\nabla_2 k(x^l,x^j) \big) \|^2 {p_n^N}(\underline{x})\dee \underline{x} \\
    & = \mb{E}_{\underline{x}\sim {p_n^N}}\big[ \| \big( (1-\nu_{n+1}) \iota^*_{k,\rho^N}\iota_{k,\rho^N}+\nu_{n+1}I_d \big)^{-1} \iota^*_{k,\rho^N}\iota_{k,\rho^N}\nabla\log \frac{\rho^N}{\pi}(\underline{x})\|^2 \big] \\
    & \le {\frac{\| \iota^*_{k,\rho^N}\iota_{k,\rho^N} \|_{L^2(\rho^N)}}{(1-\nu_{n+1})\| \iota^*_{k,\rho^N}\iota_{k,\rho^N} \|_{L^2(\rho^N)}+\nu_{N+1}}} \mb{E}_{\underline{x}\sim {p_n^N}} [ I_{\nu_{n+1},\text{Stein}}(\rho^N|\pi) ]\\
      &   \le {\frac{B}{(1-\nu_{n+1})B+\nu_{n+1}}} \mb{E}_{\underline{x}\sim {p_n^N}} [ I_{\nu_{n+1},\text{Stein}}(\rho^N|\pi) ],
\end{align*}
where the second identity follows from the equivalence between the matrix-form and the operator-form of the regularized SVGD and the last inequality follows from the spectral decomposition definition of $I_{\nu_{n+1},\text{Stein}}(\rho^N|\pi)$ and properties of the integral operator $\iota^*_{k,\rho^N}$. See Appendix \ref{app:RKHS} for more details.
\end{proof}

\begin{lemma}\label{lem:upper bound psi 2} 
    Under Assumption \ref{assump:convergence}, we have
    \begin{align*}
        \int \| J\Tm_{n+1}(\underline{x}) \|_{\HS}^2 p_{n}^N(\underline{x})\dee x &\le 2Bd(1-\nu_{n+1})^2\nu_{n+1}^{-3}\mb{E}_{\underline{x}\sim {p_n^N}}[I_{\nu_{n+1},\text{Stein}}(\rho^N|\pi)] \\
        &\quad + \nu_{n+1}^{-2}A^2 B^2 d (2N+3) {\mb{E}_{\underline{x}\sim {p_n^N}} [\big( \frac{1}{N} \sum_{i=1}^N V(x^i)\big)^{2\alpha}]}\\
        &\quad + \nu_{n+1}^{-2}B^2 d^2 (2N+3) + 3\nu_{n+1}^{-2} B^2 C_V^2 d.
    \end{align*}    
\end{lemma}
\namedproof{Proof of Lemma \ref{lem:upper bound psi 2}.}
    If we denote $\Tm_n=(\Tm_n^{1},\cdots, \Tm_n^{N})^\intercal$ and $\Tm_n^{ik}$ the $k^{th}$ coordinate in $\Tm_n^{i}$. Then $J\Tm_n(\underline{x})=\big( \partial_{jl}\Tm_n^{ik}(\underline{x}) \big)_{j,i\in [N],l,k\in [d]}$, where $\partial_{jl}$ denoting the partial derivative against the $l^{th}$ coordinate of $x^j$ (denoted as $x^{jl}$). Now we compute each entry in the Jacobian:
\begin{align*}
    &\quad \partial_{jl}\Tm_{n+1}^{ik}(\underline{x}) \\
    &= \partial_{jl}\big[ \frac{1}{N}\sum_{r_1=1}^N \big( \frac{1-\nu_{n+1}}{N}K(\underline{x})+\nu_{n+1}I_N \big)^{-1}_{i,r_1}\sum_{r_2=1}^N k(x^{r_1},x^{r_2})\partial_k V(x^{r_2})-\partial_{2,k}k(x^{r_1},x^{r_2}) \big] \\
    & = \underbrace{\frac{1}{N}\sum_{r_1=1}^N \partial_{jl}\big( K^{\nu_{n+1}}(\underline{x}) \big)^{-1}_{i,r_1}\sum_{r_2=1}^N k(x^{r_1},x^{r_2})\partial_k V(x^{r_2})-\partial_{2,k}k(x^{r_1},x^{r_2})}_{\coloneqq S_0(i,j,k,l)} \\
    &\quad +  \underbrace{\frac{1}{N}\big( K^{\nu_{n+1}}(\underline{x})\big)^{-1}_{i,j}\sum_{r_2=1}^N \partial_{1,l}k(x^{j},x^{r_2})\partial_k V(x^{r_2})-\partial_{1,l}\partial_{2,k}k(x^{j},x^{r_2})}_{\coloneqq S_1(i,j,k,l)} \\
    & \quad + \underbrace{\frac{1}{N}\sum_{r_1=1}^N \big( K^{\nu_{n+1}}(\underline{x})\big)^{-1}_{i,r_1} \big(\partial_{2,l}k(x^{r_1},x^{j})\partial_k V(x^{j})+k(x^{r_1},x^j)\partial^2_{k,l}V(x^j)-\partial^2_{2,k,l}k(x^{r_1},x^{j}) \big)}_{\coloneqq S_2(i,j,k,l)},
\end{align*}
where $K^{\nu_{n+1}}(\underline{x})\coloneqq \frac{1-\nu_{n+1}}{N}K(\underline{x})+\nu_{n+1}I_N$. $\partial_{1,l}k(\cdot,\cdot)$/$\partial_{2,l}k(\cdot,\cdot)$ denotes taking partial derivative wrt. the $l^{th}$ coordinate of the first/second variable in $k$ and $\partial^2_{2,k,l}k(\cdot,\cdot)$ denotes taking the second order partial derivative wrt. the $l^{th}$ and $k^{th}$ coordinates of the second variable in $k$. 

\vspace{.2cm}

\noindent Next, we bound the three terms $S_0,S_1$ and $S_2$ respectively.

\noindent For $S_0(i,j,k,l)$, notice that 
\begin{align*}
    \partial_{jl} (K^{\nu_{n+1}}(\underline{x}))^{-1} = -\frac{1-\nu_{n+1}}{N}(K^{\nu_{n+1}}(\underline{x}))^{-1} \partial_{jl}K(\underline{x}) (K^{\nu_{n+1}}(\underline{x}))^{-1}.
\end{align*}
In particular, for all $s_1, s_2\in [N]$, $$(\partial_{jl} K(\underline{x}))_{s_1,s_2} = 1_{\{s_1=j\}} \partial_{1,l}k(x^j,x^{s_2}) + 1_{\{s_2=j\}}\partial_{2,l} k(x^{s_1},x^j).$$ Hence $\partial_{jl}K(\underline{x})$ is sparse (with $(2N-1)$ non-zero bounded entries) and $$\|\partial_{jl}K(\underline{x})\|_{\mathrm{op}}\le \|\partial_{jl}K(\underline{x})\|_{\HS}\le \sqrt{2N}B.$$ Based on this estimation, we have
\begin{align*}
    &\quad \sum_{i,j,k,l} |S_0(i,j,k,l)|^2 \\
    &= \sum_{i,j,k,l}\big| \frac{1}{N} \sum_{r_1=1}^N \partial_{jl}\big((K^{\nu_{n+1}})^{-1}\big)_{i,r_1}\sum_{r_2=1}^N k(x^{r_1},x^{r_2})\partial_k V(x^{r_2})-\partial_{2,k}k(x^{r_1},x^{r_2})\big|^2 \\
    & = \sum_{i,j,k,l}\big| \frac{1-\nu_{n+1}}{N^2} \sum_{r_1=1}^N \big((K^{\nu_{n+1}})^{-1} \partial_{jl}K (K^{\nu_{n+1}})^{-1}\big)_{i,r_1}\sum_{r_2=1}^N k(x^{r_1},x^{r_2})\partial_k V(x^{r_2})-\partial_{2,k}k(x^{r_1},x^{r_2})\big|^2 \\
     &\le \frac{(1-\nu_{n+1})^2}{N^4}\sum_{i,j,k,l} \|   (K^{\nu_{n+1}})^{-1} \partial_{jl} K \|_{\mathrm{op}}^2 \big| \sum_{r_1=1}^N (K^{\nu_{n+1}})^{-1}_{i,r_1} \sum_{r_2=1}^N k(x^{r_1},x^{r_2})\partial_k V(x^{r_2})-\partial_{2,k}k(x^{r_1},x^{r_2})\big|^2 \\
    &\le \frac{(1-\nu_{n+1})^2}{N^4}\sum_{i,j,k,l} \| \partial_{jl} K \|_{\HS}^2 \|  (K^{\nu_{n+1}})^{-1}\|_{\mathrm{op}}^2 \big| \sum_{r_1=1}^N (K^{\nu_{n+1}})^{-1}_{i,r_1} \sum_{r_2=1}^N k(x^{r_1},x^{r_2})\partial_k V(x^{r_2})-\partial_{2,k}k(x^{r_1},x^{r_2})\big|^2 \\
    &\le  \frac{(1-\nu_{n+1})^2}{N^4} \frac{2N B^2}{\nu_{n+1}^2} N d \sum_{i=1}^N \| \sum_{r_1=1}^N (K^{\nu_{n+1}})^{-1}_{i,r_1} \sum_{r_2=1}^N k(x^{r_1},x^{r_2})\nabla V(x^{r_2})-\nabla_2 k(x^{r_1},x^{r_2})\|^2\\
    &= 2B^2 d \nu_{n+1}^{-2}(1-\nu_{n+1})^2 \| (K^{\nu_{n+1}})^{-1} \big( \frac{1}{N}\sum_{r_2=1}^N k(\underline{x},x^{r_2})\nabla V(x^{r_2})-\nabla_2 k(\underline{x},x^{r_2}) \big) \|^2,
    \end{align*}
    where we hide the dependence of $K^{\nu_{n+1}}$ and $\partial_{jl}K$ on $\underline{x}$ before the second line to hide notational overload.
   Taking the expectation, we have 
    \begin{align*}
        &\quad \int \sum_{i,j,k,l}|S_0(i,j,k,l)|^2 {p_n^N}(\underline{x})\dee x\\
        &\le 2B^2 (1-\nu_{n+1})^2\nu_{n+1}^{-2} d \int \| \big( (1-\nu_{n+1}) \iota^*_{k,\rho^N}\iota_{k,\rho^N}+\nu_{n+1}I_d \big)^{-1} \iota^*_{k,\rho^N}\iota_{k,\rho^N}\nabla\log \frac{\rho^N}{\pi}(\underline{x})\|^2 {p_n^N}(\underline{x})\dee \underline{x} \\
        &\le 2(1-\nu_{n+1})^2 \nu_{n+1}^{-3} B^2 d \mb{E}_{\underline{x}\sim {p_n^N}}[I_{\nu_{n+1},\text{Stein}}(\rho^N|\pi)],
    \end{align*}
    where the last inequality follows from the spectral decomposition of $I_{\nu_{n+1},\text{Stein}}(\rho^N|\pi)]$ in $L^2(\rho^N)$.

    \vspace{.2cm}

   \noindent For $S_1(i,j,k,l)$, we have
   \begin{align*}
       &\quad \sum_{i,j,k,l} |S_1(i,j,k,l)|^2 \\
       &= \sum_{i,j,k,l}\big| \frac{1}{N}\big( K^{\nu_{n+1}}(\underline{x}) \big)^{-1}_{i,j}\sum_{r_2=1}^N \partial_{1,l}k(x^{j},x^{r_2})\partial_k V(x^{r_2})-\partial_{1,l}\partial_{2,k}k(x^{j},x^{r_2})\big|^2 \\
       &\le  \frac{1}{N^2} \sum_{k,l} \| \big( K^{\nu_{n+1}}(\underline{x}) \big)^{-1} \|_{\mathrm{F}}^2 \max_{j} \big( \sum_{r_2=1}^N \partial_{1,l}k(x^{j},x^{r_2})\partial_k V(x^{r_2})-\partial_{1,l}\partial_{2,k}(x^{j},x^{r_2})\big)^2 \\
       &\le \frac{2}{N^2}\frac{N}{\nu_{n+1}^2}  \big( B^2  d  N\sum_{r_2=1}^N \| \nabla V(x^{r_2})\|^2 +  B^2 d^2 N^2\big) \\
       &\le 2\nu_{n+1}^{-2}A^2B^2 N d \big( \frac{1}{N} \sum_{i=1}^N V(x^i)\big)^{2\alpha} + 2\nu_{n+1}^{-2}B^2 N d^2, 
   \end{align*}
   where the last identity follows from Assumption \ref{assump:convergence}-(b) and Jensen's inequality.
   
    \vspace{.2cm}

    \noindent Last, for $S_2(i,j,k,l)$, we have
   \begin{align*}
       &\quad \sum_{i,j,k,l} |S_2(i,j,k,l)|^2 \\
       &= \sum_{i,j,k,l}\big| \frac{1}{N}\sum_{r_1=1}^N \big( K^{\nu_{n+1}}(\underline{x}) \big)^{-1}_{i,r_1} \big(\partial_{2,l}k(x^{r_1},x^{j})\partial_k V(x^{j})+k(x^{r_1},x^j)\partial^2_{k,l}V(x^j)-\partial^2_{2,k,l}k(x^{r_1},x^{j}) \big)\big|^2 \\
       &\le \| \big( K^{\nu_{n+1}}(\underline{x}) \big)^{-1} \|_2^2 \sum_{i,j,k,l}\big| \frac{1}{N}  \big(\partial_{2,l}k(x^{i},x^{j})\partial_k V(x^{j})+k(x^{i},x^j)\partial^2_{k,l}V(x^j)-\partial^2_{2,k,l}k(x^{i},x^{j}) \big)\big|^2 \\
       &\le 3\nu_{n+1}^{-2}\frac{1}{N^2}\bigg(  \sum_{j} Nd  B^2  \|\nabla V(x^{j})\|^2 + B^2 N^2 \big(\sup_x\| H_V(x) \|_{\HS}^2\big) + B^2 N^2 d^2 \bigg) \\
       &\le 3\nu_{n+1}^{-2}A^2B^2 d \big( \frac{1}{N} \sum_{i=1}^N V(x^i)\big)^{2\alpha} + 3\nu_{n+1}^{-2}  B^2 C_V^2  d+ 3\nu_{n+1}^{-2} B^2  d^2. 
   \end{align*}
\end{proof}
\begin{remark} Lemma \ref{lem:upper bound psi 1} and Lemma \ref{lem:upper bound psi 2} are adapted versions of \cite[Lemma 4]{balasubramanian2024improved}. We emphasize (1) our upper bounds depend on the regularized KSD in expectation, which reflect the geometry of R-SVGD and (2) our bounds reduce to those in \cite[Lemma 4]{balasubramanian2024improved} when $\nu_{n+1}=1$.   
\end{remark}
\begin{lemma}\label{lem:Jacobian pointwise bound} Let $\{X^i_n\}_{i\in [N], n\ge 0}$ denotes the sequence of $N$ particles along the R-SVGD \eqref{eq:algorithm}. Under Assumption \ref{assump:convergence}, if $\{(h_n, \nu_n)\}_{n\ge 0}$ satisfies that for all $n\ge 0$:
\begin{align*}
    h_n\le \nu_n/B,\quad B h_{n+1}\nu_{n+1}^{-1}\le \frac{1}{16(1-\alpha)^2 C_V A^2 \sum_{l=1}^n B h_l \nu_l^{-1}} ,
\end{align*}
We have that for all $n\ge 1$:
     \begin{align}\label{eq:upper bound Jacobian norm}
        \| J\Tm_{n+1}(\underline{X}_n) \|_{\HS}^2 \le  &  9\nu_{n+1}^{-4}A^2B^4 N d \bigg[ M^{2\alpha} \bigg(d^{\frac{1}{1-\alpha}}+\frac{1}{N}\sum_{i=1}^N V(x^i(0))\bigg)^{2\alpha} \big[ (\sum_{l=1}^n h_l\nu_l^{-1})^{\frac{2\alpha}{1-\alpha}} \vee 1\big] \bigg] \nonumber \\
        & +9\nu_{n+1}^{-4} B^4 N d^2+ 3\nu_{n+1}^{-2}B^2 C_V^2 d.
    \end{align}
\end{lemma}
\begin{remark}
    Lemma \ref{lem:Jacobian pointwise bound} is an analogue to \cite[Lemma 4 part 2]{balasubramanian2024improved}. This result is necessary since we need to pointwisely bound the term $\| (\id - t J\Tm_{n+1}(\underline{x}) )^{-1} \|_{2}^2$ in \eqref{eq:term psi 2}. Compared to Lemma \ref{lem:upper bound psi 2}, the order of $\nu$ increases by $2$.
\end{remark} 
\namedproof{Proof of Lemma \ref{lem:Jacobian pointwise bound}.} According to the proof of Lemma \ref{lem:upper bound psi 2}, we pointwisely bound $\sum_{i,j,k,l}|S_0(i,j,k,l)|^2$. For simplicity, we denote $\underline{X}_n$ by $\underline{x}$ in the proof. We have
\begin{align*}
    &\quad \sum_{i,j,k,l} |S_0(i,j,k,l)|^2 \\
    &\le 2B^2 d \nu_{n+1}^{-2}(1-\nu_{n+1})^2 \frac{1}{N^2} \| (K^{\nu_{n+1}}(\underline{x}))^{-1} \big( \sum_{r_2=1}^N k(\underline{x},x^{r_2})\nabla V(x^{r_2})-\nabla_2 k(\underline{x},x^{r_2}) \big) \|^2 \\
    &\le 2B^2d\nu_{n+1}^{-4}(1-\nu_{n+1})^2 \frac{1}{N^2} \sum_{r_1=1}^N\|   \sum_{r_2=1}^N k(x^{r_1},x^{r_2})\nabla V(x^{r_2})-\nabla_2 k(x^{r_1},x^{r_2})\|^2 \\
    &\le 4B^2d\nu_{n+1}^{-4}(1-\nu_{n+1})^2 \big( A^2B^2N (\frac{1}{N}\sum_j V(x^j))^{2\alpha}+NB^2d \big)\\
    & = 4 A^2 B^4 \nu_{n+1}^{-4}(1-\nu_{n+1})^2 N d  \big(\frac{1}{N}\sum_j V(x^j)\big)^{2\alpha}+4B^4 \nu_{n+1}^{-4}(1-\nu_{n+1})^2 N d^2
    \end{align*}
    where the last inequality follows from Assumption \ref{assump:convergence} and Jensen's inequality. Combined with the upper bounds of $\sum_{i,j,k,l}|S_1(i,j,k,l)|^2$ and $\sum_{i,j,k,l}|S_2(i,j,k,l)|^2$ derived in the proof of Lemma \ref{lem:upper bound psi 2}, we have
    \begin{align*}
        \| J\Tm_{n+1}(\underline{x}) \|_{\HS}^2\le &\nu_{n+1}^{-2}A^2B^2 d \big( 4B^2 (\nu_{n+1}^{-1}-1)^2N +2N  +3\big)\big(\frac{1}{N}\sum_j V(x^j)\big)^{2\alpha}\\
        & + \nu_{n+1}^{-2} B^2 d^2 \big(  4B^2(\nu_{n+1}^{-1}-1)^2 N + 2N +3 \big)+ 3\nu_{n+1}^{-2}B^2 C_V^2 d \\
        \le & 9\nu_{n+1}^{-4}A^2B^4 N d \big(\frac{1}{N}\sum_j V(x^j)\big)^{2\alpha}+9\nu_{n+1}^{-4} B^4 N d^2+ 3\nu_{n+1}^{-2}B^2 C_V^2  d.
    \end{align*}   
The bound in the lemma now follows from Lemma \ref{lem:trajectory bound of V empirical expectation}.
\end{proof}
\begin{lemma}\label{lem:trajectory bound of V empirical expectation} Under assumptions in Lemma \ref{lem:Jacobian pointwise bound}, we have for all $n\ge 1$,
 \begin{align}
     \frac{1}{N}\sum_{i=1}^N V(X^i_n) \le M \left(d^{\frac{1}{1-\alpha}}+\frac{1}{N}\sum_{i=1}^N V(X^i_0)\right) \big[ (\sum_{l=1}^n h_l\nu_l^{-1})^{\frac{1}{1-\alpha}} \vee 1\big],
 \end{align}
 where $M$ is a positive constant only depending on constants appearing in Assumption \ref{assump:convergence} (independent of $N,d$).
\end{lemma}
\namedproof{Proof of Lemma \ref{lem:trajectory bound of V empirical expectation}.} For all $n\ge 0$,
\begin{align}\label{eq:equ 1}
    &  \frac{1}{N}\sum_{i=1}^N V(X^i_{n+1}) - \frac{1}{N}\sum_{i=1}^N V(X^i_n) \nonumber\\
    = & \frac{1}{N} \sum_{i=1}^N \langle \nabla V(X^i_n) , X^i_{n+1} - X^i_n  \rangle \nonumber\\
    & + \frac{1}{N}\sum_{i=1}^N \int_0^1 (1-s) \langle X^i_{n+1}-X^i_n , H_V\big( X^i_n - s h_{n+1} \Tm_{n+1}^{i}(X^i_n)  \big) (X^i_{n+1}-X^i_n) \rangle \dee s \nonumber\\
    \le & \frac{1}{N} \sum_{i=1}^N \langle \nabla V(X^i_n) , X^i_{n+1} - X^i_n  \rangle + \frac{C_V h_{n+1}^2}{2N} \| \Tm_{n+1}(\underline{X}_n) \|^2,
\end{align}
where
\begin{align*}
    &\quad \| \Tm_{n+1}(\underline{X}_n) \|^2 \\
    &= \frac{1}{N^2}\sum_{i=1}^N \| \sum_{l=1}^N \big( K^{\nu_{n+1}}(\underline{X}_n)\big)^{-1}_{i,l} \big( \sum_{j=1}^N k(X^l_n,X^j_n)\nabla V(X^j_n)-\nabla_2 k(X^l_n,X^j_n) \big) \|^2 \\
    &\le \frac{2}{N^2}\sum_{i=1}^N \| \sum_{l,j=1}^N \big( K^{\nu_{n+1}}(\underline{X}_n) \big)^{-1}_{i,l}  k(X^l_n,X^j_n)\nabla V(X^j_n) \|^2  + \frac{2}{N^2}\sum_{i=1}^N \| \sum_{l,j=1}^N \big( K^{\nu_{n+1}}(\underline{X}_n) \big)^{-1}_{i,l} \nabla_2 k(X^l_n,X^j_n)  \|^2 \\
    & = \underbrace{\frac{2}{N^2} \| \big( K^{\nu_{n+1}}(\underline{X}_n)\big)^{-1} K(\underline{X}_n) {\nabla V}(\underline{X}_n)  \|_{\mathrm{F}}^2}_{\coloneqq \text{\rom{1}}}  + \underbrace{\frac{2}{N^2} \| \big( K^{\nu_{n+1}}(\underline{X}_n) \big)^{-1} \sum_{j=1}^N\nabla_2 k(\underline{X}_n,X^j_n)  \|_{\mathrm{F}}^2}_{\coloneqq\text{\rom{2}}} .
\end{align*}
For \rom{1}, we have 
\begin{align*}
\mathrm{\rom{1}}   &\le 2N^{-2}\| \big( K^{\nu_{n+1}}(\underline{X}_n)\big)^{-1} K(\underline{X}_n) \|_{2}^2 \|{\nabla V}(\underline{X}_n)  \|_{\mathrm{F}}^2\\
    &\le 2N^{-2}\big( \frac{1-\nu_{n+1}}{N} \| K(\underline{X}_n) \|_{2} + \nu_{n+1} \big)^{-2} \| K(\underline{X}_n) \|_{2}^2 \|{\nabla V}(\underline{X}_n)  \|_{\mathrm{F}}^2\\
    &\le  2N^{-2}\big( \frac{1-\nu_{n+1}}{N} \| K(\underline{X}_n) \|_{\mathrm{F}} + \nu_{n+1} \big)^{-2} \| K(\underline{X}_n) \|_{\mathrm{F}}^2 \|{\nabla V}(\underline{X}_n)  \|_{\mathrm{F}}^2 \\
    &\le  2\big( B (1-\nu_{n+1}) + \nu_{n+1} \big)^{-2} B^2 \|{\nabla V}(\underline{X}_n)  \|_{\mathrm{F}}^2 \\
    & = 2\big( B (1-\nu_{n+1}) + \nu_{n+1} \big)^{-2} B^2\sum_{i=1}^N \| \nabla V(X^i_n) \|^2\\
    &\le 2A^2\big( B (1-\nu_{n+1}) + \nu_{n+1} \big)^{-2} B^2 N \big(\frac{1}{N} \sum_{i=1}^N V(X^i_n) \big)^{2\alpha},
\end{align*}
where the last inequality follows from Assumption \ref{assump:convergence} and Jensen's inequality, noting $\alpha\in [0,1/2]$.

\vspace{.2cm}

\noindent For \rom{2}, we have
\begin{align*}
  \mathrm{\rom{2}}  &\le 2N^{-2}\| \big( K^{\nu_{n+1}}(\underline{X}_n)\big)^{-1} \|_2^2 \| \sum_{j=1}^N\nabla_2 k(\underline{X}_n,X^j_n)  \|_{\mathrm{F}}^2 \le 2\nu_{n+1}^{-2} N d B^2. 
\end{align*}
Combine the above two estimations, we get
\begin{align*}
    \| \Tm_{n+1}(\underline{X}_n) \|^2 
    &\le 2A^2\big( B (1-\nu_{n+1}) + \nu_{n+1} \big)^{-2} B^2 N \big(\frac{1}{N} \sum_{i=1}^N V(X^i_n) \big)^{2\alpha} +2\nu_{n+1}^{-2} N d B^2.
\end{align*}
For the other term in \eqref{eq:equ 1}, we have
\begin{align*}
    &\quad \frac{1}{N} \sum_{i=1}^N \langle \nabla V(X^i_n) , X^i_{n+1} - X^i_n  \rangle  \\
    & = -\frac{h_{n+1}}{N^2} \sum_{i=1}^N \langle \nabla V(X^i_n) , \sum_{l=1}^N \big( K^{\nu_{n+1}}(\underline{X}_n)\big)^{-1}_{i,l} \big( \sum_{j=1}^N k(X^l_n,X^j_n)\nabla V(X^j_n)-\nabla_2 k(X^l_n,X^j_n) \big)  \rangle \\
    & = -\frac{h_{n+1}}{N^2} \sum_{i,j=1}^N \langle \nabla V(X^i_n) , \left(\big( K^{\nu_{n+1}}(\underline{X}_n) \big)^{-1} K(\underline{X}_n) \right)_{i,j}{\nabla V}(X^j_n) \rangle \\
    &\quad + \frac{h_{n+1}}{N^2} \sum_{i,l=1}^N \langle \nabla V(X^i_n) , \left(\big( K^{\nu_{n+1}}(\underline{X}_n) \big)^{-1}\right)_{i,l}  \big(\sum_{j=1}^N \nabla_2 k(X^l_n,X^j_n) \big) \rangle \\
    &\le \frac{h_{n+1}}{N^2} \sum_{i=1}^N   \|  \nabla V(X^i_n) \| \cdot \| \big( K^{\nu_{n+1}}(\underline{X}_n) \big)^{-1} \|_{2}\| \sum_{j=1}^N \nabla_2 k(X^i_n,X^j_n)  \| \\
    &\le {h_{n+1}} \nu_{n+1}^{-1} B\sqrt{d} A \big(\frac{1}{N} \sum_{i=1}^N V(X^i_n) \big)^{\alpha},
\end{align*}
where the last inequality follows from Assumption \ref{assump:convergence} and Jensen's inequality.
For simplicity, we denote $f_n\coloneqq \frac{1}{N}\sum_{i=1}^N V(X^i_n)$. Then, with the above estimates, we have
\begin{align*}
      f_{n+1} - f_n \le  {h_{n+1}} \nu_{n+1}^{-1} B A \sqrt{d} f_n^{\alpha}  + {C_V h_{n+1}^2}A^2\big( B (1-\nu_{n+1}) + \nu_{n+1} \big)^{-2} B^2 f_n^{2\alpha} + {C_V h_{n+1}^2}\nu_{n+1}^{-2}B^2 d.
\end{align*}
If we further denote $\eta_{n+1}\coloneqq B h_{n+1} /\nu_{n+1}$, since $\alpha\in (0,1/2)$, apply that $f_n^{2\alpha}\le 1+f_n$ and we get
\begin{align*}
    f_{n+1}\le (1 + C_V A^2 \eta_{n+1}^2) f_n  + A \sqrt{d} \eta_{n+1} f_n^\alpha +  C_V (A^2+d) \eta_{n+1}^2.
 \end{align*}
 We will harness the above recursive bound to obtain the claimed bound in the lemma. First, we handle the case $0<n\le n^*$ with $n^*\coloneqq \sup\{ n\ge 0 : \sum_{i=1}^n \eta_i\le 1 \}$. Fix $L>0$, define
 \begin{align*}
     \tau_L \coloneqq \sup \{ n\ge 0 : f_n\le L \} \wedge n^*.
 \end{align*}
 Then for all $1\le n\le \tau_L$, we have $\sup_{1\le n\le n^*} \eta_n\le 1$, and therefore
 \begin{align*}
     f_n  & \le \sum_{l=1}^n \big( A \sqrt{d} \eta_l L^\alpha + C_V(A^2+d)\eta_l^2 \big)\prod_{r=l+1}^{n} (1+C_V A^2 \eta_r^2)  + f_0\prod_{l=1}^n (1+C_V A^2 \eta_l^2)\\
     &\le \sum_{l=1}^n \big( A \sqrt{d} \eta_l L^\alpha + C_V (A^2+d)\eta_l^2 \big)\exp\big( C_V A^2 \sum_{r=l+1}^{n} \eta_r^2 \big)  + f_0 \exp\big( C_V A^2 \sum_{l=1}^{n-1} \eta_l^2 \big) \\
     &\le \exp(C_V A^2) ( A \sqrt{d} L^\alpha + +C_V A^2+ C_V d ) + \exp(C_V A^2) f_0.
 \end{align*}
 Hence, taking $L=\hat{M}(d+f_0)$ for some suitable $\hat{M}\ge 1$ depending only on constants in Assumption \ref{assump:convergence} (independent to $d$ and $f_0$), we conclude from the above bound that $f_{\tau_L}<L$ and hence $\tau_L = n^*$. Thus, $f_n\le \hat{M}(d+f_0)$ for all $0\le n\le n^*$. Now we handle the case $n^*+1\le n\le T$. We proceed by induction. Let 
 \begin{align*}
     \beta \coloneqq M \big( d^{\frac{1}{1-\alpha}} + f_0\big), \quad \text{where } M=\hat{M} \vee [16(1-\alpha)^2(A+C_V A^2 + C_V)]^{\frac{1}{1-\alpha}}.
 \end{align*}
  Then, by the above bound, note that for $n=n^*+1$, $f_n\le \hat{M}(d+f_0)\le \beta (\sum_{i=1}^n \eta_i )^{\frac{1}{1-\alpha}}$. Suppose that for some $n^*+1\le n\le T$, $f_n\le \beta (\sum_{i=1}^n \eta_i )^{\frac{1}{1-\alpha}}$, taking any $\eta_{n+1} \le 1 \wedge \tfrac{1}{16(1-\alpha)^2 C_V A^2 \sum_{i=1}^n \eta_i } $, then we have
 \begin{align*}
     f_{n+1} & \le (1+C_V A^2 \eta_{n+1}^2) \beta (\sum_{i=1}^n \eta_i )^{\frac{1}{1-\alpha}} + A\sqrt{d}\eta_{n+1} \beta^\alpha (\sum_{i=1}^n \eta_i )^{\frac{\alpha}{1-\alpha}} + C_V (A^2+d) \eta_{n+1}^2 \\
     &=  \beta (\sum_{i=1}^{n+1} \eta_i )^{\frac{1}{1-\alpha}}\left[ (1+C_V A^2 \eta_{n+1}^2)\left( \frac{\sum_{i=1}^n \eta_i }{\sum_{i=1}^{n+1} \eta_i } \right)^{\frac{1}{1-\alpha}} + A\sqrt{d}\beta^{-(1-\alpha)} \frac{(\sum_{i=1}^n \eta_i)^{\frac{\alpha}{1-\alpha}}\eta_{n+1} }{(\sum_{i=1}^{n} \eta_i+\eta_{n+1})^{\frac{1}{1-\alpha}} } \right.\\
     &\qquad\qquad\qquad\qquad\left. + C_V (A^2+d) \beta^{-1} \frac{\eta_{n+1}^2}{(\sum_{i=1}^{n} \eta_i+\eta_{n+1})^{\frac{1}{1-\alpha}} } \right] \\
     &\le \beta (\sum_{i=1}^{n+1} \eta_i )^{\frac{1}{1-\alpha}}\left[ (1+C_V A^2 \eta_{n+1}^2)-\left\{ \frac{\eta_{n+1}}{8(1-\alpha)^2\sum_{i=1}^n \eta_i} - \frac{A\sqrt{d}+C_V(A^2+d)}{\beta^{1-\alpha}}\frac{\eta_{n+1}}{\sum_{i=1}^n \eta_i} \right\} \right]\\
     &\le \beta (\sum_{i=1}^{n+1} \eta_i )^{\frac{1}{1-\alpha}}\left[ (1+C_V A^2 \eta_{n+1}^2)-\left\{ \frac{\eta_{n+1}}{8(1-\alpha)^2\sum_{i=1}^n \eta_i} - \frac{A\sqrt{d}+C_V(A^2+d)}{M^{1-\alpha }d}\frac{\eta_{n+1}}{\sum_{i=1}^n \eta_i} \right\} \right]\\
     &\le \beta (\sum_{i=1}^{n+1} \eta_i )^{\frac{1}{1-\alpha}}\left[ (1+C_V A^2 \eta_{n+1}^2)- \frac{\eta_{n+1}}{16(1-\alpha)^2\sum_{i=1}^n \eta_i}  \right] \\
     &\le \beta (\sum_{i=1}^{n+1} \eta_i )^{\frac{1}{1-\alpha}},     
 \end{align*}
 where the first inequality follows from the fact that $(1+x)^{-\frac{1}{1-\alpha}}+\tfrac{1}{8(1-\alpha)^2}x\le 1$ for all $\alpha\in [0,\tfrac{1}{2}]$ and $x>0$ and the fact that $\eta_{n+1}\le 1 \le (\beta (\sum_{i=1}^n \eta_i)^{\frac{1}{1-\alpha}} )^\alpha$. The second last inequality follows from the definition of $M$ and the last inequality follows from the upper bound of $\eta_{n+1}$. The claimed bounds follows from absorbing $B$ into the constant $M$ by taking any $\eta_n \le 1 $. 
\end{proof} 

\subsection{Rates under constant $h$ and $\nu$}\label{append:optimal scaling}

\namedproof{Proof of Corollary \ref{cor:optimal discussion}}
We first prove part (1).  Under the constant stepsize and regularization parameter $\nu=1-\frac{1}{N}$, results in Theorem \ref{thm:decay of regularized fisher svgd} convert to
\begin{align}\label{eq:simplified condition}
    \mb{E}_{{p^N_{0,K}}} \left[\frac{1}{T} \sum_{n=1}^T  I_{\nu,\text{Stein}}(\rho^N_{n-1}|\pi)  \right]&\le \frac{2\gamma d }{Th} +\frac{2c^*d}{N} + \frac{2 \beta_1 d}{N^2} + \frac{\theta \beta_2 }{B^2 N^{\frac{5}{2}} h } + \frac{\theta^2 }{4Nh},
\end{align}
with the stepsize satisfying
\begin{align}\label{hchoose}
    h = C_0^{-1} \theta N^{-\frac{1}{2}} \min\{ (Th)^{-\frac{\alpha}{1-\alpha}}, 1 \} =\mc{O}( (Th)^{-1})
\end{align}
with $C_0\lesssim d^{\frac{1+\alpha}{2(1-\alpha)}} + d + K^\alpha \sqrt{d}$. Then choosing $T=N^{\frac{2}{1-\alpha}}$ and $\theta=\sqrt{N/T}=N^{-\frac{1+\alpha}{2(1-\alpha)}}$, and $h=C_0^{-(1-\alpha)}N^{-\frac{1+\alpha}{1-\alpha}}$, we can check $h$ satisfies \eqref{hchoose}, and we obtain that
\begin{align*}
     \mb{E}_{{p^N_{0,K}}} \left[\frac{1}{T} \sum_{n=1}^T  I_{\nu,\text{Stein}}(\rho^N_{n-1}|\pi)  \right] \lesssim \frac{d(d^{\frac{1+\alpha}{2}} + d^{1-\alpha} + K^{\alpha(1-\alpha)}d^{\frac{1-\alpha}{2}} )}{N}.
\end{align*}

\noindent Now we prove part (2). Under the constant stepsize and regularization parameter $\nu_n\equiv \nu={\Theta}(N^{-c})$ for some $c\in [0,\tfrac{1}{3})$, results in Theorem \ref{thm:decay of fisher} convert to
\begin{align*}
    \mb{E}_{{p^N_{0,K}}} \left[\frac{1}{T} \sum_{n=1}^T  I(\rho^N_{n-1}|\pi)  \right]&\lesssim \frac{\gamma d }{Th} +\frac{d}{N} + \frac{ d}{N^{1-3c}} + \frac{\theta }{ N^{\frac{3}{2}} h } + \frac{\theta^2 }{N^{1+2c}h},
\end{align*}
and the condition on $h$ becomes
\begin{align}\label{eq:simplified condition 2}
    h =  C_0^{-1} \theta N^{-3c-\frac{1}{2} -\frac{\alpha}{1-\alpha} c} (Th)^{\frac{-\alpha}{1-\alpha}}  \le \frac{1}{Th N^c}.
\end{align}
When $c<\tfrac{1-\alpha}{3-2\alpha}$, for any $\alpha\in [0,\tfrac{1}{2}]$, taking $T=N^{\frac{3}{2} + \max\{ 0, \frac{1}{2}-2c \} }$, $\theta=N^{\frac{3}{2}}/T= N^{-\max\{ 0, \frac{1}{2}-2c \}}$ and $$h  = C_0^{-(1-\alpha)} N^{1-\alpha-c(3-2\alpha)-\frac{3}{2}-\max\{ 0, \frac{1}{2}-2c \}},$$ 
we can check $h$ satisfies \eqref{eq:simplified condition 2}, and we get
\begin{align*}
    \mb{E}_{{p^N_{0,K}}} \left[\frac{1}{T} \sum_{n=1}^T I(\rho^N_n|\pi)  \right]\lesssim \frac{d(d^{\frac{1+\alpha}{2}} + d^{1-\alpha} + K^{\alpha(1-\alpha)}d^{\frac{1-\alpha}{2}}) }{N^{1-\alpha-c(3-2\alpha)}}.
\end{align*}
Similarly, according to Theorem \ref{thm:decay of fisher}, we get same optimal order estimation of $W_1(\mu^N_{T,av},\pi)^2$.
\end{proof}

\section{Relation between regularized Fisher information and KSD}\label{append:KSD and regularized Fisher}

As explained in \cite{he2024regularized},
\begin{align*}
    \mathrm{KSD}(\rho|\pi) & = \left\langle \iota_{k,\rho}^* \nabla \log \frac{\rho}{\pi},  \iota_{k,\rho}^* \nabla \log \frac{\rho}{\pi}\right\rangle_{\mc{H}_k^d},  \\
    I_{\nu,\text{Stein}}(\rho|\pi) &= \left\langle \iota_{k,\rho}^* \nabla \log \frac{\rho}{\pi}, \left((1-\nu)\iota_{k,\rho}^*\iota_{k,\rho}+\nu I_d \right)^{-1} \iota_{k,\rho}^* \nabla \log \frac{\rho}{\pi}\right\rangle_{\mc{H}_k^d}. 
\end{align*}
Then we have
\begin{align*}
    \mathrm{KSD}(\rho|\pi) &\le \|(1-\nu)\iota_{k,\rho}^*\iota_{k,\rho}+\nu I_d \|_{\mc{H}_k^d\to \mc{H}_k^d} I_{\nu,\text{Stein}}(\rho|\pi) \\
    &\le (B+1) I_{\nu,\text{Stein}}(\rho|\pi).
\end{align*}
Therefore, $I_{\nu,\text{Stein}}(\bar{\rho}^N|\pi)\to 0$ implies that $\mathrm{KSD}(\bar{\rho}^N|\pi)\to 0$.
\section{Convexity of regularized Fisher information}
In this section, we show that the functional $\rho\mapsto I_{\nu,\text{Stein}}(\rho|\pi)$ is convex.
\begin{lemma}[Convexity of $\rho\mapsto I_{\nu,\text{Stein}}(\rho|\pi)$]\label{lem:convexity of regularized fisher} Assume Assumption \ref{assump:kernel} and Assumption \ref{assump:target} hold. The functional $I_{\nu,\text{Stein}}(\cdot|\pi):\mc{P}(\mb{R}^d)\to \mb{R}_{+}$ is convex, where $I_{\nu,\text{Stein}}(\cdot|\pi)$ is defined by \eqref{eq:regularized fisher} for any $\nu\in (0,1]$.     
\end{lemma}

\namedproof{Proof of Lemma \ref{lem:convexity of regularized fisher}.}
    Let $\rho_1,\rho_2\in \mc{P}(\mb{R}^d)$ and $\rho=\theta\rho_1+(1-\theta)\rho_2\in \mc{P}(\mb{R}^d)$ for $\theta \in (0,1)$. Let $\phi_i= \big((1-\nu)\iota_{k,\rho_i}^*\iota_{k,\rho_i}+\nu I_d\big)^{-1}\iota_{k,\rho_i}^*\nabla\log\frac{\rho_i}{\pi}$ for $i=1,2$ and $\phi= \big((1-\nu)\iota_{k,\rho}^*\iota_{k,\rho}+\nu I_d\big)^{-1}\iota_{k,\rho}^*\nabla\log\frac{\rho}{\pi}$. Our goal is to prove 
    \begin{align}\label{eq:convexity}
        I_{\nu,\text{Stein}}(\rho|\pi)\le \theta I_{\nu,\text{Stein}}(\rho_1|\pi)+(1-\theta)I_{\nu,\text{Stein}}(\rho_2|\pi).
    \end{align}
    We have
    \begin{align*}
        I_{\nu,\text{Stein}}(\rho|\pi)=&\left\langle -\int \nabla_2 k(\cdot,y)\rho(y)\dee y+\int k(\cdot,y)\nabla V(y)\rho(y)\dee y, \phi \right\rangle_{\mc{H}_k^d} \\
        =&\theta\left\langle -\int \nabla_2 k(\cdot,y)\rho_1(y)\dee y+\int k(\cdot,y)\nabla V(y)\rho_1(y)\dee y, \phi \right\rangle_{\mc{H}_k^d}\\
        &+(1-\theta)\left\langle -\int \nabla_2 k(\cdot,y)\rho_2(y)\dee y+\int k(\cdot,y)\nabla V(y)\rho_2(y)\dee y, \phi \right\rangle_{\mc{H}_k^d}\\
        =&\theta I_{\nu,\text{Stein}}(\rho_1|\pi)+(1-\theta)I_{\nu,\text{Stein}}(\rho_2|\pi)\\
        &+\theta\left\langle -\int \nabla_2 k(\cdot,y)\rho_1(y)\dee y+\int k(\cdot,y)\nabla V(y)\rho_1(y)\dee y, \phi-\phi_1 \right\rangle_{\mc{H}_k^d}\\
        &+(1-\theta)\left\langle -\int \nabla_2 k(\cdot,y)\rho_2(y)\dee y+\int k(\cdot,y)\nabla V(y)\rho_2(y)\dee y, \phi-\phi_2 \right\rangle_{\mc{H}_k^d}.
    \end{align*}
    It suffices to show 
    \begin{align*}
        &\theta\underbrace{\left\langle -\int \nabla_2 k(\cdot,y)\rho_1(y)\dee y+\int k(\cdot,y)\nabla V(y)\rho_1(y)\dee y, \phi-\phi_1 \right\rangle_{\mc{H}_k^d}}_{I_1}
        \\
        +&(1-\theta)\underbrace{\left\langle -\int \nabla_2 k(\cdot,y)\rho_2(y)\dee y+\int k(\cdot,y)\nabla V(y)\rho_2(y)\dee y, \phi-\phi_2 \right\rangle_{\mc{H}_k^d}}_{I_2}\le 0
    \end{align*}
    For simplicity, we denote $\mc{T}_{k,\rho}\coloneqq \iota_{k,\rho}^*\iota_{k,\rho}$ for all $\rho$. Then we can write $I_1$ as
    \begin{align*}
         &\left\langle \iota_{k,\rho_1}^*\nabla\log\frac{\rho_1}{\pi}, \big((1-\nu)\mc{T}_{k,\rho}+\nu I_d\big)^{-1}\iota_{k,\rho}^*\nabla\log\frac{\rho}{\pi}-\big((1-\nu)\mc{T}_{k,\rho_1}+\nu I_d\big)^{-1}\iota_{k,\rho_1}^*\nabla\log\frac{\rho_1}{\pi} \right\rangle_{\mc{H}_k^d} \\&=\theta\left\langle \iota_{k,\rho_1}^*\nabla\log\frac{\rho_1}{\pi}, \big((1-\nu)\mc{T}_{k,\rho}+\nu I_d\big)^{-1}\iota_{k,\rho_1}^*\nabla\log\frac{\rho_1}{\pi}-\big((1-\nu)\mc{T}_{k,\rho_1}+\nu I_d\big)^{-1}\iota_{k,\rho_1}^*\nabla\log\frac{\rho_1}{\pi} \right\rangle_{\mc{H}_k^d}\\
        &+(1-\theta)\left\langle \iota_{k,\rho_1}^*\nabla\log\frac{\rho_1}{\pi}, \big((1-\nu)\mc{T}_{k,\rho}+\nu I_d\big)^{-1}\iota_{k,\rho_2}^*\nabla\log\frac{\rho_2}{\pi}-\big((1-\nu)\mc{T}_{k,\rho_1}+\nu I_d\big)^{-1}\iota_{k,\rho_1}^*\nabla\log\frac{\rho_1}{\pi} \right\rangle_{\mc{H}_k^d}\\
        &=\left\langle \iota_{k,\rho_1}^*\nabla\log\frac{\rho_1}{\pi}, \bigg(\theta\big((1-\nu)\mc{T}_{k,\rho}+\nu I_d\big)^{-1}-\big((1-\nu)\mc{T}_{k,\rho_1}+\nu I_d\big)^{-1}\bigg)\iota_{k,\rho_1}^*\nabla\log\frac{\rho_1}{\pi} \right\rangle_{\mc{H}_k^d}\\
        &+(1-\theta)\left\langle \iota_{k,\rho_1}^*\nabla\log\frac{\rho_1}{\pi}, \big((1-\nu)\mc{T}_{k,\rho}+\nu I_d\big)^{-1}\iota_{k,\rho_2}^*\nabla\log\frac{\rho_2}{\pi} \right\rangle_{\mc{H}_k^d}
    \end{align*}
    where the first identity follows from the fact that $\rho\mapsto\iota_{k,\rho}^*\nabla\log\frac{\rho}{\pi}$ is linear.
    Since $\rho\mapsto \mc{T}_{k,\rho}=\iota_{k,\rho}^*\iota_{k,\rho}$ is linear, we have 
    \begin{align*}
     &\quad \theta\big((1-\nu)\mc{T}_{k,\rho}+\nu I_d\big)^{-1}-\big((1-\nu)\mc{T}_{k,\rho_1}+\nu I_d\big)^{-1}\\
     &=   \big(\theta^{-1}(1-\nu)\mc{T}_{k,\rho}+\theta^{-1}\nu I_d\big)^{-1}-\big((1-\nu)\mc{T}_{k,\rho_1}+\nu I_d\big)^{-1}\\
     &=\big(\theta^{-1}(1-\nu)\mc{T}_{k,\rho}+\theta^{-1}\nu I_d\big)^{-1}\big((1-\nu)\mc{T}_{k,\rho_1}+\nu I_d- \theta^{-1}(1-\nu)\mc{T}_{k,\rho}-\theta^{-1}\nu I_d\big)\big((1-\nu)\mc{T}_{k,\rho_1}+\nu I_d\big)^{-1}\\
     &=\big(\theta^{-1}(1-\nu)\mc{T}_{k,\rho}+\theta^{-1}\nu I_d\big)^{-1}\big((1-\theta^{-1})\nu I_d-\theta^{-1}(1-\theta)(1-\nu)\mc{T}_{k,\rho_2} \big)\big((1-\nu)\mc{T}_{k,\rho_1}+\nu I_d\big)^{-1}\\
     &=-(1-\theta)\big((1-\nu)\mc{T}_{k,\rho}+\nu I_d\big)^{-1}\big((1-\nu)\mc{T}_{k,\rho_2} +\nu I_d\big)\big((1-\nu)\mc{T}_{k,\rho_1}+\nu I_d\big)^{-1}.
    \end{align*}
    Therefore,
    \small{
    \begin{align*}
         I_1&=(1-\theta)\left\langle \iota_{k,\rho_1}^*\nabla\log\frac{\rho_1}{\pi}, \big((1-\nu)\mc{T}_{k,\rho}+\nu I_d\big)^{-1}\iota_{k,\rho_2}^*\nabla\log\frac{\rho_2}{\pi} \right\rangle_{\mc{H}_k^d}\\
         &-(1-\theta)\left\langle \iota_{k,\rho_1}^*\nabla\log\frac{\rho_1}{\pi}, \big((1-\nu)\mc{T}_{k,\rho}+\nu I_d\big)^{-1}\big((1-\nu)\mc{T}_{k,\rho_2} +\nu I_d\big)\big((1-\nu)\mc{T}_{k,\rho_1}+\nu I_d\big)^{-1}\iota_{k,\rho_1}^*\nabla\log\frac{\rho_1}{\pi} \right\rangle_{\mc{H}_k^d}.
    \end{align*}}
    Similarly, for $I_2$, we have
    \begin{align*}
        I_2&=\left\langle \iota_{k,\rho_2}^*\nabla\log\frac{\rho_2}{\pi}, \big((1-\nu)\mc{T}_{k,\rho}+\nu I_d\big)^{-1}\iota_{k,\rho}^*\nabla\log\frac{\rho}{\pi}-\big((1-\nu)\mc{T}_{k,\rho_2}+\nu I_d\big)^{-1}\iota_{k,\rho_2}^*\nabla\log\frac{\rho_2}{\pi} \right\rangle_{\mc{H}_k^d} \\
        &=\theta\left\langle \iota_{k,\rho_2}^*\nabla\log\frac{\rho_2}{\pi}, \big((1-\nu)\mc{T}_{k,\rho}+\nu I_d\big)^{-1}\iota_{k,\rho_1}^*\nabla\log\frac{\rho_1}{\pi}-\big((1-\nu)\mc{T}_{k,\rho_2}+\nu I_d\big)^{-1}\iota_{k,\rho_2}^*\nabla\log\frac{\rho_2}{\pi} \right\rangle_{\mc{H}_k^d}\\
        &+(1-\theta)\left\langle \iota_{k,\rho_2}^*\nabla\log\frac{\rho_2}{\pi}, \big((1-\nu)\mc{T}_{k,\rho}+\nu I_d\big)^{-1}\iota_{k,\rho_2}^*\nabla\log\frac{\rho_2}{\pi}-\big((1-\nu)\mc{T}_{k,\rho_2}+\nu I_d\big)^{-1}\iota_{k,\rho_2}^*\nabla\log\frac{\rho_2}{\pi} \right\rangle_{\mc{H}_k^d}\\
        &=\left\langle \iota_{k,\rho_2}^*\nabla\log\frac{\rho_2}{\pi}, \bigg((1-\theta)\big((1-\nu)\mc{T}_{k,\rho}+\nu I_d\big)^{-1}-\big((1-\nu)\mc{T}_{k,\rho_2}+\nu I_d\big)^{-1}\bigg)\iota_{k,\rho_2}^*\nabla\log\frac{\rho_2}{\pi}\right\rangle_{\mc{H}_k^d}\\
        &+\theta\left\langle \iota_{k,\rho_2}^*\nabla\log\frac{\rho_2}{\pi}, \big((1-\nu)\mc{T}_{k,\rho}+\nu I_d\big)^{-1}\iota_{k,\rho_1}^*\nabla\log\frac{\rho_1}{\pi} \right\rangle_{\mc{H}_k^d}\\
        &=\theta\left\langle \iota_{k,\rho_2}^*\nabla\log\frac{\rho_2}{\pi}, \big((1-\nu)\mc{T}_{k,\rho}+\nu I_d\big)^{-1}\iota_{k,\rho_1}^*\nabla\log\frac{\rho_1}{\pi} \right\rangle_{\mc{H}_k^d}\\
         &-\theta\left\langle \iota_{k,\rho_2}^*\nabla\log\frac{\rho_2}{\pi}, \big((1-\nu)\mc{T}_{k,\rho}+\nu I_d\big)^{-1}\big((1-\nu)\mc{T}_{k,\rho_1} +\nu I_d\big)\big((1-\nu)\mc{T}_{k,\rho_2}+\nu I_d\big)^{-1}\iota_{k,\rho_2}^*\nabla\log\frac{\rho_2}{\pi} \right\rangle_{\mc{H}_k^d}
    \end{align*}
    If we denote $T_{k,\mu}=\mc{T}_{k,\mu} +\nu I_d$ for all $\mu\in \mc{P}(\mb{R}^d)$, then we have
    \begin{align*}
        \theta I_1+(1-\theta)I_2 =& 2\theta(1-\theta)\left\langle \iota_{k,\rho_1}^*\nabla\log\frac{\rho_1}{\pi}, T_{k,\rho}^{-1}\iota_{k,\rho_2}^*\nabla\log\frac{\rho_2}{\pi} \right\rangle_{\mc{H}_k^d}\\
        &-\theta(1-\theta)\left\langle \iota_{k,\rho_1}^*\nabla\log\frac{\rho_1}{\pi}, T_{k,\rho}^{-1}T_{k,\rho_2}T_{k,\rho_1}^{-1}\iota_{k,\rho_1}^*\nabla\log\frac{\rho_1}{\pi} \right\rangle_{\mc{H}_k^d}\\
         &-\theta(1-\theta)\left\langle \iota_{k,\rho_2}^*\nabla\log\frac{\rho_2}{\pi}, T_{k,\rho}^{-1}T_{k,\rho_1}T_{k,\rho_2}^{-1}\iota_{k,\rho_2}^*\nabla\log\frac{\rho_2}{\pi} \right\rangle_{\mc{H}_k^d}.
    \end{align*}
    Even though $T_{k,\rho}$ and $T_{k,\mu}^{-1}$ doesn't commute for general $\rho,\mu\in \mc{P}(\mb{R}^d)$, we can apply Young's inequality to prove the above quantity is non-positive by writing the three terms as
    \begin{align*}
        \left\langle \iota_{k,\rho_1}^*\nabla\log\frac{\rho_1}{\pi}, T_{k,\rho}^{-1}\iota_{k,\rho}^*\nabla\log\frac{\rho_2}{\pi} \right\rangle_{\mc{H}_k^d}&=\left\langle T_{k,\rho_1}^{-1}\iota_{k,\rho_1}^*\nabla\log\frac{\rho_1}{\pi}, \big(T_{k,\rho_1}T_{k,\rho}^{-1}T_{k,\rho_2}\big)T_{k,\rho_2}^{-1}\iota_{k,\rho}^*\nabla\log\frac{\rho_2}{\pi} \right\rangle_{\mc{H}_k^d} \\
        \left\langle \iota_{k,\rho_1}^*\nabla\log\frac{\rho_1}{\pi}, T_{k,\rho}^{-1}T_{k,\rho_2}T_{k,\rho_1}^{-1}\iota_{k,\rho_1}^*\nabla\log\frac{\rho_1}{\pi} \right\rangle_{\mc{H}_k^d}&=\left\langle T_{k,\rho_1}^{-1}\iota_{k,\rho_1}^*\nabla\log\frac{\rho_1}{\pi}, \big(T_{k,\rho_1}T_{k,\rho}^{-1}T_{k,\rho_2}\big)T_{k,\rho_1}^{-1}\iota_{k,\rho_1}^*\nabla\log\frac{\rho_1}{\pi} \right\rangle_{\mc{H}_k^d} \\
        \left\langle \iota_{k,\rho_2}^*\nabla\log\frac{\rho_2}{\pi}, T_{k,\rho}^{-1}T_{k,\rho_1}T_{k,\rho_2}^{-1}\iota_{k,\rho_2}^*\nabla\log\frac{\rho_2}{\pi} \right\rangle_{\mc{H}_k^d}&=\left\langle T_{k,\rho_2}^{-1}\iota_{k,\rho_1}^*\nabla\log\frac{\rho_1}{\pi}, \big(T_{k,\rho_1}T_{k,\rho}^{-1}T_{k,\rho_2}\big)T_{k,\rho_2}^{-1}\iota_{k,\rho_2}^*\nabla\log\frac{\rho_2}{\pi} \right\rangle_{\mc{H}_k^d}
    \end{align*}
    where the three identities follows from the self-adjointness of $T_{k,\mu}$ and $T_{k,\mu}^{-1}$ for all $\mu\in \mc{P}(\mb{R}^d)$. Since $T_{k,\rho_1}T_{k,\rho}^{-1}T_{k,\rho_2}:\mc{H}_k^d\to \mc{H}_k^d$ is positive definite, it is also self-adjoint, i.e., $$T_{k,\rho_1}T_{k,\rho}^{-1}T_{k,\rho_2}=(T_{k,\rho_1}T_{k,\rho}^{-1}T_{k,\rho_2})^*=T_{k,\rho_2}T_{k,\rho}^{-1}T_{k,\rho_1}.$$ Therefore, the non-negativity follows from applying Young's inequality $\| x\|^2+\| y\|^2\ge 2\langle x,y\rangle$ for 
    \begin{align*}
    x=(T_{k,\rho_1}T_{k,\rho}^{-1}T_{k,\rho_2})^{1/2}T_{k,\rho_1}^{-1}\iota_{k,\rho_1}^*\nabla\log\frac{\rho_1}{\pi}\quad\text{and}\quad y=(T_{k,\rho_1}T_{k,\rho}^{-1}T_{k,\rho_2})^{1/2}T_{k,\rho_1}^{-1}\iota_{k,\rho_2}^*\nabla\log\frac{\rho_2}{\pi}.
    \end{align*}
\end{proof}

\section{Preliminaries on Reproducing Kernel Hilbert Spaces (RKHS)}\label{app:RKHS}
This appendix collects RKHS facts used in the analysis. We refer to~\cite{steinwart2008support,berlinet2011reproducing,paulsen2016introduction} for detailed expositions.

Let $\mathcal{H}_k$ be a separable reproducing kernel Hilbert space (RKHS) of real-valued functions on $\mathbb{R}^d$ with reproducing kernel $k:\mathbb{R}^d\times\mathbb{R}^d\to\mathbb{R}$ and norm $\|\cdot\|_{\mathcal{H}_k}$. Recall the reproducing property:
\[
f(x)=\langle f, k(x,\cdot)\rangle_{\mathcal{H}_k}\qquad \forall f\in\mathcal{H}_k,\ x\in\mathbb{R}^d.
\]
The next proposition summarizes standard embedding and operator-theoretic properties that we repeatedly use, under Assumption~\ref{assump:kernel}; see~\cite[Lem.~4.23, Thms.~4.26--4.27]{steinwart2008support}.

\begin{proposition}[Basic RKHS properties~\cite{steinwart2008support}]\label{prop:RKHS_property_app}
\begin{itemize}
    \item[(i)] (\emph{Boundedness and $L_\infty$-embedding}) The kernel $k$ is bounded (i.e., $\sup_{x}k(x,x)<\infty$) if and only if every $f\in\mathcal{H}_k$ is bounded. In this case, the inclusion (embedding) $\iota_\infty:\mathcal{H}_k\to L_\infty(\mathbb{R}^d)$ is continuous and
    \[
    \|\iota_\infty\|_{\mathcal{H}_k\to L_\infty}=\|k\|_\infty,\qquad 
    \|k\|_\infty:=\sup_{x\in\mathbb{R}^d}\sqrt{k(x,x)}.
    \]
    \emph{Intuition:} $k(x,x)$ controls the size of the point-evaluation functional $f\mapsto f(x)$ via the reproducing property.

    \item[(ii)] (\emph{$L_2(\mu)$-embedding and adjoint}) Let $\mu$ be a $\sigma$-finite measure on $\mathbb{R}^d$ and assume $k$ is measurable and satisfies
    \[
    \|k\|_{L_2(\mu)}:=\Big(\int_{\mathbb{R}^d} k(x,x)\,d\mu(x)\Big)^{1/2}<\infty.
    \]
    Then $\mathcal{H}_k\subset L_2(\mu)$ and the inclusion $\iota_{k,\mu}:\mathcal{H}_k\to L_2(\mu)$ is continuous with
    \[
    \|\iota_{k,\mu}\|_{\mathcal{H}_k\to L_2(\mu)}\le \|k\|_{L_2(\mu)}.
    \]
    Moreover, the adjoint $\iota_{k,\mu}^*:L_2(\mu)\to\mathcal{H}_k$ is the (kernel) integral operator
    \[
    (\iota_{k,\mu}^* g)(x)=\int_{\mathbb{R}^d} k(x,y)\,g(y)\,d\mu(y),\qquad g\in L_2(\mu).
    \]
    \emph{Intuition:} $\iota_{k,\mu}^*$ maps an $L_2(\mu)$ function to a smoother RKHS element by averaging it against $k(\cdot,y)$.

    \item[(iii)] (\emph{Density vs.\ injectivity}) Under (ii), $\mathcal{H}_k$ is dense in $L_2(\mu)$ if and only if $\iota_{k,\mu}^*$ is injective. Equivalently, $\iota_{k,\mu}^*$ has dense range in $\mathcal{H}_k$ if and only if $\iota_{k,\mu}$ is injective.

    \item[(iv)] (\emph{Hilbert--Schmidt embedding and compactness}) Under (ii), the inclusion $\iota_{k,\mu}:\mathcal{H}_k\to L_2(\mu)$ is Hilbert--Schmidt with
    \[
    \|\iota_{k,\mu}\|_{\mathrm{HS}}=\|k\|_{L_2(\mu)}.
    \]
    Consequently, the associated integral operator
    \[
    \mathcal{T}_{k,\mu}:=\iota_{k,\mu}\iota_{k,\mu}^*:L_2(\mu)\to L_2(\mu)
    \]
    is compact, positive, self-adjoint, and trace-class (nuclear).
\end{itemize}
\end{proposition}

\paragraph{Vector-valued inner products.}
For $f\in\mathcal{H}_k^d$ with $f=[f_1,\dots,f_d]^\top$ and $g\in\mathcal{H}_k$, we use the componentwise pairing
\[
\langle f,g\rangle_{\mathcal{H}_k}\in\mathbb{R}^d,\qquad 
(\langle f,g\rangle_{\mathcal{H}_k})_i:=\langle f_i,g\rangle_{\mathcal{H}_k}.
\]
Similarly, for $f\in L_2^d(\mu)$ and $g\in L_2(\mu)$,
\[
\langle f,g\rangle_{L_2(\mu)}\in\mathbb{R}^d,\qquad 
(\langle f,g\rangle_{L_2(\mu)})_i:=\langle f_i,g\rangle_{L_2(\mu)}.
\]

\paragraph{Range of $\mathcal{T}_{k,\mu}^{1/2}$.}
Since $\mathcal{T}_{k,\mu}=\iota_{k,\mu}\iota_{k,\mu}^*$, one can identify
\[
\mathrm{Ran}(\mathcal{T}_{k,\mu}^{1/2})=\mathcal{H}_k \subset L_2(\mu),
\]
see, e.g.,~\cite{cucker2007learning}. Intuitively, $\mathcal{T}_{k,\mu}^{1/2}$ links the $L_2(\mu)$ geometry to the RKHS geometry.

\paragraph{Spectral representation.}
Let $(\lambda_i,e_i)_{i\ge 1}$ be eigenpairs of $\mathcal{T}_{k,\mu}$ with $\lambda_1\ge \lambda_2\ge \cdots >0$ and $\{e_i\}_{i\ge 1}$ orthonormal in $\mathrm{Ran}(\mathcal{T}_{k,\mu})$. Then, for all $f\in L_2(\mu)$,
\begin{equation}\label{eq:SD_of_Tk_app}
\mathcal{T}_{k,\mu}f=\sum_{i=1}^\infty \lambda_i\,\langle f,e_i\rangle_{L_2(\mu)}\,e_i.
\end{equation}
Computing this decomposition for general $(k,\mu)$ is typically difficult and is only known in special cases (see, e.g.,~\cite{minh2006mercer,scetbon2021spectral}). In this work, the spectral form~\eqref{eq:SD_of_Tk_app} is used purely as an analysis tool; the practical algorithm does not require explicit knowledge of $(\lambda_i,e_i)$.

\begin{remark}\label{rem:nullspace_app}
The notation above is simplest when $\mathcal{T}_{k,\mu}$ has a trivial null space, in which case $\overline{\mathrm{Ran}(\mathcal{T}_{k,\mu})}=L_2(\mu)$ and the eigenfunctions form an orthonormal basis of $L_2(\mu)$. Our arguments do \emph{not} require this: if $\ker(\mathcal{T}_{k,\mu})\neq\{0\}$, then $\overline{\mathrm{Ran}(\mathcal{T}_{k,\mu})}\subset L_2(\mu)$, and one may extend $\{e_i\}$ (by adding a basis of the null space) without affecting the conclusions.
\end{remark}

\section{A roadmap to checking \eqref{eq:regularizer condition dynamics} and \eqref{eq:fisher equivalence} for finite-dimensional RKHS.}\label{append:sufficient condition}  

For the sake of explicit and tangible conditions as requested by reviewers, we consider a finite-dimensional RKHS
\[
\mathcal{H}_k=\Bigl\{f_a:f_a(x)=\sum_{i=1}^M a_i\Phi_i(x),\ a\in\mathbb{R}^M\Bigr\},
\]
with feature map $\Phi(x)=(\Phi_1(x),\dots,\Phi_M(x))^\top\in\mathbb{R}^M$ and kernel
\[
k(x,y)=\sum_{i=1}^M \Phi_i(x)\Phi_i(y)=\Phi(x)^\top\Phi(y).
\]
Then
\[
\mathcal{H}_k^d=\{f_A:f_A(x)=A^\top \Phi(x),\ A\in\mathbb{R}^{M\times d}\}.
\]

Let $i_{k,\mu}:\mathcal{H}_k^d\to L_2^d(\mu)$ denote the inclusion map and
$i_{k,\mu}^*:L_2^d(\mu)\to \mathcal{H}_k^d$ its adjoint. For any $\mu$ such that
$I(\mu\mid\pi)$ and $\mc{J}(\mu,\pi)$ are well-defined, Appendix G yields
\[
I(\mu\mid\pi)
=
\Bigl\langle
(i_{k,\mu}^*i_{k,\mu})^{-1}i_{k,\mu}^*\nabla \log \frac{\mu}{\pi},
\,i_{k,\mu}^*\nabla \log \frac{\mu}{\pi}
\Bigr\rangle_{\mathcal{H}_k^d},
\]
and
\begin{align*}
\|\mc{J}(\mu,\pi)\|_{L_2^d(\mu)}^2
&=
\langle (i_{k,\mu}i_{k,\mu}^*)^{-\gamma(t)}\nabla \log \frac{\mu}{\pi}, (i_{k,\mu}i_{k,\mu}^*)^{-\gamma(t)}\nabla\log \frac{\mu}{\pi} \rangle_{L_2^d(\mu)}\\
&=
\Bigl\langle
(i_{k,\mu}^*i_{k,\mu})^{-2\gamma(t)-1}i_{k,\mu}^*\nabla \log \frac{\mu}{\pi},
\,i_{k,\mu}^*\nabla \log \frac{\mu}{\pi}
\Bigr\rangle_{\mathcal{H}_k^d},
\end{align*}
where inverses are understood as Moore--Penrose pseudoinverses on the relevant ranges.

Under the finite-dimensional feature representation, the operator
$i_{k,\mu}^*i_{k,\mu}:\mathcal{H}_k^d\to \mathcal{H}_k^d$ is represented by the matrix
\[
C(\mu):=\int \Phi(x)\Phi(x)^\top\,\mu(dx)\in\mathbb{R}^{M\times M}.
\]
Indeed, for $f_A(x)=A^\top\Phi(x)$,
\[
\langle f_A,i_{k,\mu}^*i_{k,\mu}f_A\rangle_{\mathcal{H}_k^d}
=
\|i_{k,\mu}f_A\|_{L_2^d(\mu)}^2
=
\int \|A^\top\Phi(x)\|^2\,\mu(dx)
=
\langle A,\,C(\mu)A\rangle_F.
\]
Assume that $\nabla \log \frac{d\mu}{d\pi}\in L_2^d(\mu)$. Under our finite-dimensional feature representation, the map
\[
A\mapsto f_A
\]
is injective, there exists a unique matrix $A(\mu)\in\mathbb{R}^{M\times d}$ such that
\[
i_{k,\mu}^*\nabla \log \frac{d\mu}{d\pi}=f_{A(\mu)}.
\]
Equivalently,
\[
i_{k,\mu}^*\nabla \log \frac{d\mu}{d\pi}(x)=A(\mu)^\top \Phi(x),
\qquad x\in\mathsf{X}.
\]
With this notation,
\[
I(\mu\mid \pi)=\langle C(\mu)^{-1}A(\mu),A(\mu)\rangle_F,
\qquad
\|\mc{J}(\mu,\pi)\|_{L_2^d(\mu)}^2
=
\langle C(\mu)^{-2\gamma(t)-1}A(\mu),A(\mu)\rangle_F.
\]
With the above representation, the RHS of (16) satisfies
$$\inf_{t\in [0,T]} \left( \frac{\langle C(\mu^N(t))^{-1} A(\mu^N(t)), A(\mu^N(t)) \rangle_{\mathrm{F}}}{2\langle C(\mu^N(t))^{-2\gamma(t)-1} A(\mu^N(t)), A(\mu^N(t)) \rangle_{\mathrm{F}}} \right)^{\frac{1}{2\gamma(t)}} \ge \frac{1}{2^{\frac{1}{2\gamma(t)}}}\inf_{t\in [0,T]}\lambda_{\min}(C(\mu^N(t))).$$ 

Therefore, assuming $\inf_{t \in [0,T]} \gamma(t)>0$, a sufficient condition for $(16)$ and $(17)$ is
\begin{equation}
\inf_{t\in[0,T]}\lambda_{\min}(C(\mu^N(t)))>0.
\label{1}
\end{equation}
According to the definition of $\mu^N(t)$, this condition can be written as
\[
\inf_{t\in[0,T]}
\lambda_{\min}
\left(
\mathbb{E}_{x(0)\sim p^N(0)}
\Bigl[
\frac1N\sum_{i=1}^N \Phi(x_i(t))\Phi(x_i(t))^\top
\Bigr]
\right)
>0.
\]
Condition $(1)$ is natural for the particle system: it requires that the
sample-covariance-type matrices generated by the feature map $\Phi$ remain sufficiently non-singular
within the time horizon.

We now illustrate this in a special case.
\subsection{Illustrating technique for Gaussian target and bilinear kernel: }
We further illustrate this condition with  $\Phi(x)=(1,x^\top)^\top$ and
\[
k(x,y)=1+x^\top y.
\]
This kernel was also used in~\cite{he2024regularized} to illustrate the conditions in mean-field setting.

The matrix $C(\mu^N(t))$ in this case takes the explicit block form
\begin{equation}
C(\mu^N(t))
=
\begin{pmatrix}
1 & \mathbb{E}_{\mu^N(t)}[x]^\top\\
\mathbb{E}_{\mu^N(t)}[x] & \mathbb{E}_{\mu^N(t)}[xx^\top]
\end{pmatrix}.
\label{2}
\end{equation}
Hence \eqref{1} holds, for example, if
\[
\inf_{t\in[0,T]}\lambda_{\min}\bigl(\operatorname{Cov}(\mu^N(t))\bigr)>0
\qquad\text{and}\qquad
\sup_{t\in[0,T]}\|\mathbb{E}_{\mu^N(t)}[x]\|<\infty.
\]

Another perspective on \eqref{1} is through the mean-field limit $\rho(t)$ of $\mu^N(t)$.
Through mean-field analysis and long-time propagation of chaos techniques, it is plausible that one can relate
 $\lambda_{\min}\bigl(\operatorname{Cov}(\mu^N(t))\bigr)$ to $\lambda_{\min}\bigl(\operatorname{Cov}(\rho(t))\bigr)$. 
 
 In the linear-kernel case
$k(x,y)=1+x^\top y$, we have
\[
\|C(\rho(t))-C(\mu^N(t))\|_{\mathrm{op}}
=
\left\|
\begin{pmatrix}
0 & \mathbb{E}_{\rho(t)}[x]^\top-\mathbb{E}_{\mu^N(t)}[y]^\top\\
\mathbb{E}_{\rho(t)}[x]-\mathbb{E}_{\mu^N(t)}[y]
&
\mathbb{E}_{\rho(t)}[xx^\top]-\mathbb{E}_{\mu^N(t)}[yy^\top]
\end{pmatrix}
\right\|_{\mathrm{op}}.
\]
Therefore,
\[
\|C(\rho(t))-C(\mu^N(t))\|_{\mathrm{op}}
\le
\|\mathbb{E}_{\rho(t)}[x]-\mathbb{E}_{\mu^N(t)}[y]\|
+
\|\mathbb{E}_{\rho(t)}[xx^\top]-\mathbb{E}_{\mu^N(t)}[yy^\top]\|_{\mathrm{op}}.
\]
Using a coupling $(x,y)$ of $\rho(t)$ and $\mu^N(t)$, we obtain
\[
\|\mathbb{E}_{\rho(t)}[x]-\mathbb{E}_{\mu^N(t)}[y]\|
\le
\mathbb{E}\|x-y\|,
\]
and
\[
\|\mathbb{E}_{\rho(t)}[xx^\top]-\mathbb{E}_{\mu^N(t)}[yy^\top]\|_{\mathrm{op}}
\le
\mathbb{E}\bigl[\|xx^\top-yy^\top\|_{\mathrm{op}}\bigr]
\le
\mathbb{E}\bigl[(\|x\|+\|y\|)\|x-y\|\bigr].
\]
Hence
\[
\|C(\rho(t))-C(\mu^N(t))\|_{\mathrm{op}}
\le
\mathbb{E}\|x-y\|
+
\mathbb{E}\bigl[(\|x\|+\|y\|)\|x-y\|\bigr].
\]
By Cauchy--Schwarz,
\[
\mathbb{E}\bigl[(\|x\|+\|y\|)\|x-y\|\bigr]
\le
\Bigl( \mathbb{E}(\|x\|+\|y\|)^2 \Bigr)^{1/2}
\Bigl( \mathbb{E}\|x-y\|^2 \Bigr)^{1/2},
\]
so that
\[
\|C(\rho(t))-C(\mu^N(t))\|_{\mathrm{op}}
\le
\Bigl(1+\mathbb{E}_{\rho(t)}[\|x\|^2]^{1/2}
+\mathbb{E}_{\mu^N(t)}[\|y\|^2]^{1/2}\Bigr)\,
W_2(\rho(t),\mu^N(t)).
\]
If
\[
\sup_{t\in[0,T]}\mathbb{E}_{\rho(t)}[\|x\|^2]<\infty,
\qquad
\sup_{t\in[0,T]}\mathbb{E}_{\mu^N(t)}[\|y\|^2]<\infty,
\]
and using a propagation-of-chaos estimate of the form
\[
W_2(\rho(t),\mu^N(t))\le f(t)\,N^{-1/2},
\]
proved in \cite[Theorem 8]{he2024regularized},
then one has
\[
\|C(\rho(t))-C(\mu^N(t))\|_{\mathrm{op}}
\lesssim
f(t)\,N^{-1/2}.
\]
Moreover, condition \eqref{1} holds with $C(\rho(t))$ in place of $C(\mu^N(t))$ uniformly on $[0,T]$ as discussed in \cite[Remark 10]{he2024regularized}). This implies that condition \eqref{1} also
holds for $C(\mu^N(t))$, and hence for $C(\mu^N(t))$ uniformly on $[0,T]$, provided $N$ is sufficiently large. 

This can also be extended to longer time intervals $T = T(N)$, eg. when $T(N) =N^{2/3}$ used in some of our rate computations, provided one has \emph{long-time propagation of chaos estimates} of the form
\[
\sup_{t \in [0,T(N)]}W_2(\rho(t),\mu^N(t))\le C\,N^{-1/2},
\]
where $C$ does not depend on $N$. Results like this are of independent interest, and it is the topic of ongoing research. 

For more structured scenarios, like with the Gaussian target and bilinear kernel, we expect to show something even stronger, namely, uniform-in-time propagation of chaos:
\[
\sup_{t \in [0,\infty)}W_2(\rho(t),\mu^N(t))\le C\,N^{-1/2}.
\]
In fact, for the (unregularized) SVGD, \cite{liu2024towards} showed that this holds and this should naturally extend to the regularized SVGD setup with $\nu$ close to $1$. We expect this result should, in fact, hold for all $\nu \in (0,1)$.

\begin{remark}
    The bi-linear kernel $k(x,y)=1+x^\top y$ does not satisfy the boundedness assumption in Assumption~\ref{assump:kernel}-(3). However, note that our central bounds in Equations~\ref{eq:convergence of regularized Stein fisher} and~\ref{eq:KL decay} do not require Assumption~\ref{assump:kernel}-(3) and hold for bi-linear kernels as well. When the target is Gaussian, the term $\mb{E}_{\underline{x}\sim p^N(t)}[C^*(\underline{x})]$ can be explicitly computed and bounded.
\end{remark}

\end{document}